\documentclass[lettersize,journal]{IEEEtran}
\usepackage{siunitx}
\usepackage{amsmath,amsfonts}
\usepackage{algorithmic}
\usepackage{algorithm}
\usepackage{array}
\usepackage[caption=false,font=normalsize,labelfont=sf,textfont=sf]{subfig}
\usepackage{textcomp}
\usepackage{stfloats}
\usepackage{url}
\usepackage{verbatim}
\usepackage{graphicx}
\DeclareSIUnit{\Nm}{N.m}
\DeclareSIUnit{\Nms}{N.m.s}
\DeclareSIUnit{\Ns}{N.s}
\DeclareSIUnit{\kgmsquare}{kg.m^2}
\DeclareSIUnit{\revolution}{rev}

\usepackage{cite}
\usepackage{nomencl}
\usepackage{xcolor}

\usepackage{comment}

\makenomenclature

\renewcommand{\nompreamble}{The following abbreviations and symbols will be used in forthcoming equations.}
\usepackage{etoolbox}
\renewcommand\nomgroup[1]{%
  \item[\bfseries
  \ifstrequal{#1}{R}{Heavy-Duty Manipulator:}{%
  \ifstrequal{#1}{M}{Manipulator Parameters:}{%
  \ifstrequal{#1}{E}{EMLA Parameters:}{%
  \ifstrequal{#1}{A}{Abbreviations:}{}}}}%
]}

\newtheorem{defka}{Definition}
\newtheorem{thm}{Theorem}

\newtheorem{prop}{Property}

\renewcommand{\figurename}{Fig.} 

\begin{document}

© 2026 IEEE. Personal use of this material is permitted.
Permission from IEEE must be obtained for all other uses,
including reprinting/republishing this material for advertising
or promotional purposes, collecting new collected works
for resale or redistribution to servers or lists, or reuse of
any copyrighted component of this work in other works.
This work has been submitted to the IEEE for possible
publication. Copyright may be transferred without notice,
after which this version may no longer be accessible.

\title{All-Electric Heavy-Duty Robotic Manipulator: \\ Surrogate-Enhanced Modeling and Adaptive Modular Control}

\author{Amir Hossein~Barjini,
        Mohammad~Bahari, Mahdi~Hejrati,
        and~Jouni~Mattila
        \thanks{Funding for this research was provided by the Research Council of Finland under the Project ``Nonlinear PDE-model-based control of flexible manipulators'' (Grant No. 355664) and the Business Finland partnership project ``Future All-Electric Rough Terrain Autonomous Mobile Manipulators'' (Grant No. 2334/31/2022).}
        \thanks{All authors are with the Department of Engineering and Natural Sciences,
        Tampere University, 7320 Tampere, Finland (e-mails: amirhossein.barjini@tuni.fi, mohammad.bahari@tuni.fi, mahdi.hejrati@tuni.fi, jouni.mattila@tuni.fi).}
        \thanks{The first two authors contributed equally to this work.}
}

\maketitle

\begin{abstract}
This paper presents a unified system-level modeling and control framework for an all-electric heavy-duty robotic manipulator (HDRM) driven by electromechanical linear actuators (EMLAs). A surrogate-enhanced actuator model, combining electromechanical dynamics with a trained neural network using experimental data, is integrated into an extended virtual decomposition control (VDC) architecture augmented by a natural adaptation law. The derived analytical HDRM model supports a hierarchical control structure that seamlessly maps high‐level force and velocity objectives to real‐time actuator commands, accompanied by a Lyapunov‐based stability proof. In multi-domain simulations of both cubic and a custom planar triangular trajectories, the proposed adaptive modular controller achieves sub-centimeter Cartesian tracking accuracy. Experimental validation of the 1-DoF platform under realistic load emulation confirms the efficacy of the proposed control strategy. These findings demonstrate that a surrogate-enhanced EMLA model embedded in the VDC approach can enable modular, real-time control of an all-electric HDRM, supporting its deployment in next-generation mobile working machines.
\end{abstract}



\nomenclature[A]{\(\textbf{EMLA}\)}{Electromechanical Linear Actuator}
\nomenclature[A]{\(\textbf{HLA}\)}{Hydraulic Linear Actuator}
\nomenclature[A]{\(\textbf{HDRM}\)}{Heavy-Duty Robotic Manipulator}
\nomenclature[A]{\(\textbf{PMSM}\)}{Permanent Magnet Synchronous Motor}
\nomenclature[A]{\(\textbf{MWM}\)}{Mobile Working Machine}
\nomenclature[A]{\(\textbf{BEV}\)}{Battery Electric Vehicle}
\nomenclature[A]{\(\textbf{DoF}\)}{Degree-of-Freedom}
\nomenclature[A]{\(\textbf{ICE}\)}{Internal Combustion Engine}
\nomenclature[A]{\(\textbf{VDC}\)}{Virtual Decomposition Control}
\nomenclature[A]{\(\textbf{DNN}\)}{Deep Neural Network}
\nomenclature[A]{\(\textbf{RMSE}\)}{Root Mean Squared Error}
\nomenclature[A]{\(\textbf{VPF}\)}{Virtual Power Flow}
\nomenclature[A]{\(\textbf{VCP}\)}{Virtual Cutting Point}

\nomenclature[E]{\(v_d, v_q\)}{Stator voltages in the $dq$ reference frame (\si{\volt})}
\nomenclature[E]{\(i_d, i_q\)}{Stator currents in the $dq$ reference frame (\si{\ampere})}
\nomenclature[E]{\(R_s\)}{Stator winding resistance (\si{\ohm})}
\nomenclature[E]{\(\lambda_d, \lambda_q\)}{Stator flux linkages in the $dq$ frame (\si{\weber})}
\nomenclature[E]{\(\lambda_m\)}{Permanent magnet flux linkage (\si{\weber})}
\nomenclature[E]{\(L_d, L_q\)}{$dq$-axis inductances (\si{\henry})}
\nomenclature[E]{\(\tau_e\)}{Electromagnetic torque of the PMSM (\si{\Nm})}
\nomenclature[E]{\(p\)}{Number of pole pairs}

\nomenclature[E]{\(\tau_C\)}{Coulomb friction of the PMSM (\si{\Nm})}
\nomenclature[E]{\(J_m\)}{Rotor inertia of the PMSM (\si{\kgmsquare})}
\nomenclature[E]{\(C_m\)}{Viscous damping coefficient of the PMSM (\si{\Nms\per\radian})}
\nomenclature[E]{\(N_{\mathrm{gear}}\)}{Gearbox reduction ratio}
\nomenclature[E]{\(\eta_{\mathrm{gear}}\)}{Mechanical efficiency of the gearbox}
\nomenclature[E]{\(\dot{\theta}_m\)}{Angular velocity of PMSM rotor (\si{\radian\per\second})}
\nomenclature[E]{\(\ddot{\theta}_m\)}{Angular acceleration of PMSM rotor (\si{\radian\per\second\squared})}
\nomenclature[E]{\(\dot{\theta}_s\)}{Angular velocity of the screw shaft (\si{\radian\per\second})}
\nomenclature[E]{\(\ddot{\theta}_s\)}{Angular acceleration of the screw shaft (\si{\radian\per\second\squared})}
\nomenclature[E]{\(\tau_s\)}{Torque applied to the screw shaft (\si{\Nm})}
\nomenclature[E]{\(\tau_s'\)}{Torque reflected from the screw-side load to the PMSM shaft (\si{\Nm})}

\nomenclature[E]{\(\rho\)}{Screw lead (\si{\meter\per\revolution})}

\nomenclature[E]{\(x_n, \dot{x}_n\)}{Position and velocity of the screw nut (\si{\meter}, \si{\meter\per\second})}
\nomenclature[E]{\(x, \dot{x}\)}{Position and velocity of the actuator rod (\si{\meter}, \si{\meter\per\second})}

\nomenclature[E]{\(\tau_{\mathrm{bd}}\)}{Back-drive torque from nut-rod elastic interaction (\si{\Nm})}
\nomenclature[E]{\(C_b\)}{Damping coefficient between nut and rod (\si{\Ns\per\meter})}
\nomenclature[E]{\(K_b\)}{Equivalent stiffness between nut and rod (\si{\newton\per\meter})}

\nomenclature[E]{\(J_s\)}{Inertia of the screw shaft (\si{\kgmsquare})}
\nomenclature[E]{\(C_s\)}{Viscous damping of the screw shaft (\si{\Nms\per\radian})}

\nomenclature[E]{\(M_{\mathrm{act}}\)}{Effective mass of the actuator rod and attached load (\si{\kilogram})}
\nomenclature[E]{\(C_{\mathrm{act}}\)}{Damping coefficient of the actuator rod (\si{\Ns\per\meter})}
\nomenclature[E]{\(F_{\mathrm{ext}}\)}{External force acting on the actuator output (\si{\newton})}

\nomenclature[E]{\(\ell_s\)}{Effective engaged length of the screw shaft under axial loading (\si{\meter})}
\nomenclature[E]{\(L_s\)}{Total axial distance between the screw shaft bearings (\si{\meter})}
\nomenclature[E]{\(K_s\)}{Axial stiffness of the screw shaft (\si{\newton\per\meter})}
\nomenclature[E]{\(K_{\mathrm{br}}\)}{Stiffness of the bearing elements (\si{\newton\per\meter})}
\nomenclature[E]{\(K_n\)}{Stiffness of the nut interface (\si{\newton\per\meter})}
\nomenclature[E]{\(K_r\)}{Axial stiffness of the rod (\si{\newton\per\meter})}
\nomenclature[E]{\(K_{\mathrm{rot}}\)}{Torsional stiffness of the screw shaft (\si{\Nm\per\radian})}
\nomenclature[E]{\(\eta_{\mathrm{EMLA}}\)}{Efficiency of the EMLA}
\nomenclature[E]{\(\hat{F}_{\mathrm{ext}}\)}{Predicted EMLA output force obtained from the DNN (\si{\newton})}
\nomenclature[E]{\(\hat{\dot{x}}\)}{Predicted EMLA linear velocity obtained from the DNN (\si{\meter\per\second})}
\nomenclature[E]{\(\hat{P}_{\mathrm{out}}\)}{Predicted EMLA mechanical output power obtained from the DNN (\si{\watt})}
\nomenclature[E]{\({P}_{\mathrm{in}}\)}{Predicted electrical input power of EMLA obtained from the DNN (\si{\watt})}
\nomenclature[E]{\(\hat{\eta}_{\text{EMLA}}\)}{Predicted EMLA efficiency by the DNN model}
\nomenclature[E]{\(\alpha\)}{Blending coefficient between physics-based and data-driven models}

\nomenclature[E]{\(P_{\mathrm{loss}}\)}{Cumulative PMSM losses including electrical and mechanical losses (\si{\watt})}
\nomenclature[E]{\(P_{\mathrm{sw}}\)}{Switching losses in the inverter (\si{\watt})}
\nomenclature[E]{\(P_{\mathrm{cu}}\)}{Copper conduction losses in the stator windings (\si{\watt})}
\nomenclature[E]{\(P_{\mathrm{core}}\)}{Core losses (hysteresis and eddy currents) in the PMSM (\si{\watt})}
\nomenclature[E]{\(P_{\mathrm{mech}}\)}{Mechanical losses including windage and bearing friction (\si{\watt})}

\nomenclature[E]{\(N\)}{Number of samples in the training dataset}
\nomenclature[E]{\(\mathcal{F}_{\mathrm{DNN}}(\cdot)\)}{Combined Function Representation of the Trained DNN Estimating Actuator Force and Velocity}
\nomenclature[E]{\(\boldsymbol{x}, \boldsymbol{y}\)}{Input and output vectors of the neural network (context-dependent units)}
\nomenclature[E]{\(\boldsymbol{W}_\ell, \boldsymbol{b}_\ell\)}{Weight matrix and bias vector of layer $\ell$ in the neural network}
\nomenclature[E]{\(L\)}{Total number of layers in the DNN}
\nomenclature[M]{$\mathbf{{}^{A}\boldsymbol{v}}$}{Linear velocity vector}
\nomenclature[M]{$\mathbf{{}^{A}\boldsymbol{\omega}}$}{Angular velocity vector}
\nomenclature[M]{$\mathbf{{}^{A}}{\boldsymbol{V}}$}{Linear/angular velocity vector}
\nomenclature[M]{$\mathbf{{}^{A}\boldsymbol{f}}$}{Force vector}
\nomenclature[M]{$\mathbf{{}^{A}\boldsymbol{\tau}}$}{Moment vector}
\nomenclature[M]{$\mathbf{{}^{A}}{\boldsymbol{F}}$}{Force/moment vector}
\nomenclature[M]{${ }^{\mathbf{A}} \mathbf{U}_{\mathbf{B}}$}{Transformation matrix}
\nomenclature[M]{${ }^{\mathbf{A}} \mathbf{R}_{\mathbf{B}}$}{Rotation matrix}
\nomenclature[M]{${ }^{\mathbf{A}} \mathbf{r}_{\mathbf{A B}}$}{Distance between \{$\mathbf{A}$\} and \{$\mathbf{B}$\}, expressed in \{$\mathbf{A}$\}}
\nomenclature[M]{$\mathbf{{}^{A}}{\boldsymbol{F}^*}$}{Net force/moment vector}
\nomenclature[M]{$\boldsymbol{M}_A$}{Mass matrix}
\nomenclature[M]{$\boldsymbol{C}_A$}{Matrix of coriolis and centrifugal terms}
\nomenclature[M]{$\mathbf{G}_{\mathbf{A}}$}{Gravitational force/moment vector term}
\nomenclature[M]{$\lambda$}{Positive Control Parameter}
\nomenclature[M]{$\mathbf{{}^{A}}{\boldsymbol{F}_r}$}{Required force/moment vector}
\nomenclature[M]{$\mathbf{{}^{A}}{\boldsymbol{F}_r^*}$}{Required net force/moment vector}
\nomenclature[M]{$\mathbf{K_A}$}{Positive-definite gain matrix}
\nomenclature[M]{$\mathbf{{}^{A}}{\boldsymbol{V}_r}$}{Required linear/angular velocity vector}
\nomenclature[M]{$\zeta_i$}{Joint angles}
\nomenclature[M]{$\beta_i$}{Positive constant angles of the second object}
\nomenclature[M]{$\gamma_i$}{Positive angles of the third object}
\nomenclature[M]{$\dot{\boldsymbol{\Pi}}$}{Linear/angular velocity vector of the end-effector}
\nomenclature[M]{$\boldsymbol{J}$}{Jacobian matrix}
\nomenclature[M]{$\Theta$}{Joint angle vector}
\nomenclature[M]{$\Lambda$}{Constant Vector of positive control parameters}
\nomenclature[M]{$r_B$}{Radius of the base mechanism}
\nomenclature[M]{$K_i$}{Positive current control gain}
\nomenclature[M]{$K_v$}{Positive velocity control gain}
\nomenclature[M]{$K_f$}{Positive force control gain}


\section{Introduction}
\label{sec:introduction}
\IEEEPARstart{R}{obot} manipulation plays a crucial role across a broad range of applications, encompassing both mobile systems~\cite{wang2023two} and stationary manipulators~\cite{ostyn2024improving,barjini2024deep}, popular examples of which are heavy-duty robotic manipulators (HDRMs), utilized in such industrial sectors as off-road machinery, construction, agriculture, and forestry~\cite{mattila2017survey}. However, challenges to achieving dexterous and adaptable manipulation capabilities remain~\cite{billard2019trends}. In particular, as the number of degrees of freedom (DOF) increases, the control of high-DoF robotic manipulators can become more complex~\cite{hejrati2025impact}.
\subsection{Controlling Heavy-Duty Robotic Manipulators}
\label{subsec:HDRM_introduction}
To address the complex challenges associated with controlling HDRMs, a range of advanced strategies has been explored in recent studies. For instance, adaptive robust control techniques \cite{mohanty2010integrated} aim to compensate for modeling uncertainties and external disturbances, while nonlinear model predictive control (NMPC) \cite{mononen2019nonlinear} enables optimal trajectory tracking under system constraints. Further, adaptive neural network-based control \cite{liang2024adaptive} leverages learning capabilities to handle nonlinearities and time-varying dynamics, and data-driven reinforcement learning approaches \cite{yao2023data} have emerged as promising tools for developing control policies through direct interaction with the environment, without requiring an explicit model. In addition, hybrid control approaches that combine backstepping, sliding mode, and neural network techniques have also been proposed to manage complex nonlinearities and completely unknown system dynamics in electro-hydraulic actuators \cite{truong2023backstepping}. Meanwhile, another widely adopted method is virtual decomposition control (VDC), a modular and model-based control approach \cite{zhu2010virtual} that, due to its structured design and performance, has attracted considerable attention in recent years and has been widely applied to the motion control of HDRMs \cite{koivumaki2015stability, hejrati2025orchestrated, petrovic2022mathematical}. A comparative analysis also demonstrated that VDC achieves an acceptable tracking error at higher speeds \cite{hejrati2025orchestrated} with validated performance in high-DoF systems, particularly through its implementation in a 13-DoF bilateral teleoperation framework for HDRM in \cite{hejrati2025robust}.

With the increasing adoption of electrified actuation systems, considerable attention has been directed toward developing control strategies specifically suited for electric manipulators driven by permanent magnet synchronous motors (PMSMs). Adaptive neural sliding mode control approaches have thus been introduced to enhance speed regulation and robustness against dynamic uncertainties in PMSM systems \cite{dat2023advanced}. In addition, model-free and observer-based sliding mode control techniques have been proposed to address the limitations of model dependency by estimating unmodeled dynamics and external disturbances through robust observers \cite{zhao2022model}. Robust model predictive control (MPC) has also been applied to three-phase PMSM drives, demonstrating improved performance in terms of current tracking accuracy and resilience to parameter variations \cite{niu2020robust}. Similarly, preview-based closed-loop control has also been investigated for electromechanical actuators to achieve robust and reliable operation in the presence of significant actuation delays \cite{4162489}. In effect, these advancements underscore the growing maturity of control solutions tailored for electrically actuated manipulators.

\subsection{Electrification of Heavy-Duty Robotic Manipulators}
\label{subsec:electrification}
The escalating urgency of the climate crisis, alongside a growing international consensus on the need for carbon neutrality, has significantly accelerated the shift toward electrified systems across both the transportation and off-highway machinery sectors~\cite{sorknaes2022electrification,bellocchi2020electrification,9826430}. Regulatory frameworks, such as the Paris Agreement~\cite{unfccc2015paris}, the European Union (EU)'s proposed 2035 ban on internal combustion engine (ICE) vehicles~\cite{eu2035ban}, and Advanced Motor Fuels (AMF) Annex 50~\cite{amf_annex50} have set ambitious targets for reducing emissions, thereby encouraging the development of low-emission powertrains and electric actuation systems in mobile working machines (MWMs)~\cite{el2022electrical,10328391}.
In response to the broader push toward electrification, both industry and academia have increasingly focused on developing modular, battery-compatible mechanisms that support scalable deployment across various classes of MWMs. One representative category is vehicle-mounted lifting and manipulation mechanisms, including loader cranes, tail lifts, and articulated booms, which occupy a transitional space between conventional MWMs and fully coordinated HDRMs~\cite{ramos2024study}. 

Traditionally powered by hydraulic linear actuators (HLAs), these systems offer a valuable platform for evaluating the integration of electromechanical linear actuators (EMLAs) under realistic load and duty-cycle conditions \cite{bahari2023performance}. Through such intermediate applications, researchers have shown that electromechanical-based architectures improve energy efficiency and enable regenerative energy recovery \cite{qu2023electrified}, and these findings provide a practical foundation for extending EMLA adoption into HDRMs, where minimizing total energy consumption and maximizing battery life are essential to the viability of fully electric systems \cite{paz2024energy}.
The replacement of HLAs with EMLAs, which are typically driven by PMSMs and integrated with high-efficiency roller-screw mechanisms, enables precise~\cite{alipour2017performance}, high-response~\cite{10102552}, smooth~\cite{tootoonchian2016cogging}, and clean actuation \cite{bahari2025system}. In addition, EMLAs also support direct electrical control of linear motion, enhanced fault detection and isolation, and predictable performance under varying load conditions \cite{10176298,zhai2025adaptive}, with recent advances in position sensing having further enabled the high‐precision feedback needed for PMSMs of EMLA~\cite{8854990,8839808,9989447}. 

Notably, by demonstrating the feasibility of all-electric actuation in real-world, industrial‐grade machinery, the fully electric Bobcat T7X compact track loader~\cite{bobcat2022t7x} and the Volvo EX03 mid-sized electric excavator~\cite{volvo2021ex03} represent significant milestones, showcasing technical maturity, commercial readiness, and operational advantages in rugged, high‐power applications with clean operation~\cite{beltrami2021electrification,pate2025industry}.
While these achievements underscore the viability of all-electric actuation in MWMs, this also opens an opportunity to investigate how such technologies can be extended to more dynamically complex, multi-DoF HDRMs, for which integrated modeling and coordinated control are critical.

\subsection{Motivations and Research Gap}
\label{subsec:research_gap}
EMLAs represent promising alternatives to traditional hydraulic systems for HDRMs, delivering clean actuation, improved energy efficiency, enhanced control precision, and significantly reduced maintenance requirements. Driven by escalating sustainability objectives and regulatory demands, the transition toward all-electric HDRMs has become increasingly essential. However, integrating EMLAs into multi-DoF manipulators poses complicated multi-domain interaction and control challenges. Despite notable progress in the actuator-level modeling and control design of single-DoF EMLAs~\cite{al2024output,liu2019impedance,ma2024adaptive}, extending these efforts to comprehensive all-electric HDRM assemblies with multiple interacting joints remains underexplored~\cite{bahari2024system}. Therefore, ensuring effective operation and the realization of the full benefits of EMLA-driven HDRMs depends on a rigorous system-level modeling and coordinated control strategy to enable accurate motion tracking and ensure robust responsiveness to disturbances~\cite{10879599,10466505,9727198}.

Motivated by these research gaps and the necessity to employ a scalable modeling and control strategy for all-electric HDRMs, there is a critical need for an integrated framework capable of capturing the synergistic interaction between actuator- and manipulator-level dynamics. In this context, the VDC framework, recognized for its modular and model-based formulation, provides a compelling foundation, but to the best of the authors' knowledge, its practical applications have thus far been limited to hydraulically actuated HDRMs. Thus, extending VDC to incorporate high-fidelity, empirically validated EMLA models that ensure rigorous physical consistency, capture essential electromechanical interactions, and guarantee closed-loop stability is imperative.

\subsection{Contributions and Structure of the Paper}
\label{subsec:contribution}
This paper addresses the paradigm shift toward HDRM electrification by developing a comprehensive, high-fidelity modeling and adaptive modular control framework for an all-electric 6-DoF manipulator actuated exclusively by EMLAs. By leveraging the VDC methodology, the presented approach captures the intricate interactions between actuator- and manipulator-level dynamics, ensuring a robust performance against parametric uncertainties under realistic conditions. The proposed framework aims to contribute to the ongoing transition toward all-electric HDRMs by addressing key modeling and control challenges that have limited the deployment of fully electric manipulation systems.

The methodological framework and primary contributions of this study are summarized as follows:
\begin{itemize}
\item Development of a surrogate-enhanced EMLA modeling framework that integrates empirical data with analytical actuator dynamics and that accurately captures nonlinear behaviors, energy conversion inefficiencies, and unmodeled losses to achieve a hybrid model of the actuation mechanism.
\item Extension of the VDC framework to systematically incorporate EMLA behavior and handle closed-chain mechanisms, including the modeling and control of four-bar mechanisms within a unified structure.
\item \textcolor{black}{Design of a surrogate-enhanced low-level EMLA controller for simultaneous current, force, and velocity regulation, integrated within an adaptive modular VDC framework for coordinated control.}
\item \textcolor{black}{Lyapunov-based stability proof demonstrating that the combined rigid-body subsystems and low-level EMLA controllers yield an asymptotically stable closed-loop system.}
\item Comprehensive validation of the proposed framework through simulations and experiments on a 1-DoF EMLA testbed, demonstrating accurate trajectory tracking under cubic and planar motions, robustness to parametric uncertainties, and consistent performance using joint-level profiles derived from 6-DoF manipulator simulations.
\end{itemize}

The remainder of the paper is organized as follows. Section~\ref{sec:mathematical_foundation} presents the theoretical and mathematical foundations of the VDC framework and derives the complete electromechanical model of the EMLA mechanism, covering rigid-body kinematics and dynamics, velocity/force transformations, the natural adaptation law (NAL), stability definitions, and the actuator’s \textit{dq}-frame electrical and mechanical dynamics. Section~\ref{sec:modeling} outlines the building of the unified system model: first the analytical HDRM model is presented, with a systematic kinematic chain formulation for each link and recursive force/moment dynamics under VDC; then, the surrogate-enhanced hybrid EMLA model is developed by blending model-based dynamics with data-driven corrections. In Section~\ref{sec:control}, we embed the actuator model into an adaptive modular VDC controller, derive the required joint and task-space velocities and forces, design low-level voltage and current controllers for each EMLA, and provide a Lyapunov-based stability proof. Section~\ref{sec:results_and_discussion} evaluates system performance through cubic and custom planar trajectories, first in simulation by analyzing tracking accuracy, and then experimentally on a 1-DoF testbed. Finally, Section~\ref{sec:conclusion} summarizes the contributions and outlines avenues for future research.

\section{Mathematical Foundations of VDC and the EMLA Mechanism}
\label{sec:mathematical_foundation}
Before presenting the main contributions, this section introduces fundamental concepts related to the VDC approach and the dynamic modeling of EMLAs. 

\subsection{VDC Preliminaries}
\label{subsec:VDC_preliminaries}
In this section, the VDC approach~\cite{zhu2010virtual}, with its mathematical foundation, is introduced to be used for dynamics modeling, controller design, and stability analysis of the manipulator. 

\subsubsection{Kinematics and Dynamics of a Rigid Body}
\label{subsubsec:kinematics_and_dynamics}
Let $\{\mathbf{A}\}$ be a three-dimensional orthogonal coordinate system, attached to a rigid body. Considering $\mathbf{{}^{A}\boldsymbol{v}}$ $\in \mathbb{R}^3$ and $\mathbf{{}^{A}\boldsymbol{\omega}}$ $\in \mathbb{R}^3$ as the linear and angular velocity vectors, expressed in frame $\{\mathbf{A}\}$, the linear/angular velocity vector of frame $\{\mathbf{A}\}$ is defined as:

\begin{equation}
\begin{aligned}
\mathbf{{}^{A}}{\boldsymbol{V}} &\stackrel{\textit{def}}{=}\left[\begin{array}{c}
\mathbf{{}^{A}\boldsymbol{v}}  \\
\mathbf{{}^{A}\boldsymbol{\omega}} 
\end{array}\right] &\in \mathbb{R}^6 
\end{aligned}.
\label{Eq:linear_angular}
\end{equation}

Similarly, let $\mathbf{{}^{A}\boldsymbol{f}}$ $\in \mathbb{R}^3$ and $\mathbf{{}^{A}\boldsymbol{\tau}}$ $\in \mathbb{R}^3$ be the force and moment vectors that are applied, respectively, and expressed in frame $\{\mathbf{A}\}$. Therefore, the force/moment vector in frame $\{\mathbf{A}\}$ is defined as:

\begin{equation}
\begin{aligned}
\mathbf{{}^{A}}{\boldsymbol{F}} &\stackrel{\textit{def}}{=}\left[\begin{array}{c}
\mathbf{{}^{A}\boldsymbol{f}}  \\
\mathbf{{}^{A}\boldsymbol{\tau}} 
\end{array}\right] &\in \mathbb{R}^6 
\end{aligned}.
\label{Eq:Force_Moment}
\end{equation}

It should be clarified that in VDC, as a systematic approach, the linear/angular velocity vector ($\boldsymbol{V} \in \mathbb{R}^6$) is computed recursively from the base of the robot to the end-effector, while the force/moment vector ($\boldsymbol{F} \in \mathbb{R}^6$) is computed recursively from the end-effector to the base, using the transformations between coordinate frames. Now, consider frame $\{\mathbf{B}\}$ to be attached to the same rigid body. Consequently, we can transform the linear/angular velocity vector and force/moment vector between these two frames, as follows:

\begin{equation}
\left\{
\begin{aligned}
\mathbf{{}^{B}}{\boldsymbol{V}} & ={ }^{\mathbf{A}} \mathbf{U}_{\mathbf{B}}^T \mathbf{{}^{A}}{\boldsymbol{V}}, \\
\mathbf{{}^{A}}{\boldsymbol{F}} & ={ }^{\mathbf{A}} \mathbf{U}_{\mathbf{B}}{ }^{\mathbf{B}} {\boldsymbol{F}},
\end{aligned}
\right.
\label{Eq:Transformation}
\end{equation}
where ${ }^{\mathbf{A}} \mathbf{U}_{\mathbf{B}}$ $\in \mathbb{R}^{6\times6}$ denotes the transformation matrix from frame $\{\mathbf{B}\}$ to frame $\{\mathbf{A}\}$, as follows:

\begin{equation}
{ }^{\mathbf{A}} \mathbf{U}_{\mathbf{B}}=\left[\begin{array}{cc}
{ }^{\mathbf{A}} \mathbf{R}_{\mathbf{B}} & \mathbf{0}_{3 \times 3} \\
\left({ }^{\mathbf{A}} \mathbf{r}_{\mathbf{A B}} \times\right) { }^{\mathbf{A}} \mathbf{R}_{\mathbf{B}} & { }^{\mathbf{A}} \mathbf{R}_{\mathbf{B}}
\end{array}\right] ,
\label{Eq:Transformation_Matrix}
\end{equation}
where ${ }^{\mathbf{A}} \mathbf{R}_{\mathbf{B}}$ $\in \mathbb{R}^{3\times3}$ is the rotation matrix from frame $\{\mathbf{B}\}$ to frame $\{\mathbf{A}\}$. Moreover, if ${ }^{\mathbf{A}} \mathbf{r}_{\mathbf{A B}} = [r_{\mathrm{x}}, r_{\mathrm{y}}, r_{\mathrm{z}}]^T$ is the vector from the origin of frame $\{\mathbf{A}\}$ to the origin of frame $\{\mathbf{B}\}$, expressed in frame $\{\mathbf{A}\}$, then ${ }^{\mathbf{A}} \mathbf{r}_{\mathbf{A B}} \times \in \mathbb{R}^{3 \times 3}$ is a skew-symmetric matrix operator defined as:
\begin{equation}
\left({ }^{\mathbf{A}} \mathbf{r}_{\mathbf{A B}} \times\right)=\left[\begin{array}{ccc}
0 & -r_{\mathrm{z}} & r_{\mathrm{y}} \\
r_{\mathrm{z}} & 0 & -r_{\mathrm{x}} \\
-r_{\mathrm{y}} & r_{\mathrm{x}} & 0
\end{array}\right].
\label{Eq:X_Operator}
\end{equation}

The net force/moment vectors of the rigid body in frame $\{\mathbf{A}\}$, defined as ${}^{\mathbf{A}} \boldsymbol{F}^* \in \mathbb{R}^6$, are calculated based on the rigid body's dynamics as:
\begin{equation}
\mathbf{M}_{\mathbf{A}} \frac{\textit{d}}{\textit{dt}}\! \left({ }^{\mathbf{A}} \boldsymbol{V}\right) + \mathbf{C}_{\mathbf{A}} \! ^{\mathbf{A}}\! \boldsymbol{V} + \mathbf{G}_{\mathbf{A}} = {}^{\mathbf{A}} \boldsymbol{F}^*,
\label{Eq:Net_Force}
\end{equation}
where $\mathbf{M}_{\mathbf{A}} \in \mathbb{R}^{6 \times 6}$ is the mass matrix, $\mathbf{C}_{\mathbf{A}} \in \mathbb{R}^{6 \times 6}$ denotes the matrix of coriolis and centrifugal terms, and $\mathbf{G}_{\mathbf{A}} \in \mathbb{R}^{6}$ corresponds to the gravity terms.

\begin{prop}
    Consider the rigid body dynamics described in (\ref{Eq:Net_Force}). The following property can be established:
    \begin{equation}
        \Tilde{Y}_{A} \theta_{A} = \mathbf{M}_{\mathbf{A}} \frac{\textit{d}}{\textit{dt}}\! \left({ }^{\mathbf{A}} \boldsymbol{V}\right) + \mathbf{C}_{\mathbf{A}} \! ^{\mathbf{A}}\! \boldsymbol{V} + \mathbf{G}_{\mathbf{A}},
    \end{equation}
    where \(\Tilde{Y}_{A}(^A\Dot{\mathcal{V}},\,^A\mathcal{V}) \in \mathbb{R}^{6\times10}\) denotes the regressor matrix and \(\theta_A(m,\,^Ar_{AB}, I_A)\) is the corresponding inertial parameter vector, as defined in \cite{hejrati2022decentralized}.
    \label{property 2}
\end{prop}

\subsubsection{Required Linear/Angular Velocity and Force/Moment Vectors}

In VDC, the required velocity differs from the desired velocity. While the desired velocity represents the intended motion, the required velocity incorporates this target along with additional terms that consider such control errors as position error. Therefore, the required joint velocities are defined as follows:
\begin{equation}
\left\{
\begin{aligned}
\dot{q}_r & = \dot{q}_d + \lambda(q_d - q) \text{ \hspace {1cm}(For revolute joints)} \\
\dot{x}_r & = \dot{x}_d + \lambda(x_d - x) \text{\hspace{1cm}(For prismatic joints)}
\end{aligned}
\right.
\label{Eq:Required_Joint_Velocity}
\end{equation}
where $\lambda$ is the positive control parameter; $\dot{q}_d$ and $\dot{x}_d$ are the desired angular and linear velocities, respectively; $q_d$ and $x_d$ are the desired angular and linear positions, and $q$ and $x$ are the measured angular and linear positions.

Once the required joint velocities are determined, the corresponding required linear/angular velocity vector in any frame $\{\mathbf{A}\}$, denoted as $\mathbf{{}^{A}\boldsymbol{V}_r} \in \mathbb{R}^6$, can be computed, which will be described in details subsequently. Eventually, once the required linear/angular velocity vectors are determined, the required net force/moment vector in frame $\{\mathbf{A}\}$, defined as ${}^{\mathbf{A}} \boldsymbol{F}^*_r \in \mathbb{R}^6$, is delineated by:
\begin{equation}
  {}^{\mathbf{A}} \boldsymbol{F}^*_r = Y_{A} \hat{\theta}_{A} + \mathbf{K_A} \left( { }^{\mathbf{A}} \boldsymbol{V_r} - { }^{\mathbf{A}} \boldsymbol{V} \right) ,
\label{Eq:Required_Force}
\end{equation}
where $\mathbf{K_A} \in \mathbb{R}^{6 \times 6}$ is a positive-definite gain matrix, \(Y_{A}(^A\Dot{\mathcal{V}_r},\,^A\mathcal{V}_r) \in \mathbb{R}^{6\times10}\) denotes the regressor matrix in the sense of Property \ref{property 2}, and \(\hat{\theta}_A\) is the estimation of \(\theta_A\), explained in the following.

\subsubsection{Natural Adaptation Law}
{For a rigid body represented by frame \{A\}, there exists a unique set of inertial parameters \mbox{\(\theta_A \in \mathbb{R}^{10}\)} that satisfy the conditions of physical consistency. As elaborated in~\mbox{\cite{hejrati2022decentralized}}, there exists a linear map \mbox{\(f: \mathbb{R}^{10} \rightarrow S(4)\)} such that:}
\begin{equation}
    f(\theta_A) = \mathcal{L}_A = \begin{bmatrix}
        0.5\,\mathrm{tr}(I_A)\cdot\mathbf{1} - I_A & h_A \\
        h_A^\top & m_A
    \end{bmatrix},
\end{equation}
\begin{equation}
    f^{-1}(\mathcal{L}_A) = \theta_A(m_A, h_A,\, \mathrm{tr}(\Sigma_A)\cdot\mathbf{1} - \Sigma_A),
\end{equation}
{where \mbox{\(m_A\)}, \mbox{\(h_A\)}, and \mbox{\(I_A\)} denote the mass, first moment of mass, and the rotational inertia tensor, respectively. The matrix \mbox{\(\mathcal{L}_A \in S(4)\)} is referred to as the pseudo-inertia matrix, with \mbox{\(S(4)\)} denoting the space of real symmetric \(4\times4\) matrices, and \mbox{\(\Sigma_A = 0.5\,\mathrm{tr}(I_A) - I_A\)}.}

{Given a true pseudo-inertia matrix \mbox{\(\mathcal{L}_A\)} and its estimate \mbox{\(\hat{\mathcal{L}}_A\)}, a Lyapunov function can be formulated using the Bregman divergence based on the log-determinant function, as follows~\cite{hejrati2023physical}:}
\begin{equation} \label{Df}
    \mathcal{D}_F(\mathcal{L}_A \,\|\, \hat{\mathcal{L}}_A) = \log\frac{|\hat{\mathcal{L}}_A|}{|\mathcal{L}_A|} + \mathrm{tr}(\hat{\mathcal{L}}_A^{-1} \mathcal{L}_A) - 4,
\end{equation}
{with the time derivative of (\mbox{\ref{Df}}) given by:}
\begin{equation}
    \Dot{\mathcal{D}}_F(\mathcal{L}_A \,\|\, \hat{\mathcal{L}}_A) = \mathrm{tr}\left(\hat{\mathcal{L}}_A^{-1} \Dot{\hat{\mathcal{L}}}_A \hat{\mathcal{L}}_A^{-1} \, \Tilde{\mathcal{L}}_A\right),
\end{equation}
{where \mbox{\(\Tilde{\mathcal{L}}_A = \hat{\mathcal{L}}_A - \mathcal{L}_A\)} represents the estimation error. Consequently, the natural adaptation law (NAL) is obtained by:}
\begin{equation}
    \Dot{\hat{\mathcal{L}}}_A = \frac{1}{\gamma} \hat{\mathcal{L}}_A \mathcal{S}_A \hat{\mathcal{L}}_A,
    \label{L adapt}
\end{equation}
where \mbox{\(\gamma > 0\)} is the adaptation gain. The matrix \mbox{\(\mathcal{S}_A\)} is a uniquely defined symmetric matrix~\cite{hejrati2022decentralized}.

\subsubsection{Virtual Power Flow (VPF) and Virtual Stability}
\begin{defka}
    Relative to frame \{A\}, the VPF is defined as the inner product between the linear/angular velocity error and the force/moment error vectors, as introduced in \mbox{\cite{zhu2010virtual}}:
    \begin{equation}\label{VPF}
    p_A = \left(\mathbf{{}^{A}}{\boldsymbol{V}}_r - \mathbf{{}^{A}}{\boldsymbol{V}}\right)^T \left(\mathbf{{}^{A}}{\boldsymbol{F}}_r - \mathbf{{}^{A}}{\boldsymbol{F}}\right),
    \end{equation}
    where \(\mathbf{{}^{A}}{\boldsymbol{V}}_r \in \mathbb{R}^6\) and \(\mathbf{{}^{A}}{\boldsymbol{F}}_r \in \mathbb{R}^6\) denote the required linear/angular velocity and force/moment vectors, respectively, corresponding to the actual vectors \(\mathbf{{}^{A}}{\boldsymbol{V}} \) and \(\mathbf{{}^{A}}{\boldsymbol{F}} \).
    \label{Def pA}
\end{defka}

\begin{defka}
    A non-negative accompanying function \(\nu(t) \in \mathbb{R}\) is a piecewise continuously differentiable function defined for all \(t \in \mathbb{R}^+\), satisfying \(\nu(0) < \infty\), and possessing a well-defined time derivative \(\Dot{\nu}(t)\) almost everywhere.
    \label{nu def}
\end{defka}

\begin{defka}
    A subsystem obtained through decomposition is said to be virtually stable if there exists a non-negative accompanying function \( \nu(t) \), such that its time derivative satisfies the inequality \cite{zhu2010virtual}:
\begin{equation}
    \dot{\nu}(t) \leq -B(t) + \sum_{\{A\} \in \Phi} p_A - \sum_{\{C\} \in \Psi} p_C,
\end{equation}
where \(B(t)\) is a positive function, \(\Phi\) and \(\Psi\) represent the sets of driven and driving virtual cutting frames of this subsystem, respectively, and \(p_A\) and \(p_C\) are the corresponding VPFs at these frames.
    \label{Def_VS}
\end{defka}
    
\begin{thm}
    Consider a complex robotic structure decomposed into subsystems with rigid body subsystem representation (\ref{Eq:Net_Force}). If each of the constituent subsystems satisfies the virtual stability condition specified in Definition~\ref{Def_VS}, then the overall system is stable \cite{zhu2010virtual}.
    \label{thm1}
\end{thm}

\subsection{\textcolor{black}{Integrated Dynamic Model of the EMLA}}
\label{subsec:emla_dynamic}
The EMLA investigated in this study consists of a PMSM driving a linear output via a gearbox and a lead screw~\cite{miric2020dynamic}. This section describes the EMLA dynamics from the electrical domain of the motor to the mechanical domain of force generation.
The electrical behavior of the PMSM in the synchronous $dq$ reference frame is governed by the following voltage and flux linkage relations~\cite{5530379}:
\begin{equation}
\left\{
\begin{aligned}
v_d &= R_s i_d + \frac{d\lambda_d}{dt} - p  \dot{\theta}_m \lambda_q \\
v_q &= R_s i_q + \frac{d\lambda_q}{dt} + p  \dot{\theta}_m \lambda_d
\end{aligned}
\right.
\label{eq:dq_voltages}
\end{equation}
where $v_d$, $v_q$ and $i_d$, $i_q$ represent the stator voltages and currents in the $dq$ frame, $R_s$ is the stator resistance, $\dot{\theta}_m$ is the mechanical angular speed, and $\lambda_d$, $\lambda_q$ are the corresponding flux linkages. For surface-mounted configurations, where $L_d = L_q$, the flux linkages simplify to $\lambda_d = L_d i_d + \lambda_m$ and $\lambda_q = L_q i_q$, with $\lambda_m$ being the flux linkage due to the permanent magnets.
To transition into the mechanical domain, the electromagnetic torque generated by the PMSM is expressed as~\cite{8948294}:
\begin{equation}
\tau_e = \frac{3}{2} p \left( \lambda_d i_q - \lambda_q i_d \right)
\label{eq:PMSM_torque}
\end{equation}
where $p$ denotes the number of pole pairs. Meanwhile, the mechanical dynamics of the PMSM rotor are governed by~\cite{hao2021coordinated}:
\begin{equation}
J_m \ddot{\theta}_m = \tau_e - \tau_s' - C_m \dot{\theta}_m -\tau_C
\label{eq:rotor_dynamics}
\end{equation}
where $J_m$ is the motor inertia, $C_m$ is the viscous damping coefficient, $\tau_C$ is the Coulomb friction, and $\tau_s'$ denotes the torque reflected to the motor shaft from the downstream screw-side load. The motor output is connected to a shaft via a gearbox with reduction ratio $ N_{\mathrm{gear}}$ and efficiency $\eta_{\mathrm{gear}}$, and the screw shaft angular velocity $\dot{\theta}_s$ is related to the motor shaft speed $\dot{\theta}_m$ as follows:
\begin{equation}
\dot{\theta}_s = \frac{1}{N_{\mathrm{gear}}} \dot{\theta}_m
\label{eq:gear_velocity}
\end{equation}
and the torque at the screw shaft is determined by:
\begin{equation}
\tau_s = \eta_{\mathrm{gear}} N_{\mathrm{gear}} \left(\tau_e - C_m \dot{\theta}_m\right)
\label{eq:gear_torque}
\end{equation}

Rotary-to-linear motion is achieved through a lead screw, characterized by its lead $\rho$ in meters per revolution. The linear velocity of the moving nut $\dot{x}_n$ is related to the screw shaft angular velocity, as follows:
\begin{equation}
\dot{x}_n = \frac{\rho}{2\pi} \dot{\theta}_s
\label{eq:linear_velocity}
\end{equation}

To model the elastic and damping effects between the screw nut and the actuator rod, a two-mass system is introduced. This formulation captures the relative compliance between the nut and rod, which may arise due to mechanical backlash, preload effects, or structural flexibilities in the screw–nut interface. The back-drive torque $\tau_{\mathrm{bd}}$ resulting from relative motion between the nut and the rod is given by:
\begin{equation}
\tau_{\mathrm{bd}} = \frac{\rho}{2\pi} \left[ C_b (\dot{x}_n - \dot{x}) + K_b (x_n - x) \right]
\label{eq:backdrive_torque}
\end{equation}
where $x_n$ and $\dot{x}_n$ are the nut's position and velocity, $x$ and $\dot{x}$ are the rod's position and velocity, and $C_b$ and $K_b$ denote the equivalent damping and stiffness terms of the screw-nut system, respectively. Thus, the rotational dynamics of the screw shaft become:
\begin{equation}
J_s \ddot{\theta}_s + C_s \dot{\theta}_s = \tau_s - \tau_{\mathrm{bd}}
\label{eq:screw_dynamics}
\end{equation}

The actuator rod, driven via the screw nut, follows the second-order linear equation, as follows:
\begin{equation}
M_{\mathrm{act}} \ddot{x} + C_{\mathrm{act}} \dot{x} + C_b (\dot{x} - \dot{x}_n) + K_b (x - x_n) = -F_{\mathrm{ext}}
\label{eq:rod_dynamics}
\end{equation}
where $M_{\mathrm{act}}$ and $C_{\mathrm{act}}$ denote the effective mass and damping of the actuator rod and its driven components, respectively, and $F_{\mathrm{ext}}$ represents the external force acting on the actuator output.
The equivalent stiffness $K_b$ of the screw-nut-rod transmission is modeled as the parallel combination of elastic elements, including shaft flexibility, bearing and interface compliance, and torsional deformation~\cite{vesterinen2024elastodynamic}:
\begin{equation}
K_b = \left( \frac{\ell_s}{K_s} + \frac{1}{2K_{\mathrm{br}}} + \frac{1}{K_n} + \frac{1}{K_r} + \frac{\rho^2 x}{4\pi^2 K_{\mathrm{rot}}} \right)^{-1}
\label{eq:ballscrew_stiffness}
\end{equation}
where the terms $K_s$, $K_{\mathrm{br}}$, $K_n$, $K_r$, and $K_{\mathrm{rot}}$ represent the stiffnesses of the screw shaft, bearing, nut interface, rod, and the torsional stiffness of the rotating parts, respectively.
The term $\ell_s$ denotes the effective length of the screw shaft under axial deformation and is defined as:
\begin{equation}
\ell_s = \frac{x (L_s - x)}{L_s}
\label{eq:shaft_effective_length}
\end{equation}
where $x$ is the actuator rod displacement and $L_s$ is the axial distance between the screw shaft bearings. The overall efficiency of the EMLA is defined later in Section~\ref{subsec:hybdrid_emla_model} as the ratio of the mechanical output power to electrical input power.

\section{System Modeling}
\label{sec:modeling}

\subsection{Analytical HDRM Model}
\label{subsec:manipulator modeling}

\begin{figure}[ht]
\centering
\includegraphics[width=2.7in]{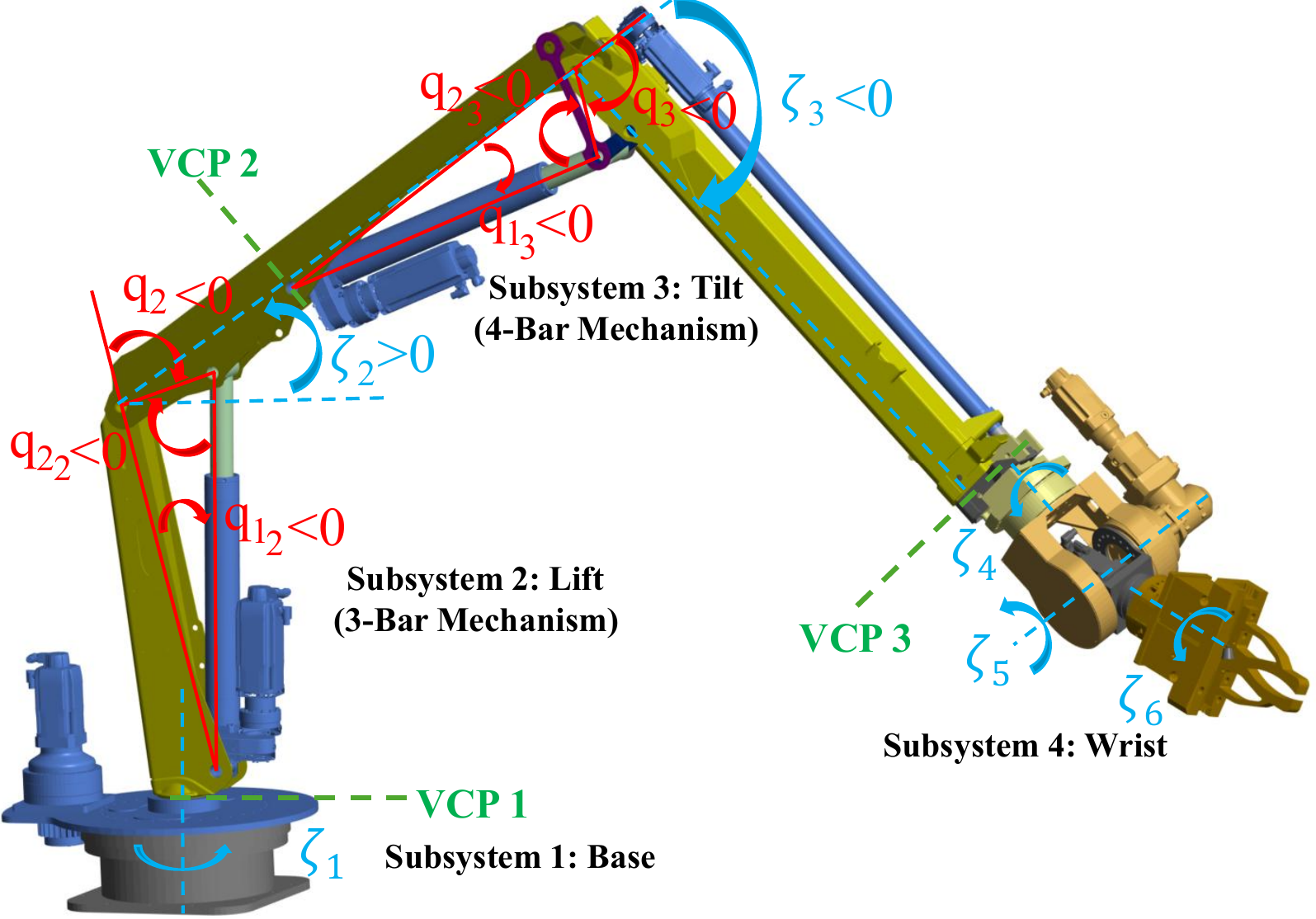}
\caption{Schematic of the electric HDRM.}
\label{Fig:SimulationPlatform}
\end{figure}

This section presents the kinematic and dynamic modeling of the manipulator, a 6-DOF electric HDRM. As illustrated in Fig. \ref{Fig:SimulationPlatform}, the manipulator is segmented into four main components: the base, lift, tilt, and wrist. Given that the system incorporates both three-bar and four-bar mechanisms, use of the VDC offers a generalized and systematic framework for deriving equations of motion, applicable to a wide range of manipulator architectures.

\subsubsection{HDRM Kinematic Formulation}
\label{subsubsec:kinematics of the manipulator}

\begin{figure*}
\centering
\includegraphics[width=4.5in]{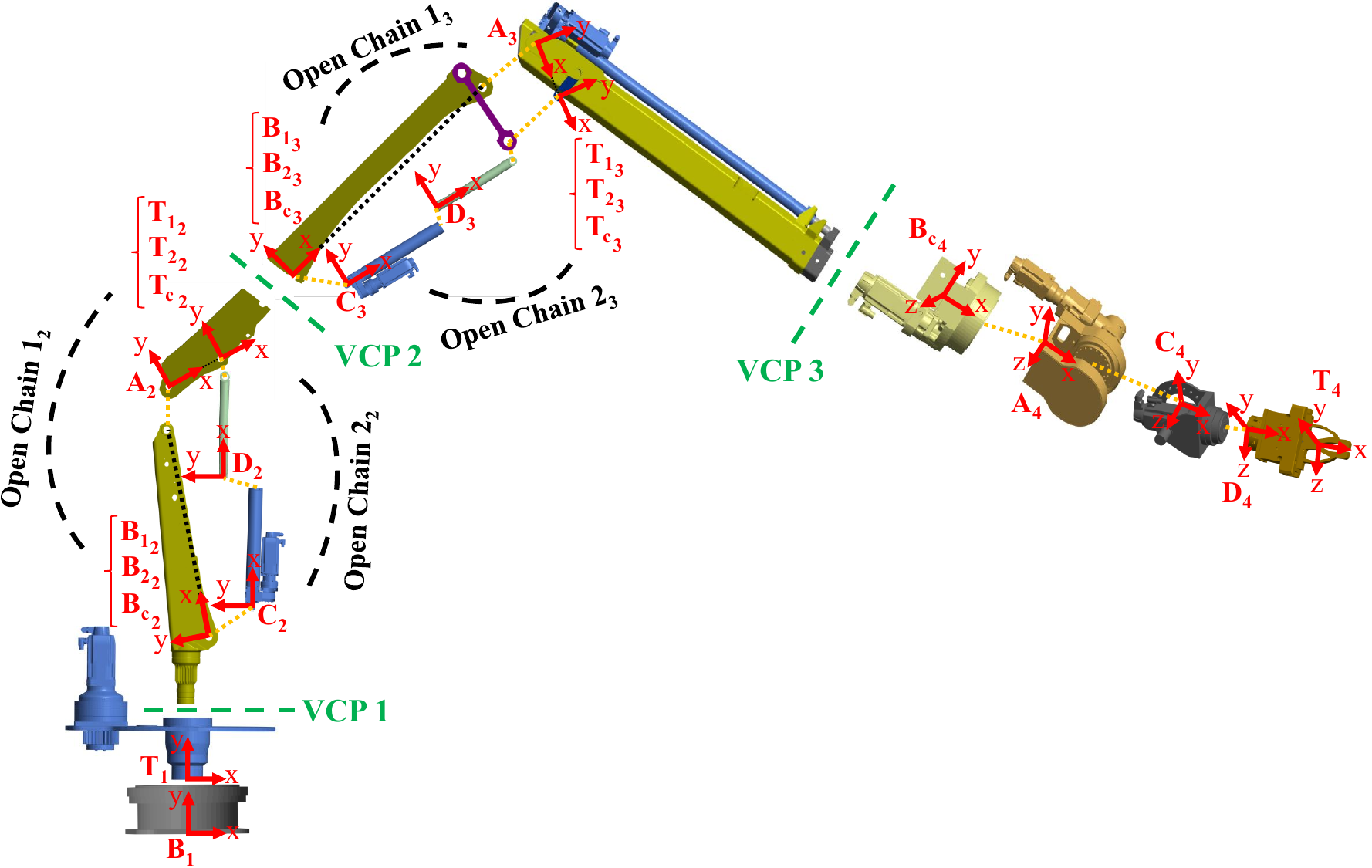}
\caption{VDC frames of the manipulator.}
\label{Fig:VDC Frames}
\end{figure*}

The manipulator’s kinematics are derived individually for each of its four structural segments. The VDC frames in conjunction with the virtual cutting points (VCPs), are presented in Fig. (\ref{Fig:VDC Frames}), though it should be mentioned that according to the VDC approach, linear/angular velocity vectors are computed recursively from the base of the robot to the wrist. In the remainder of this paper, the following parameters are used based on the VDC:
\begin{equation}
\left\{
\begin{aligned}
x_f &= [1,0,0,0,0,0]^T, \\
y_f &= [0,1,0,0,0,0]^T, \\
z_f &= [0,0,1,0,0,0]^T, \\
x_{\tau} &= [0,0,0,1,0,0]^T, \\
y_{\tau} &= [0,0,0,0,1,0]^T, \\
z_{\tau} &= [0,0,0,0,0,1]^T.
\end{aligned}
\right.
\label{Eq:Zeta}
\end{equation}

\begin{itemize}
    \item Base Motion:
\end{itemize}

According to the VDC approach, the linear/angular velocity vector for the first object is calculated as follows:
\begin{equation}
\mathbf{{}^{T_1}}{\boldsymbol{V}} = \mathbf{y_{\tau}} \dot{\zeta}_1 + { }^{\mathbf{B_1}} \mathbf{U}_{\mathbf{T_1}}^T \mathbf{{}^{B_1}}{\boldsymbol{V}},
\label{Eq:Base_Kinematics}
\end{equation}
where $\mathbf{{}^{B_1}}{\boldsymbol{V}}$ denotes the ground and is a zero vector in our study. In the next step, the linear/angular velocity vector of the second object is calculated, using the following equation:
\begin{equation}
\mathbf{{}^{B_{c_2}}}{\boldsymbol{V}} = { }^{\mathbf{T_1}} \mathbf{U}_{\mathbf{B_{c_2}}}^T \mathbf{{}^{T_1}}{\boldsymbol{V}}.
\label{Eq:Base_to_Lift_Kinematics}
\end{equation}

\begin{itemize}
    \item Lift Motion (Three-Bar Mechanism):
\end{itemize}

As it is the second object, all variables have a subscript of 2. The schematic view of the second object is shown in Fig. (\ref{Fig:3Bar}), though it should be noted that the signs of the angles $q_2$, $q_{1_2}$, $q_{2_2}$, and $\zeta_2$ are defined counterclockwise. Therefore, we have: 

\begin{figure}[ht]
\centering
\includegraphics[width=1.2in]{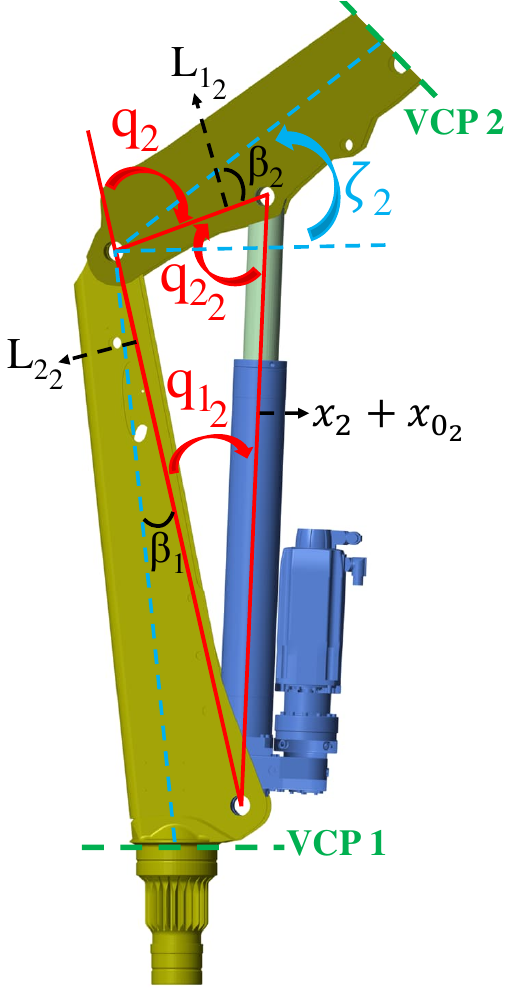}
\caption{Schematic of the lift (3-bar mechanism).}
\label{Fig:3Bar}
\end{figure}

\begin{equation}
\beta_1 + (\pi+q_2) + (\beta_2-\zeta_2) = \frac{\pi}{2},
\label{Eq:3_Bar_1}
\end{equation}
where $\beta_1$ and $\beta_2$ are two positive constant angles. Therefore, the lift angle ($q_2$) can be derived based on the measurement from the sensor ($\zeta_2$), as follows:

\begin{equation}
q_2 = \left(-\frac{\pi}{2}-\beta_1 - \beta_2 \right) + \zeta_2.
\label{Eq:3_Bar_2}
\end{equation}

Based on the law of cosines, we have:

\begin{equation}
\left\{
\begin{aligned}
x_2 & = \sqrt{L_{1_2}^2 + L_{2_2}^2 + 2L_{1_2}L_{2_2}\cos{q_2}} - x_{0_2}, \\
q_{1_2} & = -\arccos\left(\frac{L_{1_2}^2 - (x_2+x_{0_2})^2 - L_{2_2}^2}{-2 (x_2+x_{0_2}) L_{2_2}} \right), \\
q_{2_2} & = -\arccos\left(\frac{L_{2_2}^2 - (x_2+x_{0_2})^2 - L_{1_2}^2}{-2 (x_2+x_{0_2}) L_{1_2}} \right).
\end{aligned}
\right.
\label{Eq:3_Bar_3}
\end{equation}

By differentiating \eqref{Eq:3_Bar_3}, the velocity equations are derived as follows:

\begin{equation}
\left\{
\begin{aligned}
\dot{x}_2 & = -\frac{L_{1_2} L_{2_2} \sin{q_2}}{x_2 + x_{0_2}} \dot{q}_2, \\
\dot{q}_{1_2} & = - \frac{(x_2 + x_{0_2}) - L_{2_2} \cos{q_{1_2}}}{(x_2 + x_{0_2}) L_{2_2} \sin{q_{1_2}}} \dot{x}_2, \\
\dot{q}_{2_2} & = - \frac{(x_2 + x_{0_2}) - L_{1_2} \cos{q_{2_2}}}{(x_2 + x_{0_2}) L_{1_2} \sin{q_{2_2}}} \dot{x}_2,
\end{aligned}
\right.
\label{Eq:3_Bar_4}
\end{equation}
where $\dot{q}_2 = \dot{\zeta}_2$, based on (\ref{Eq:3_Bar_2}). It should be noted that (\ref{Eq:3_Bar_3}) and (\ref{Eq:3_Bar_4}) define the relation between the lift passive joint and the EMLA motion.

The linear/angular velocity vectors for the open chain $1_2$ are calculated as follows:
\begin{equation}
\left\{
\begin{aligned}
\mathbf{{}^{A_2}}{\boldsymbol{V}} & = \mathbf{z_{\tau}} \dot{q}_2 + { }^{\mathbf{B_{1_2}}} \mathbf{U}_{\mathbf{A_2}}^T \mathbf{{}^{B_{1_2}}}{\boldsymbol{V}}, \\
\mathbf{{}^{T_{1_2}}}{\boldsymbol{V}} & =  { }^{\mathbf{A_2}} \mathbf{U}_{\mathbf{T_{1_2}}}^T \mathbf{{}^{A_2}}{\boldsymbol{V}},
\end{aligned}
\right.
\label{Eq:Lift_Kinematics_Openchain1}
\end{equation}

Similarly, for the open chain $2_2$, we have:
\begin{equation}
\left\{
\begin{aligned}
\mathbf{{}^{C_2}}{\boldsymbol{V}} & = \mathbf{z_{\tau}} \dot{q}_{1_2} + { }^{\mathbf{B_{2_2}}} \mathbf{U}_{\mathbf{C_2}}^T \mathbf{{}^{B_{2_2}}}{\boldsymbol{V}}, \\
\mathbf{{}^{D_2}}{\boldsymbol{V}} & = \mathbf{x_{f}} \dot{x}_2 + { }^{\mathbf{C_2}} \mathbf{U}_{\mathbf{D_2}}^T \mathbf{{}^{C_2}}{\boldsymbol{V}}, \\
\mathbf{{}^{T_{2_2}}}{\boldsymbol{V}} & = \mathbf{z_{\tau}} \dot{q}_{2_2} + { }^{\mathbf{D_2}} \mathbf{U}_{\mathbf{T_{2_2}}}^T \mathbf{{}^{D_2}}{\boldsymbol{V}},
\end{aligned}
\right.
\label{Eq:Lift_Kinematics_Openchain2}
\end{equation}

Because the beginning and end points of the open chains $1_2$ and $2_2$ coincide, the following constraints apply:
\begin{equation}
\left\{
\begin{aligned}
\mathbf{{}^{B_{1_2}}}{\boldsymbol{V}} & = \mathbf{{}^{B_{2_2}}}{\boldsymbol{V}} = \mathbf{{}^{B_{c_2}}}{\boldsymbol{V}}, \\
\mathbf{{}^{T_{1_2}}}{\boldsymbol{V}} & = \mathbf{{}^{T_{2_2}}}{\boldsymbol{V}} = \mathbf{{}^{T_{c_2}}}{\boldsymbol{V}},
\end{aligned}
\right.
\label{Eq:Lift_Kinematics_Constraint}
\end{equation}

Eventually, the linear/angular velocity vector for the base of the third object is calculated as follows:
\begin{equation}
\mathbf{{}^{B_{c_3}}}{\boldsymbol{V}} = { }^{\mathbf{T_{c_2}}} \mathbf{U}_{\mathbf{B_{c_3}}}^T \mathbf{{}^{T_{c_2}}}{\boldsymbol{V}}.
\label{Eq:Lift_to_Tilt_Kinematics}
\end{equation}

\begin{itemize}
    \item Tilt Motion (Four-Bar Mechanism):
\end{itemize}

In this section, we derive the kinematic equations governing tilt motion, which is a four-bar linkage mechanism. As this corresponds to the third object in the system, all related variables carry the subscript 3. Note that the angles $q_3$, $q_{1_3}$, $q_{2_3}$, and $\zeta_3$ are defined as positive in the counterclockwise direction.

\begin{figure}[ht]
\centering
\includegraphics[width=3.5in]{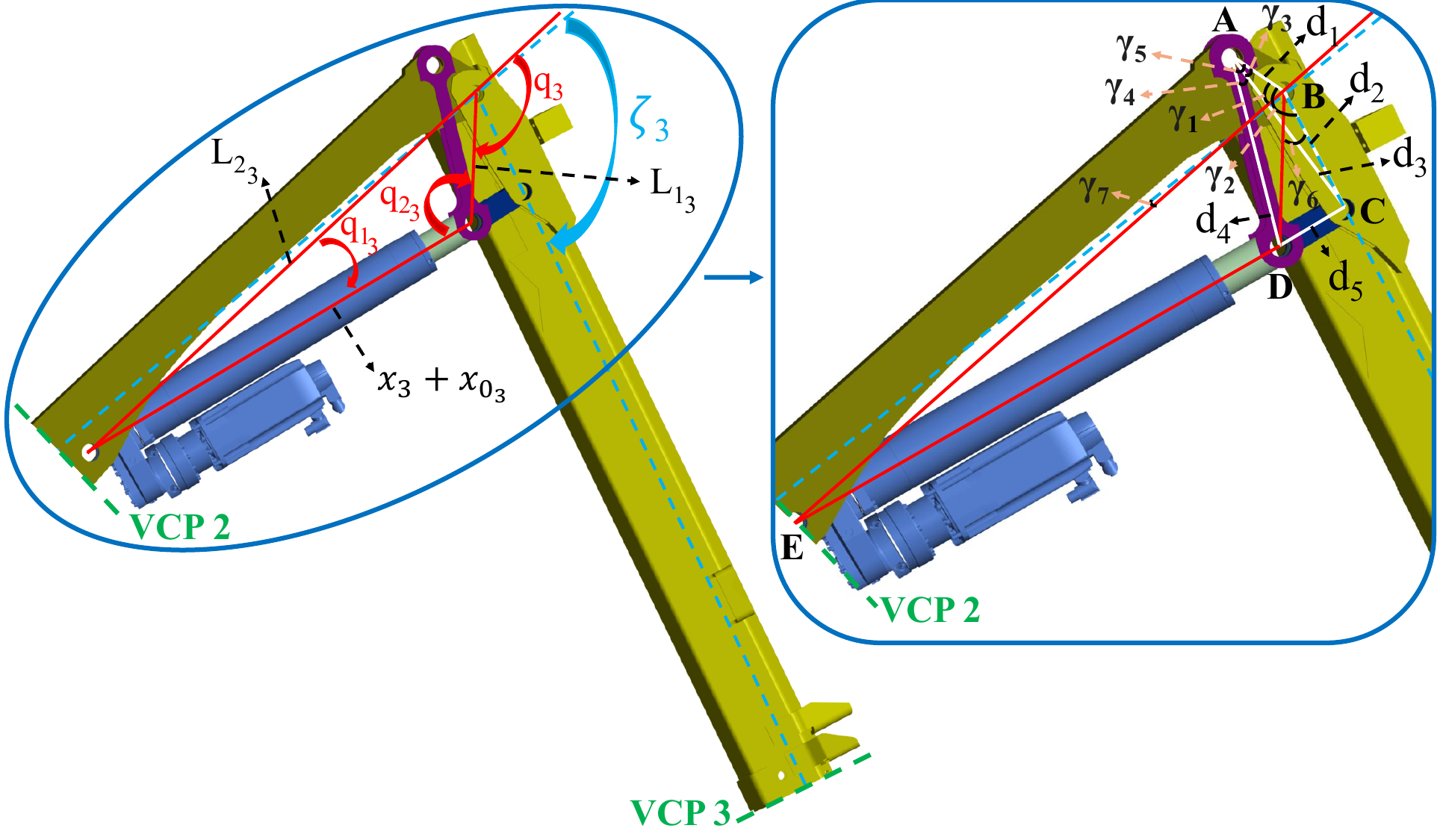}
\caption{Schematic of the tilt (4-bar mechanism).}
\label{Fig:4Bar}
\end{figure}

Referring to Fig. (\ref{Fig:4Bar}), the geometric relationships within triangle $ABC$ yield:
\begin{equation}
\left\{
\begin{aligned}
\gamma_2 & =  \gamma_1 + (\pi + \zeta_3), \\
d_3 & = \sqrt{d_1^2 + d_2^2 - 2 d_1 d_2 \cos{\gamma_2}}, \\
\gamma_3 & = \arcsin\left(\frac{d_2}{d_3} \sin{\gamma_2} \right),
\end{aligned}
\right.
\label{Eq:4_Bar_1}
\end{equation}
where $\gamma_1$ is a constant angle, $\zeta_3$ denotes the angle measured by the sensor, and $d_1$ and $d_2$ represent constant lengths. From triangle $ACD$, it follows that:
\begin{equation}
\gamma_4 = \arccos\left(\frac{d_3^2 + d_4^2 - d_5^2}{2 d_3 d_4} \right).
\label{Eq:4_Bar_2}
\end{equation}

The parameters $d_4$ and $d_5$ are constant lengths known from the mechanism's geometry. Now, by considering triangle $ABD$, we obtain:
\begin{equation}
\left\{
\begin{aligned}
\gamma_5 & = \gamma_3 + \gamma_4, \\
L_{1_3} & = \sqrt{d_1^2 + d_4^2 - 2 d_1 d_4 \cos{\gamma_5}}).
\end{aligned}
\right.
\label{Eq:4_Bar_3}
\end{equation}

Here, $L_{1_3}$ denotes a key length that plays a critical role in the following analysis. Now, by considering triangle $CDE$, we obtain:
\begin{equation}
\gamma_6 = \arccos\left(\frac{d_2^2 + L_{1_3}^2 - d_5^2}{2 d_2  L_{1_3}} \right).
\label{Eq:4_Bar_4}
\end{equation}

Given $\gamma_6$ from Equation~(\ref{Eq:4_Bar_4}) and the sensor-measured angle $\zeta_3$, it follows that:
\begin{equation}
q_3 = -\gamma_7 - \gamma_6 + \zeta_3,
\label{Eq:4_Bar_5}
\end{equation}
where $\gamma_7$ is a constant angle.

Similar to the lift motion, using the law of cosines, we have:
\begin{equation}
\left\{
\begin{aligned}
x_3 & = \sqrt{L_{1_3}^2 + L_{2_3}^2 + 2L_{1_3}L_{2_3}\cos{q_3}} - x_{0_3}, \\
q_{1_3} & = -\arccos\left(\frac{L_{1_3}^2 - (x_3+x_{0_3})^2 - L_{2_3}^2}{-2 (x_3+x_{0_3}) L_{2_3}} \right), \\
q_{2_3} & = -\arccos\left(\frac{L_{2_3}^2 - (x_3+x_{0_3})^2 - L_{1_3}^2}{-2 (x_3+x_{0_3}) L_{1_3}} \right).
\end{aligned}
\right.
\label{Eq:4_Bar_6}
\end{equation}

Now, by differentiating (\ref{Eq:4_Bar_6}), using the geometric parameters from (\ref{Eq:4_Bar_1}) to (\ref{Eq:4_Bar_5}), the variables $\dot{x}_3$, $\dot{q}_3$, $\dot{q}_{1_3}$, and $\dot{q}_{2_3}$ are calculated.

The linear/angular velocity vectors corresponding to the open chain \(1_3\) are obtained as follows:
\begin{equation}
\left\{
\begin{aligned}
\mathbf{{}^{A_3}}{\boldsymbol{V}} & = \mathbf{z_{\tau}} \dot{q}_3 + { }^{\mathbf{B_{1_3}}} \mathbf{U}_{\mathbf{A_3}}^T \mathbf{{}^{B_{1_3}}}{\boldsymbol{V}}, \\
\mathbf{{}^{T_{1_3}}}{\boldsymbol{V}} & =  { }^{\mathbf{A_3}} \mathbf{U}_{\mathbf{T_{1_3}}}^T \mathbf{{}^{A_3}}{\boldsymbol{V}},
\end{aligned}
\right.
\label{Eq:Tilt_Kinematics_Openchain1}
\end{equation}

In a similar manner, the linear/angular velocity vectors for the open chain \(2_3\) are given by:
\begin{equation}
\left\{
\begin{aligned}
\mathbf{{}^{C_3}}{\boldsymbol{V}} & = \mathbf{z_{\tau}} \dot{q}_{1_3} + { }^{\mathbf{B_{2_3}}} \mathbf{U}_{\mathbf{C_3}}^T \mathbf{{}^{B_{2_3}}}{\boldsymbol{V}}, \\
\mathbf{{}^{D_3}}{\boldsymbol{V}} & = \mathbf{x_{f}} \dot{x}_3 + { }^{\mathbf{C_3}} \mathbf{U}_{\mathbf{D_3}}^T \mathbf{{}^{C_3}}{\boldsymbol{V}}, \\
\mathbf{{}^{T_{2_3}}}{\boldsymbol{V}} & = \mathbf{z_{\tau}} \dot{q}_{2_3} + { }^{\mathbf{D_3}} \mathbf{U}_{\mathbf{T_{2_3}}}^T \mathbf{{}^{D_3}}{\boldsymbol{V}},
\end{aligned}
\right.
\label{Eq:Tilt_Kinematics_Openchain2}
\end{equation}

As the initial and terminal points of the open chains \(1_3\) and \(2_3\) coincide, the following constraint equations must be satisfied:
\begin{equation}
\left\{
\begin{aligned}
\mathbf{{}^{B_{1_3}}}{\boldsymbol{V}} & = \mathbf{{}^{B_{2_3}}}{\boldsymbol{V}} = \mathbf{{}^{B_{c_3}}}{\boldsymbol{V}}, \\
\mathbf{{}^{T_{1_3}}}{\boldsymbol{V}} & = \mathbf{{}^{T_{2_3}}}{\boldsymbol{V}} = \mathbf{{}^{T_{c_3}}}{\boldsymbol{V}},
\end{aligned}
\right.
\label{Eq:Tilt_Kinematics_Constraint}
\end{equation}

Finally, the linear/angular velocity vector at the base of the fourth object is computed as follows:
\begin{equation}
\mathbf{{}^{B_{c_4}}}{\boldsymbol{V}} = { }^{\mathbf{T_{c_3}}} \mathbf{U}_{\mathbf{B_{c_4}}}^T \mathbf{{}^{T_{c_3}}}{\boldsymbol{V}}.
\label{Eq:Tilt_to_Wrist_Kinematics}
\end{equation}

\begin{itemize}
    \item Wrist Motion:
\end{itemize}

At this stage, we proceed to derive the kinematic equations associated with the wrist mechanism. As the wrist constitutes the fourth element in the kinematic chain, all corresponding variables are denoted with the subscript 4 to maintain consistency in notation. The kinematic relationship from the wrist base frame, denoted as $\mathbf{B_{c_4}}$, to the wrist tip frame, denoted as $\mathbf{T_4}$, which corresponds to the end effector, is formulated as follows:
\begin{equation}
\left\{
\begin{aligned}
\mathbf{{}^{A_4}}{\boldsymbol{V}} & = \mathbf{x_{\tau}} \dot{\zeta}_4 + { }^{\mathbf{B_{c_4}}} \mathbf{U}_{\mathbf{A_4}}^T \mathbf{{}^{B_{c_4}}}{\boldsymbol{V}}, \\
\mathbf{{}^{C_4}}{\boldsymbol{V}} & = \mathbf{z_{\tau}} \dot{\zeta}_5 + { }^{\mathbf{A_4}} \mathbf{U}_{\mathbf{C_4}}^T \mathbf{{}^{A_4}}{\boldsymbol{V}}, \\
\mathbf{{}^{D_4}}{\boldsymbol{V}} & = \mathbf{x_{\tau}} \dot{\zeta}_6 + { }^{\mathbf{C_4}} \mathbf{U}_{\mathbf{D_4}}^T \mathbf{{}^{C_4}}{\boldsymbol{V}}, \\
\mathbf{{}^{T_4}}{\boldsymbol{V}} & = { }^{\mathbf{D_4}} \mathbf{U}_{\mathbf{T_4}}^T \mathbf{{}^{D_4}}{\boldsymbol{V}},
\end{aligned}
\right.
\label{Eq:Wrist_Kinematics}
\end{equation}

\subsubsection{HDRM Dynamic Formulation}
\label{subsubsec:dynamics of the manipulator}

With the linear/angular velocity vectors of each component obtained, we substitute them into (\ref{Eq:Net_Force}) to compute the net force/moment vectors ($\boldsymbol{F}^*$) for each individual component. According to the VDC framework, the force/moment vectors are computed in a recursive manner, propagating from the wrist toward the base of the manipulator, as follows:

\begin{itemize}
    \item Wrist Motion:
\end{itemize}

The dynamic behavior of the fourth component in the system, corresponding to the wrist, is described by the following equations:
\begin{equation}
\left\{
\begin{aligned}
\mathbf{{}^{D_4}}{\boldsymbol{F}} &= \mathbf{{}^{D_4}}{\boldsymbol{F}^*} + { }^{\mathbf{D_4}} \mathbf{U}_{\mathbf{T_4}} \mathbf{{}^{T_4}}{\boldsymbol{F}},\\
\mathbf{{}^{C_4}}{\boldsymbol{F}} &= \mathbf{{}^{C_4}}{\boldsymbol{F}^*} + { }^{\mathbf{C_4}} \mathbf{U}_{\mathbf{D_4}} \mathbf{{}^{D_4}}{\boldsymbol{F}},\\
\mathbf{{}^{A_4}}{\boldsymbol{F}} &= \mathbf{{}^{A_4}}{\boldsymbol{F}^*} + { }^{\mathbf{A_4}} \mathbf{U}_{\mathbf{C_4}} \mathbf{{}^{C_4}}{\boldsymbol{F}},\\
\mathbf{{}^{B_{c_4}}}{\boldsymbol{F}} &= \mathbf{{}^{B_{c_4}}}{\boldsymbol{F}^*} + { }^{\mathbf{B_{c_4}}} \mathbf{U}_{\mathbf{A_4}} \mathbf{{}^{A_4}}{\boldsymbol{F}},
\end{aligned}
\right.
\label{Eq:Wrist_Dynamics}
\end{equation}
where $\mathbf{{}^{T_4}}\boldsymbol{F}$ denotes the force/moment vector exerted on the end-effector by the environment, which can include contributions from the payload.

\begin{itemize}
    \item Tilt Motion (Four-Bar Mechanism):
\end{itemize}

Having obtained the force/moment vector at the base of the wrist ($\mathbf{B_{c_4}}$), we can now transfer it to the base of the tilt mechanism ($\mathbf{B_{c_3}}$) using the following equation. A comprehensive derivation of this relation is provided in \cite{petrovic2022mathematical}:
\begin{equation}
\begin{aligned}
\mathbf{{}^{B_{c_3}}}{\boldsymbol{F}} & = \mathbf{{}^{B_{1_3}}}{\boldsymbol{F}^*} + { }^{\mathbf{B_{1_3}}} \mathbf{U}_{\mathbf{A_3}} \mathbf{{}^{A_3}}{\boldsymbol{F}^*} + { }^{\mathbf{B_{2_3}}} \mathbf{U}_{\mathbf{C_3}} \mathbf{{}^{C_3}}{\boldsymbol{F}}^* \\
+& { }^{\mathbf{B_{2_3}}} \mathbf{U}_{\mathbf{C_3}} { }^{\mathbf{C_3}} \mathbf{U}_{\mathbf{D_3}} \mathbf{{}^{D_3}}{\boldsymbol{F}^*} + { }^{\mathbf{B_{1_3}}} \mathbf{U}_{\mathbf{A_3}} { }^{\mathbf{A_3}} \mathbf{U}_{\mathbf{B_{c_4}}} \mathbf{{}^{B_{c_4}}}{\boldsymbol{F}}.
\end{aligned}
\label{Eq:Wrist_to_Tilt_Dynamics}
\end{equation}

Furthermore, the force exerted by the linear actuator (EMLA) in the tilt mechanism is computed as follows:
\begin{equation}
\begin{aligned}
f_3 & = \mathbf{x_f^T} \mathbf{{}^{D_3}}{\boldsymbol{F}^*} - \frac{\mathbf{z_{\tau}^T} \left(\mathbf{{}^{A_3}}{\boldsymbol{F}^*} + { }^{\mathbf{A_3}} \mathbf{U}_{\mathbf{B_{c_4}}} \mathbf{{}^{B_{c_4}}}{\boldsymbol{F}}  \right)}{L_{1_3} \sin{q_{2_3}}}\\
-& \frac{\mathbf{z_{\tau}^T} \left( \mathbf{{}^{C_3}}{\boldsymbol{F}^*} + \mathbf{{}^{D_3}}{\boldsymbol{F}^*} \right) + \mathbf{y_f^T} \left( \mathbf{{}^{D_3}}{\boldsymbol{F}^*} \right) \left( x_3 + x_{0_3}- l_{c_3} \right) }{\left( x_3 + x_{0_3} \right) \tan q_{2_3}}.
\end{aligned}
\label{Eq:Actuator_Tilt_Dynamics}
\end{equation}

\begin{itemize}
    \item Lift Motion (Three-Bar Mechanism):
\end{itemize}

Similarly, the force/moment vector calculated at the base of the tilt joint ($\mathbf{B_{c_3}}$) is subsequently mapped to the base of the lift mechanism ($\mathbf{B_{c_2}}$) using the transformation equation presented below \cite{petrovic2022mathematical}.
\begin{equation}
\begin{aligned}
&\mathbf{{}^{B_{c_2}}}{\boldsymbol{F}} = \mathbf{{}^{B_{1_2}}}{\boldsymbol{F}^*} + { }^{\mathbf{B_{1_2}}} \mathbf{U}_{\mathbf{A_2}} \mathbf{{}^{A_2}}{\boldsymbol{F}^*} + { }^{\mathbf{B_{2_2}}} \mathbf{U}_{\mathbf{C_2}} \mathbf{{}^{C_2}}{\boldsymbol{F}}^*\\
+& { }^{\mathbf{B_{2_2}}} \mathbf{U}_{\mathbf{C_2}} { }^{\mathbf{C_2}} \mathbf{U}_{\mathbf{D_2}} \mathbf{{}^{D_2}}{\boldsymbol{F}^*} + { }^{\mathbf{B_{1_2}}} \mathbf{U}_{\mathbf{A_2}} { }^{\mathbf{A_2}} \mathbf{U}_{\mathbf{B_{c_3}}} \mathbf{{}^{B_{c_3}}}{\boldsymbol{F}}.
\end{aligned}
\label{Eq:Tilt_to_Lift_Dynamics}
\end{equation}

The force generated by the linear actuator (EMLA) within the lift mechanism is determined using the following expression:
\begin{equation}
\begin{aligned}
f_2 & = \mathbf{x_f^T} \; \mathbf{{}^{D_2}}{\boldsymbol{F}^*} - \frac{\mathbf{z_{\tau}^T} \left(\mathbf{{}^{A_2}}{\boldsymbol{F}^*} + { }^{\mathbf{A_2}} \mathbf{U}_{\mathbf{B_{c_3}}} \mathbf{{}^{B_{c_3}}}{\boldsymbol{F}}  \right)}{L_{1_2} \sin{q_{2_2}}}\\
-& \frac{\mathbf{z_{\tau}^T} \left( \mathbf{{}^{C_2}}{\boldsymbol{F}^*} + \mathbf{{}^{D_2}}{\boldsymbol{F}^*} \right) + \mathbf{y_f^T} \left( \mathbf{{}^{D_2}}{\boldsymbol{F}^*} \right) \left( x_2 + x_{0_2}- l_{c_2} \right) }{\left( x_2 + x_{0_2} \right) \tan q_{2_2}}.
\end{aligned}
\label{Eq:Actuator_Lift_Dynamics}
\end{equation}

\begin{itemize}
    \item Base Motion:
\end{itemize}

Given the force/moment vector $\mathbf{{}^{B_{c_2}}{\boldsymbol{F}}}$ from object 2, the corresponding force/moment vector at the base can be obtained using the following relation:
\begin{equation}
\mathbf{{}^{T_1}}{\boldsymbol{F}} = \mathbf{{}^{T_1}}{\boldsymbol{F}^*} + { }^{\mathbf{T_1}} \mathbf{U}_{\mathbf{B_{c_2}}} \mathbf{{}^{B_{c_2}}}{\boldsymbol{F}}.
\label{Eq:Base_Dynamics1}
\end{equation}

Finally, the force generated by the EMLA contributing to the base motion is given by:
\begin{equation}
f_1 = \frac{1}{r_B}y_{\tau}^T \left(\mathbf{{}^{T_1}}{\boldsymbol{F}} \right),
\label{Eq:Base_Dynamics2}
\end{equation}
where $r_B$ denotes the radius of the base mechanism.

\subsection{Surrogate-Enhanced Hybrid Model of the EMLA}
\label{subsec:hybdrid_emla_model}
\begin{figure}[ht]
\centering
\includegraphics[width=0.91\linewidth]{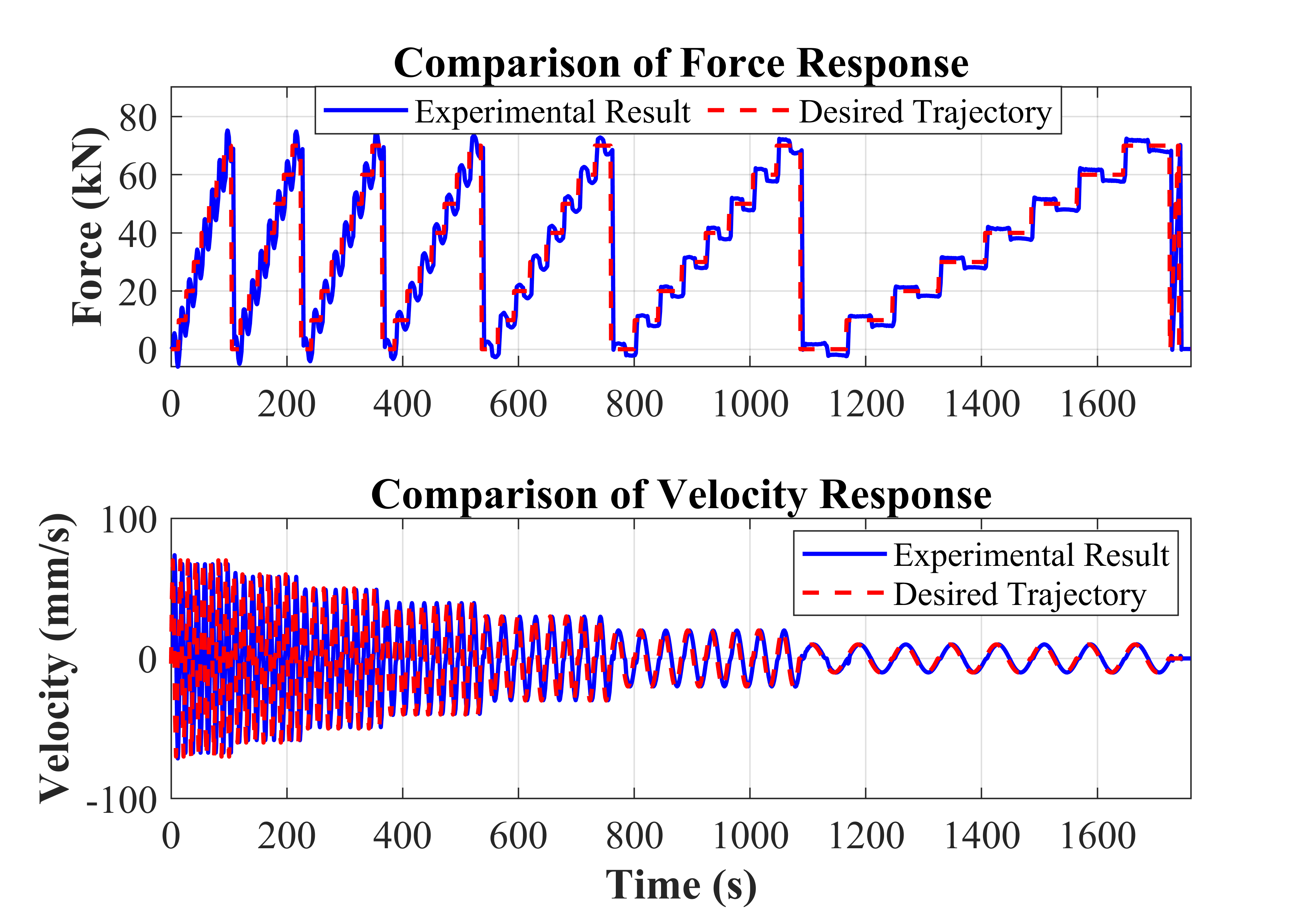}
\caption{Comparison of the experimental force and velocity responses with the desired trajectory.}
\label{fig:exp_vs_des}
\end{figure}

To complement the physics-based model described in the previous section, this study leverages experimental data to characterize the dynamic behavior of the EMLA under realistic operating conditions. A long-duration test campaign was conducted, during which the EMLA was driven across a wide range of loads (0--70~kN) and speeds (0--70~mm/s), as illustrated in Fig.~\ref{fig:exp_vs_des}. 
The recorded dataset includes synchronized measurements of actuator output force ($F_{\text{ext}}$), linear velocity ($\dot{x}$), and motor-side electrical signals $(v_d, v_q, i_d, i_q)$, from which the instantaneous mechanical output power and electrical input power were evaluated at each operating point. The EMLA efficiency $\eta_{\text{EMLA}}(\dot{\theta}_m, \tau_e)$ was then computed as follows:
\begin{equation}
    \eta_{\text{EMLA}}(\dot{\theta}_m, \tau_e) = \frac{\dot{x}^{\top} F_{\text{ext}}}{\dot{\theta}_m^{\top} \tau_e + P_{\text{loss}} (v_d, v_q, i_d, i_q)},
\label{eq:emla_eff}
\end{equation}
where $P_{\text{loss}}$ represents cumulative motor losses. As the raw efficiency values are computationally expensive to evaluate in real-time applications, a data-driven surrogate model was developed. Deep neural network (DNN) was trained on the experimental dataset to approximate the actuator behavior. The DNN receives PMSM torque ($\tau_e$) and angular velocity ($\dot{\theta}_m$) as inputs and predicts the corresponding actuator force ($\hat{F}_{\text{ext}}$) and linear velocity ($\hat{\dot{x}}$), and the inferred outputs enable reconstruction of the mechanical power ($\hat{P}_{\mathrm{out}}$) and estimation of the predicted efficiency as follows:
\begin{equation}
\begin{aligned}
&
\begin{bmatrix}
\hat{F}_{\text{ext}} \\
\hat{\dot{x}}
\end{bmatrix}
= \mathcal{F}_{\text{DNN}} \left(
\begin{bmatrix}
\tau_e \\
\dot{\theta}_m
\end{bmatrix}
\right), \\[6pt]
&\hat{P}_{\mathrm{out}} = \hat{\dot{x}}^{\top} \cdot \hat{F}_{\text{ext}}, \\[4pt]
&{P}_{\mathrm{in}} = \dot{\theta}_m^{\top} \cdot \tau_e, \\[6pt]
&P_{\mathrm{loss}} = P_{\mathrm{sw}} + P_{\mathrm{cu}} + P_{\mathrm{core}} + P_{\mathrm{mech}}, \\[4pt]
&\hat{\eta}_{\text{EMLA}} =
\begin{cases}
\displaystyle \frac{\hat{P}_{\mathrm{out}}}{{P}_{\mathrm{in}} + P_{\mathrm{loss}}}, & \text{if } \tau_e > 0 \text{ and } \dot{\theta}_m > 0 \quad \text{(Q-I)} \\[8pt]
\displaystyle \frac{{P}_{\mathrm{in}} - P_{\mathrm{loss}}}{\hat{P}_{\mathrm{out}}}, & \text{if } \tau_e > 0 \text{ and } \dot{\theta}_m < 0 \quad \text{(Q-IV)} \\[8pt]
\text{NaN}, & \text{otherwise}
\end{cases}
\end{aligned}
\label{eq:dnn_prediction_efficiency}
\end{equation}
with $\mathcal{F}_{\text{DNN}}(\boldsymbol{x})$ denoting the DNN function, and \(P_{\mathrm{sw}}\), \(P_{\mathrm{cu}}\), \(P_{\mathrm{core}}\), and \(P_{\mathrm{mech}}\) denoting the switching losses, copper‐conduction losses, core losses including hysteresis and eddy currents, and mechanical losses including windage and bearing friction of the PMSM~\cite{9123428,8401696}. The final network architecture consisted of a feedforward DNN with five hidden layers with sizes $[64, 48, 32, 16, 8]$, trained using the Levenberg--Marquardt (\texttt{trainlm}) algorithm. A \texttt{mapminmax} normalization strategy was applied to both inputs and targets, and the model was trained. The dataset was split into training, validation, and test sets using a 70:20:10 ratio. 


The trained model achieved high predictive accuracy, with correlation coefficients exceeding \(R > 0.995\) across the training, validation, and test sets, and mean squared errors of 0.00013, 0.00476, and 0.00219, respectively, indicating low overfitting and strong generalization over the operating range.

Rather than replacing the physics-based model outright, the DNN is trained to estimate the residual dynamics not captured by the analytical formulation, such as unmodeled losses and nonlinearities~\cite{tahkola2020surrogate}, thereby refining the predicted force, velocity, and efficiency. The resulting hybrid model combines analytical predictions with data-driven corrections using formulation in \eqref{eq:residual_blending_hat}, where each final output is computed as the sum of the physics-based estimate and a scaled neural residual. The coefficient $\alpha \in [0,1]$ governs the influence of the correction, and this structure ensures the hybrid model remains physically interpretable and dynamically consistent, while achieving high empirical accuracy suitable for integration into real-time control and an energy-aware simulation:
\begin{equation}
\left\{
\begin{aligned}
F_{\mathrm{hyb}} &= F_{\mathrm{ext}}(\dot{\theta}_m, \tau_e) + \alpha  \left( \hat{F}_{\mathrm{ext}} - F_{\mathrm{ext}}(\dot{\theta}_m, \tau_e) \right), \\
\dot{x}_{\mathrm{hyb}} &= \dot{x}(\dot{\theta}_m, \tau_e) + \alpha  \left( \hat{\dot{x}} - \dot{x}(\dot{\theta}_m, \tau_e) \right), \\
\eta_{\mathrm{hyb}} &= \eta_{\mathrm{EMLA}}(\dot{\theta}_m, \tau_e) \\
&\qquad + \alpha \left(\hat{\eta}_{\mathrm{EMLA}} - \eta_{\mathrm{EMLA}}(\dot{\theta}_m, \tau_e) \right).
\end{aligned}
\right.
\label{eq:residual_blending_hat}
\end{equation}

\begin{figure}[ht]
\centering
\includegraphics[width=2.4in]{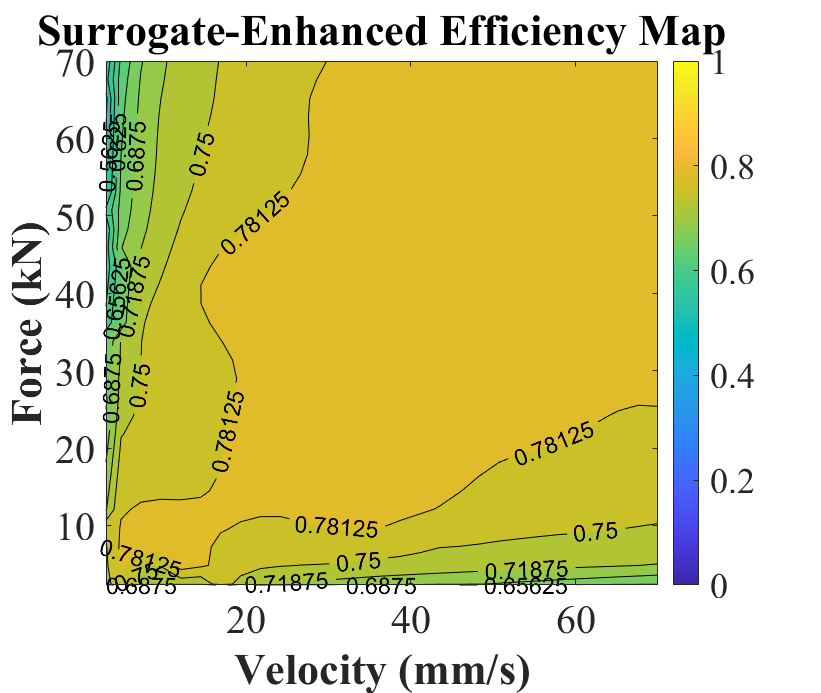}
\caption{Surrogate-enhanced efficiency map of the EMLA, combining physics-based modeling with DNN predictions. The map illustrates actuator efficiency across varying load forces and linear velocities, capturing both model-driven dynamics and experimentally-informed nonlinearities.}
\label{fig:emla_hybrid_eff_map}
\end{figure}

Finally, Fig.~\ref{fig:emla_hybrid_eff_map} illustrates the resulting surrogate-enhanced efficiency map that reflects the actuator's performance across the full range of operating conditions, derived from the hybrid formulation in \eqref{eq:residual_blending_hat}. The resulting hybrid EMLA model is embedded into the VDC framework and is used to simulate the full 6-DoF all-electric robotic manipulator. 
%
%
%
\section{Control}
\label{sec:control}
Building upon the dynamic models presented earlier, this section derives the linear/angular velocity vectors, as well as the corresponding force/moment vectors, required for control within the VDC framework. 

First, consider the relationship between the task space and joint space, expressed using the Denavit–Hartenberg (DH) convention, as follows:

\begin{equation}
\mathbf{\dot{\Pi}} = \mathbf{J} \mathbf{\dot{\Theta}},
\label{Eq:Inverse Kinematics}
\end{equation}
where $\dot{\boldsymbol{\Pi}} \in \mathbb{R}^6$ denotes the linear/angular velocity vector of the end-effector in Cartesian space, $\mathbf{J} \in \mathbb{R}^{6 \times 6}$ is the Jacobian matrix, and $\boldsymbol{\Theta} = [\zeta_1, \zeta_2, \zeta_3, \zeta_4, \zeta_5, \zeta_6]^T$ represents the joint angle vector. Given a desired end-effector velocity $\dot{\boldsymbol{\Pi}}_d$ in Cartesian space, the required joint velocities can be computed using the closed-loop inverse kinematics (CLIK) method, as follows \cite{sciavicco2001modelling}:
\begin{equation}
\mathbf{\dot{\Theta}_r} = \mathbf{J}^{-1} (\mathbf{\dot{\Pi}_d} + \mathbf{\Lambda} (\mathbf{\Pi_d} -  \mathbf{\Pi}))  ,
\label{Eq:Inverse Kinematics2}
\end{equation}
where $\mathbf{\Lambda} \in \mathbb{R}^{6\times6}$ is a constant positive-definite matrix.

\subsection{VDC Controller Design for the Manipulator (High-Level)}
\label{subsub:high_level}

\subsubsection{Required Linear/Angular Velocity Vectors}

Now that the required joint velocities have been obtained from (\ref{Eq:Inverse Kinematics2}), the corresponding linear/angular velocity vectors can be derived by substituting the required velocity vectors into the equations presented in the modeling chapter (see Section~\ref{subsubsec:kinematics of the manipulator}).
\begin{itemize}
    \item Base Motion:
\end{itemize}

\begin{equation}
\mathbf{{}^{T_1}}{\boldsymbol{V}_r} = \mathbf{y_{\tau}} \dot{\zeta}_{1_r} + { }^{\mathbf{B_1}} \mathbf{U}_{\mathbf{T_1}}^T \mathbf{{}^{B_1}}{\boldsymbol{V}_r},
\label{Eq:Base_Required_Kinematics}
\end{equation}

\begin{equation}
\mathbf{{}^{B_{c_2}}}{\boldsymbol{V}_r} = { }^{\mathbf{T_1}} \mathbf{U}_{\mathbf{B_{c_2}}}^T \mathbf{{}^{T_1}}{\boldsymbol{V}_r}.
\label{Eq:Base_to_Lift_Required_Kinematics}
\end{equation}
where $\dot{\zeta}_{1_r}$ is derived from (\ref{Eq:Inverse Kinematics2}).

\begin{itemize}
    \item Lift Motion:
\end{itemize}

\begin{equation}
\left\{
\begin{aligned}
\mathbf{{}^{A_2}}{\boldsymbol{V}_r} & = \mathbf{z_{\tau}} \dot{q}_{2_r} + { }^{\mathbf{B_{1_2}}} \mathbf{U}_{\mathbf{A_2}}^T \mathbf{{}^{B_{1_2}}}{\boldsymbol{V}_r}, \\
\mathbf{{}^{T_{1_2}}}{\boldsymbol{V}_r} & =  { }^{\mathbf{A_2}} \mathbf{U}_{\mathbf{T_{1_2}}}^T \mathbf{{}^{A_2}}{\boldsymbol{V}_r},
\end{aligned}
\right.
\label{Eq:Lift_Required_Kinematics_Openchain1}
\end{equation}

\begin{equation}
\left\{
\begin{aligned}
\mathbf{{}^{C_2}}{\boldsymbol{V}_r} & = \mathbf{z_{\tau}} \dot{q}_{{1_2}_r} + { }^{\mathbf{B_{2_2}}} \mathbf{U}_{\mathbf{C_2}}^T \mathbf{{}^{B_{2_2}}}{\boldsymbol{V}_r}, \\
\mathbf{{}^{D_2}}{\boldsymbol{V}_r} & = \mathbf{x_{f}} \dot{x}_{2_r} + { }^{\mathbf{C_2}} \mathbf{U}_{\mathbf{D_2}}^T \mathbf{{}^{C_2}}{\boldsymbol{V}_r}, \\
\mathbf{{}^{T_{2_2}}}{\boldsymbol{V}_r} & = \mathbf{z_{\tau}} \dot{q}_{{2_2}_r} + { }^{\mathbf{D_2}} \mathbf{U}_{\mathbf{T_{2_2}}}^T \mathbf{{}^{D_2}}{\boldsymbol{V}_r},
\end{aligned}
\right.
\label{Eq:Lift_Required_Kinematics_Openchain2}
\end{equation}
where $\dot{x}_{2_r}$, $\dot{q}_{2_r}$ $\dot{q}_{{1_2}_r}$, and $\dot{q}_{{2_2}_r}$ are readily obtained by substituting $\dot{\zeta}_{2_r}$ from (\ref{Eq:Inverse Kinematics2}) into (\ref{Eq:3_Bar_4}). Then, the required linear/angular velocity vector at the base of the next object (tilt) is obtained as follows:
\begin{equation}
\mathbf{{}^{B_{c_3}}}{\boldsymbol{V}_r} = { }^{\mathbf{T_{c_2}}} \mathbf{U}_{\mathbf{B_{c_3}}}^T \mathbf{{}^{T_{c_2}}}{\boldsymbol{V}_r}.
\label{Eq:Lift_to_Tilt_Required_Kinematics}
\end{equation}

\begin{itemize}
    \item Tilt Motion:
\end{itemize}

\begin{equation}
\left\{
\begin{aligned}
\mathbf{{}^{A_3}}{\boldsymbol{V}_r} & = \mathbf{z_{\tau}} \dot{q}_{3_r} + { }^{\mathbf{B_{1_3}}} \mathbf{U}_{\mathbf{A_3}}^T \mathbf{{}^{B_{1_3}}}{\boldsymbol{V}_r}, \\
\mathbf{{}^{T_{1_3}}}{\boldsymbol{V}_r} & =  { }^{\mathbf{A_3}} \mathbf{U}_{\mathbf{T_{1_3}}}^T \mathbf{{}^{A_3}}{\boldsymbol{V}_r},
\end{aligned}
\right.
\label{Eq:Tilt_Required_Kinematics_Openchain1}
\end{equation}

\begin{equation}
\left\{
\begin{aligned}
\mathbf{{}^{C_3}}{\boldsymbol{V}_r} & = \mathbf{z_{\tau}} \dot{q}_{{1_3}_r} + { }^{\mathbf{B_{2_3}}} \mathbf{U}_{\mathbf{C_3}}^T \mathbf{{}^{B_{2_3}}}{\boldsymbol{V}_r}, \\
\mathbf{{}^{D_3}}{\boldsymbol{V}_r} & = \mathbf{x_{f}} \dot{x}_{3_r} + { }^{\mathbf{C_3}} \mathbf{U}_{\mathbf{D_3}}^T \mathbf{{}^{C_3}}{\boldsymbol{V}_r}, \\
\mathbf{{}^{T_{2_3}}}{\boldsymbol{V}_r} & = \mathbf{z_{\tau}} \dot{q}_{{2_3}_r} + { }^{\mathbf{D_3}} \mathbf{U}_{\mathbf{T_{2_3}}}^T \mathbf{{}^{D_3}}{\boldsymbol{V}_r},
\end{aligned}
\right.
\label{Eq:Tilt_Required_Kinematics_Openchain2}
\end{equation}

The quantities $\dot{x}_{3_r}$, $\dot{q}_{3_r}$, $\dot{q}_{{1_3}_r}$, and $\dot{q}_{{2_3}_r}$ can be determined by substituting $\dot{\zeta}_{3_r}$, obtained from (\ref{Eq:Inverse Kinematics2}), into the set of Equations (\ref{Eq:4_Bar_1}--\ref{Eq:4_Bar_6}). Subsequently, the required linear/angular velocity vector at the base of the following component (i.e., the wrist) is computed as:
\begin{equation}
\mathbf{{}^{B_{c_4}}}{\boldsymbol{V}_r} = { }^{\mathbf{T_{c_3}}} \mathbf{U}_{\mathbf{B_{c_4}}}^T \mathbf{{}^{T_{c_3}}}{\boldsymbol{V}_r}.
\label{Eq:Tilt_to_Wrist_Inverse_Kinematics}
\end{equation}

\begin{itemize}
    \item Wrist Motion:
\end{itemize}

\begin{equation}
\left\{
\begin{aligned}
\mathbf{{}^{A_4}}{\boldsymbol{V}_r} & = \mathbf{x_{\tau}} \dot{\zeta}_{4_r} + { }^{\mathbf{B_{c_4}}} \mathbf{U}_{\mathbf{A_4}}^T \mathbf{{}^{B_{c_4}}}{\boldsymbol{V}_r}, \\
\mathbf{{}^{C_4}}{\boldsymbol{V}_r} & = \mathbf{z_{\tau}} \dot{\zeta}_{5_r} + { }^{\mathbf{A_4}} \mathbf{U}_{\mathbf{C_4}}^T \mathbf{{}^{A_4}}{\boldsymbol{V}_r}, \\
\mathbf{{}^{D_4}}{\boldsymbol{V}_r} & = \mathbf{x_{\tau}} \dot{\zeta}_{6_r} + { }^{\mathbf{C_4}} \mathbf{U}_{\mathbf{D_4}}^T \mathbf{{}^{C_4}}{\boldsymbol{V}_r}, \\
\mathbf{{}^{T_4}}{\boldsymbol{V}_r} & = { }^{\mathbf{D_4}} \mathbf{U}_{\mathbf{T_4}}^T \mathbf{{}^{D_4}}{\boldsymbol{V}_r},
\end{aligned}
\right.
\label{Eq:Wrist_Kinematics_c}
\end{equation}
where $\dot{\zeta}_{4_r}$, $\dot{\zeta}_{5_r}$, and $\dot{\zeta}_{6_r}$ denote the required joint angular velocities associated with the wrist, and their values are directly determined from (\ref{Eq:Inverse Kinematics2}).

\subsubsection{Required Force/Moment Vectors}

Having derived the required linear/angular velocity vectors for each component, the corresponding force/moment vectors ($\boldsymbol{F}^*_r$) for all components can now be computed using (\ref{Eq:Required_Force}). Accordingly, following the VDC methodology, the required force/moment vectors for each component are derived by substituting the corresponding required terms into the dynamic equations introduced in the modeling section (see Section~\ref{subsubsec:dynamics of the manipulator}).

\begin{itemize}
    \item Wrist Motion:
\end{itemize}

For the wrist object, the required force/moment vectors are governed by the following equations:
\begin{equation}
\left\{
\begin{aligned}
\mathbf{{}^{D_4}}{\boldsymbol{F}_r} &= \mathbf{{}^{D_4}}{\boldsymbol{F}_r^*} + { }^{\mathbf{D_4}} \mathbf{U}_{\mathbf{T_4}} \mathbf{{}^{T_4}}{\boldsymbol{F}_r},\\
\mathbf{{}^{C_4}}{\boldsymbol{F}_r} &= \mathbf{{}^{C_4}}{\boldsymbol{F}_r^*} + { }^{\mathbf{C_4}} \mathbf{U}_{\mathbf{D_4}} \mathbf{{}^{D_4}}{\boldsymbol{F}_r},\\
\mathbf{{}^{A_4}}{\boldsymbol{F}_r} &= \mathbf{{}^{A_4}}{\boldsymbol{F}_r^*} + { }^{\mathbf{A_4}} \mathbf{U}_{\mathbf{C_4}} \mathbf{{}^{C_4}}{\boldsymbol{F}_r},\\
\mathbf{{}^{B_{c_4}}}{\boldsymbol{F}_r} &= \mathbf{{}^{B_{c_4}}}{\boldsymbol{F}_r^*} + { }^{\mathbf{B_{c_4}}} \mathbf{U}_{\mathbf{A_4}} \mathbf{{}^{A_4}}{\boldsymbol{F}_r},
\end{aligned}
\right.
\label{Eq:Wrist_Required_Dynamics}
\end{equation}

\begin{itemize}
    \item Tilt Motion:
\end{itemize}

Now, by substituting the required force/moment vectors into (\ref{Eq:Wrist_to_Tilt_Dynamics}) and (\ref{Eq:Actuator_Tilt_Dynamics}), the following expressions are obtained:

\begin{equation}
\begin{aligned}
\mathbf{{}^{B_{c_3}}}{\boldsymbol{F}_r} & = \mathbf{{}^{B_{1_3}}}{\boldsymbol{F}_r^*} + { }^{\mathbf{B_{1_3}}} \mathbf{U}_{\mathbf{A_3}} \mathbf{{}^{A_3}}{\boldsymbol{F}_r^*} + { }^{\mathbf{B_{2_3}}} \mathbf{U}_{\mathbf{C_3}} \mathbf{{}^{C_3}}{\boldsymbol{F}}_r^* \\
+& { }^{\mathbf{B_{2_3}}} \mathbf{U}_{\mathbf{C_3}} { }^{\mathbf{C_3}} \mathbf{U}_{\mathbf{D_3}} \mathbf{{}^{D_3}}{\boldsymbol{F}_r^*} + { }^{\mathbf{B_{1_3}}} \mathbf{U}_{\mathbf{A_3}} { }^{\mathbf{A_3}} \mathbf{U}_{\mathbf{B_{c_4}}} \mathbf{{}^{B_{c_4}}}{\boldsymbol{F}_r},
\end{aligned}
\label{Eq:Wrist_to_Tilt_Required_Dynamics}
\end{equation}

\begin{equation}
\begin{aligned}
f_{r_3} & = \mathbf{x_f^T} \mathbf{{}^{D_3}}{\boldsymbol{F}_r^*} - \frac{\mathbf{z_{\tau}^T} \left(\mathbf{{}^{A_3}}{\boldsymbol{F}_r^*} + { }^{\mathbf{A_3}} \mathbf{U}_{\mathbf{B_{c_4}}} \mathbf{{}^{B_{c_4}}}{\boldsymbol{F}_r}  \right)}{L_{1_3} \sin{q_{2_3}}}\\
-& \frac{\mathbf{z_{\tau}^T} \left( \mathbf{{}^{C_3}}{\boldsymbol{F}_r^*} + \mathbf{{}^{D_3}}{\boldsymbol{F}_r^*} \right) + \mathbf{y_f^T} \left( \mathbf{{}^{D_3}}{\boldsymbol{F}_r^*} \right) \left( x_3 + x_{0_3}- l_{c_3} \right) }{\left( x_3 + x_{0_3} \right) \tan q_{2_3}}.
\end{aligned}
\label{Eq:Actuator_Tilt_Required_Dynamics}
\end{equation}

\begin{itemize}
    \item Lift Motion:
\end{itemize}

For the lift motion, substituting the required force/moment vectors into (\ref{Eq:Tilt_to_Lift_Dynamics}) and (\ref{Eq:Actuator_Lift_Dynamics}) yields the following result:

\begin{equation}
\begin{aligned}
\mathbf{{}^{B_{c_2}}}{\boldsymbol{F}_r} & = \mathbf{{}^{B_{1_2}}}{\boldsymbol{F}_r^*} + { }^{\mathbf{B_{1_2}}} \mathbf{U}_{\mathbf{A_2}} \mathbf{{}^{A_2}}{\boldsymbol{F}_r^*} + { }^{\mathbf{B_{2_2}}} \mathbf{U}_{\mathbf{C_2}} \mathbf{{}^{C_2}}{\boldsymbol{F}}_r^* \\
+& { }^{\mathbf{B_{2_2}}} \mathbf{U}_{\mathbf{C_2}} { }^{\mathbf{C_2}} \mathbf{U}_{\mathbf{D_2}} \mathbf{{}^{D_2}}{\boldsymbol{F}_r^*} + { }^{\mathbf{B_{1_2}}} \mathbf{U}_{\mathbf{A_2}} { }^{\mathbf{A_2}} \mathbf{U}_{\mathbf{B_3}} \mathbf{{}^{B_3}}{\boldsymbol{F}_r},
\end{aligned}
\label{Eq:Tilt_to_Lift_Required_Dynamics}
\end{equation}

\begin{equation}
\begin{aligned}
f_{r_2} & = \mathbf{x_f^T} \mathbf{{}^{D_2}}{\boldsymbol{F}_r^*} - \frac{\mathbf{z_{\tau}^T} \left(\mathbf{{}^{A_2}}{\boldsymbol{F}_r^*} + { }^{\mathbf{A_2}} \mathbf{U}_{\mathbf{B_{c_3}}} \mathbf{{}^{B_{c_3}}}{\boldsymbol{F}_r}  \right)}{L_{1_2} \sin{q_{2_2}}}\\
-& \frac{\mathbf{z_{\tau}^T} \left( \mathbf{{}^{C_2}}{\boldsymbol{F}_r^*} + \mathbf{{}^{D_2}}{\boldsymbol{F}_r^*} \right) + \mathbf{y_f^T} \left( \mathbf{{}^{D_2}}{\boldsymbol{F}_r^*} \right) \left( x_2 + x_{0_2}- l_{c_2} \right) }{\left( x_2 + x_{0_2} \right) \tan q_{2_2}}.
\end{aligned}
\label{Eq:Actuator_Lift_Required_Dynamics}
\end{equation}

\begin{itemize}
    \item Base Motion:
\end{itemize}

Finally, the required force/moment vector at the base of the manipulator, along with the corresponding required force at the base EMLA, are given by:
\begin{equation}
\mathbf{{}^{T_1}}{\boldsymbol{F}_r} = \mathbf{{}^{T_1}}{\boldsymbol{F}_r^*} + { }^{\mathbf{T_1}} \mathbf{U}_{\mathbf{B_{c_2}}} \mathbf{{}^{B_{c_2}}}{\boldsymbol{F}_r},
\label{Eq:Base_Required_Dynamics1}
\end{equation}
\begin{equation}
f_{r_1} = \frac{1}{r_B}y_{\tau}^T \left(\mathbf{{}^{T_1}}{\boldsymbol{F}}_r \right).
\label{Eq:Actuator_Base_Required_Dynamics2}
\end{equation}

\subsection{Low-Level control}
\label{subsub:low_level}
Given the required forces and velocities, the EMLAs must now produce the corresponding motions and forces. By integrating the surrogate-enhanced EMLA model introduced in Section~\ref{subsec:hybdrid_emla_model} into the actuator control framework, the following low-level controller is designed for each EMLA to ensure the stability of the closed-loop system:
\begin{equation}
\left\{
\begin{aligned}
v_{d_r} &= R_s i_d + L_d \frac{di_{d_r}}{dt} - p  \dot{\theta}_m \lambda_q  + K_i(i_{d_r}-i_d),\\
v_{q_r} &= R_s i_q + L_q \frac{di_{q_r}}{dt} + p  \dot{\theta}_m \lambda_d \\
&+ K_f(F_{\mathrm{hyb}_r} - F_{\mathrm{hyb}}) + K_v(\dot{x}_r -\dot{x}).
\end{aligned}
\label{eq:voltages_Control}
\right.
\end{equation}

Here, $K_i$, $K_f$, and $K_v$ are positive control gains, while $i_d$ and $i_q$ denote the currents of the PMSM. The external force $F_{\mathrm{hyb}}$ is obtained from the surrogate-enhanced EMLA model described in (\ref{eq:residual_blending_hat}), while $F_{\mathrm{hyb}_r}$ represents the required force for each EMLA, as derived in~\eqref{Eq:Actuator_Tilt_Required_Dynamics}, ~\eqref{Eq:Actuator_Lift_Required_Dynamics}, and~\eqref{Eq:Actuator_Base_Required_Dynamics2}. The term $\dot{x}$ denotes the actual linear velocity, and $\dot{x}_r$ is the corresponding required linear velocity for each EMLA. It is worth noting that the model-based terms in~\eqref{eq:voltages_Control} serve to compensate for the dynamics of the actuator. Meanwhile, the error feedback terms are designed to regulate the tracking performance in current, force, and velocity simultaneously, ensuring stable closed-loop behavior.

 The required current $i_{d_r}$ is calculated as follows:
\begin{equation}
\frac{d}{dt}i_{d_r} = \frac{d}{dt}i_{d_d} + \lambda_i \left( i_{d_d} - i_d \right),
\label{eq:current_d_required}
\end{equation}
where $i_{d_d}$ is the desired current, and $\lambda_i$ is a positive control gain. It is worth mentioning that setting $\frac{d}{dt}i_{d_d} = i_{d_d}=0$ corresponds to the conventional strategy for efficient PMSM operation~\cite{9855395,9363639}.

The required current \( i_{q_r} \) is computed from~\eqref{eq:residual_blending_hat} when employing the EMLA dynamics, as outlined in Section~\ref{subsec:emla_dynamic}, as follows:
\begin{equation}
\begin{aligned}
&F_{\mathrm{hyb}_r} = (1-\alpha) \left( \frac{2 \pi N_{\mathrm{gear}}}{\rho} \right) \left[ \frac{3}{2} p \left( \lambda_d i_{q_r} - \lambda_q i_{d_r} \right) \right. \\
& \left.-\left( J_m + \frac{M_{\mathrm{act}}}{N_{\mathrm{gear}}+\eta_f} \right) \ddot{\theta}_m - C_m \dot{\theta}_m - \tau_C \right] + \alpha  \hat{F}_{\mathrm{ext}}.
\end{aligned}
\label{eq:current_q_required}
\end{equation}

\subsection{Stability Proof}
\label{subsec:stability proof}

As mentioned earlier, the high-level controller computes the required forces and velocities for the rigid body subsystem (Section~\ref{subsub:high_level}), while the low-level controller generates these commands at the actuator level (Section~\ref{subsub:low_level}). In this section, VPFs are used to link the rigid body and actuator subsystems to establish the overall stability of the entire system.

\begin{thm}
    Consider the manipulator illustrated in Fig.~\ref{Fig:SimulationPlatform}, which is decomposed into four subsystems. Each subsystem, governed by the high-level controller~(\ref{Eq:Required_Force}), the adaptation laws~(\ref{L adapt}), and the low-level controller~\eqref{eq:voltages_Control}, is virtually stable in the sense of Definition~\ref{Def_VS}.
\label{thm:virtual_stability}
\end{thm}

\textit{Proof}: Define the non-negative accompanying function for the $j^{\text{th}}$ subsystem as:
\begin{equation}
\nu_j = \nu_{rb_j} + \nu_{act_j} \quad,
\label{eq:nu_j}
\end{equation}
where $\nu_{rb_j}$ and $\nu_{act_j}$ represent the non-negative accompanying functions associated with the rigid bodies and the actuator, respectively.

Let the non-negative accompanying function for the rigid body part be chosen in the sense of Definition \ref{nu def}, as follows:
\begin{equation}
\begin{split}
{\nu}_{rb_j} &= \sum_{ A \in \Upsilon_j} \left[ \dfrac{1}{2} \, \left( { }^{\mathbf{A}} \boldsymbol{V_r} - { }^{\mathbf{A}} \boldsymbol{V} \right)^T \, \mathbf{M}_{\mathbf{A}} \, \left( { }^{\mathbf{A}} \boldsymbol{V_r} - { }^{\mathbf{A}} \boldsymbol{V} \right) \right.\\
&\left. \quad \quad \quad +  \gamma \mathcal{D}_F(\mathcal{L}_{A}\rVert \hat{\mathcal{L}}_{A}) \right],
\end{split}
\label{eq:nu_rb_j}
\end{equation}
where $\mathcal{D}_F(\mathcal{L}_{A}\rVert \hat{\mathcal{L}}_{A})$ is defined as in~(\ref{Df}). Furthermore, $\Upsilon_1 = \left\lbrace T_1 \right\rbrace$ corresponds to the base subsystem, while $\Upsilon_j = \left\lbrace A_{j}, B_{c_j}, C_{j}, D_{j} \right\rbrace$ for $j = 2, 3, 4$ denote the lift, tilt, and wrist subsystems, respectively.
Taking the time derivative of~\eqref{eq:nu_rb_j} and following the procedure explained in \cite[Appendix C]{petrovic2022mathematical} and \cite[Appendix A]{hejrati2025orchestrated} yield the follows:
\begin{equation}
    \begin{split}
        &\Dot{{\nu}}_{rb_j} \leq -\sum_{ A \in \Upsilon_j} \left[ \left( { }^{\mathbf{A}} \boldsymbol{V_r} - { }^{\mathbf{A}} \boldsymbol{V} \right)^T\,\mathbf{K_A}  \left( { }^{\mathbf{A}} \boldsymbol{V_r} - { }^{\mathbf{A}} \boldsymbol{V} \right) \right] + {V\!P\!F\!s}_j, \\
        &{V\!P\!F\!s}_1 = p_{T_1} - p_{B_{c_2}}, \\
        &{V\!P\!F\!s}_2 = p_{B_{c_2}} - p_{B_{c_3}} + (F_{\mathrm{hyb}_{r_2}} - F_{\mathrm{hyb}_2}) (\dot{x}_{r_2} -\dot{x}_2), \\
        &{V\!P\!F\!s}_3 = p_{B_{c_3}} - p_{B_{c_4}} + (F_{\mathrm{hyb}_{r_3}} - F_{\mathrm{hyb}_3}) (\dot{x}_{r_3} -\dot{x}_3), \\
        &{V\!P\!F\!s}_1 = p_{B_{c_4}} - p_{T_{4}},
    \end{split}
\label{eq:d_nu_rb}
\end{equation}
where $p_A$ is defined in Definition~\ref{Def pA}. 

For the actuator dynamics, we consider the actual forces generated by each EMLA derived from~\eqref{eq:residual_blending_hat}, based on the EMLA dynamics described in Section~\ref{subsec:emla_dynamic}, and we proceed to design the required forces as follows:
\begin{equation}
\left\{
\begin{aligned}
&F_{\mathrm{hyb}} = (1-\alpha) \left( \frac{2 \pi N_{\mathrm{gear}}}{\rho} \right) \left[ \frac{3}{2} p \left( \lambda_d i_q - \lambda_q i_d \right) \right. \\
& \left.-\left( J_m + \frac{M_{\mathrm{act}}}{N_{\mathrm{gear}}+\eta_f} \right) \ddot{\theta}_m - C_m \dot{\theta}_m - \tau_C \right] + \alpha  \hat{F}_{\mathrm{ext}}, \\
&F_{\mathrm{hyb}_r} = (1-\alpha) \left( \frac{2 \pi N_{\mathrm{gear}}}{\rho} \right) \left[ \frac{3}{2} p \left( \lambda_d i_{q_r} - \lambda_q i_{d_r} \right) \right. \\
& \left.-\left( J_m + \frac{M_{\mathrm{act}}}{N_{\mathrm{gear}}+\eta_f} \right) \ddot{\theta}_m - C_m \dot{\theta}_m - \tau_C \right] + \alpha  \hat{F}_{\mathrm{ext}}.
\end{aligned}
\right.
\label{eq:Actual_Required_Force_EMLA}
\end{equation}
By subtracting these equations, the force error for each EMLA can be computed as follows:
\begin{equation}
\begin{aligned}
&F_{\mathrm{hyb}_r} - F_{\mathrm{hyb}}  \\
&=(1-\alpha) \left( \frac{2 \pi N_{\mathrm{gear}}}{\rho} \right) \left( \frac{3}{2} p \right) \left[ \lambda_d (i_{q_r} - i_q) - \lambda_q (i_{d_r} - i_d) \right] \\
&=\beta_1 (i_{q_r} - i_q) - \beta_2 (i_{d_r} - i_d),
\end{aligned}
\label{eq:Error_Force_EMLA}
\end{equation}
where \( \beta_1 \) and \( \beta_2 \) are introduced to encapsulate system parameters and simplify the equation. Differentiating~\eqref{eq:Error_Force_EMLA}, substituting \( \frac{di_{d_r}}{dt} \) and \( \frac{di_{q_r}}{dt} \) from~\eqref{eq:voltages_Control} and \( \frac{di_{d}}{dt} \) and \( \frac{di_{q}}{dt} \) from~\eqref{eq:dq_voltages}, and using \( v_{d_r} = v_d \) and \( v_{q_r} = v_q \), yields:
\begin{equation}
\begin{aligned}
\frac{d}{dt}(F_{\mathrm{hyb}_r} - F_{\mathrm{hyb}}) &= \frac{\beta_1}{L_q} \left[ -K_f(F_{\mathrm{hyb}_r} - F_{\mathrm{hyb}}) - K_v(\dot{x}_r -\dot{x}) \right] \\
&+ \frac{\beta_2}{L_d} \left[ K_i(i_{d_r}-i_d) \right].
\end{aligned}
\label{eq:D_Error_Force_EMLA}
\end{equation}
Now, consider the non-negative auxiliary function $\nu_{act_j}$ associated with the lift and tilt closed chains ($j = 2, 3$), defined as follows:
\begin{equation}
\nu_{act_j} =  \frac{K_{1_j}}{2}(F_{\mathrm{hyb}_{r_j}} - F_{\mathrm{hyb}_j})^2 + \frac{K_{2_j}}{2} (i_{d_{r_j}} -i_{d_j})^2.
\label{eq:v_EMLA}
\end{equation}
Differentiating this equation, in conjunction with using~\eqref{eq:voltages_Control} and~\eqref{eq:D_Error_Force_EMLA}, yields:
\begin{equation}
\begin{aligned}
&\dot{\nu}_{act_j}  =  K_{1_j} (F_{\mathrm{hyb}_{r_j}} - F_{\mathrm{hyb}_j}) \frac{d}{dt}(F_{\mathrm{hyb}_{r_j}} - F_{\mathrm{hyb}_j})\\
&\quad \quad \quad + K_{2_j} (i_{d_{r_j}} -i_{d_j})\frac{d}{dt}(i_{d_{r_j}} -i_{d_j})\\
&\quad \quad =  K_{1_j} (F_{\mathrm{hyb}_{r_j}} - F_{\mathrm{hyb}_j}) \left( \frac{\beta_{1_j}}{L_{q_j}} \left[ -K_{f_j}(F_{\mathrm{hyb}_{r_j}} - F_{\mathrm{hyb}_j}) \right. \right.\\
& \quad \quad \quad -  \left.\left. K_{v_j}(\dot{x}_{r_j} -\dot{x}_j) \right] +  \frac{\beta_{2_j}}{L_{d_j}} \left[ K_{i_j} (i_{d_{r_j}}-i_{d_j}) \right] \right)\\
& \quad \quad \quad  + \frac{K_{2_j}}{L_{d_j}} (i_{d_{r_j}} -i_{d_j}) \left( -K_{i_j} (i_{d_{r_j}}-i_{d_j}) \right)\\
&\quad \quad = -\frac{K_{1_j} \beta_{1_j} K_{f_j}}{L_{q_j}} (F_{\mathrm{hyb}_{r_j}} - F_{\mathrm{hyb}_j})^2 \\
& \quad \quad \quad  - \frac{K_{1_j} \beta_{1_j} K_{v_j}}{L_{q_j}} (F_{\mathrm{hyb}_{r_j}} - F_{\mathrm{hyb}_j}) (\dot{x}_{r_j} -\dot{x}_j) \\
& \quad \quad \quad + \frac{K_{1_j} \beta_{2_j} K_{i_j}}{L_{d_j}} (F_{\mathrm{hyb}_{r_j}} - F_{\mathrm{hyb}_j}) (i_{d_{r_j}} -i_{d_j})\\
& \quad \quad \quad - \frac{K_{2_j} K_{i_j}}{L_{d_j}} (i_{d_{r_j}} -i_{d_j})^2.
\end{aligned}
\label{eq:dv_EMLA1}
\end{equation}
For the term~\((F_{\mathrm{hyb}_{r_j}} - F_{\mathrm{hyb}_j})(i_{d_{r_j}} - i_{d_j})\), applying Young's inequality yields:
\begin{equation}
\begin{aligned}
&\dot{\nu}_{act_j} \leq -\frac{K_{1_j} \beta_{1_j} K_{f_j}}{L_{q_j}} (F_{\mathrm{hyb}_{r_j}} - F_{\mathrm{hyb}_j})^2\\
& \quad \quad \quad   - \frac{K_{1_j} \beta_{1_j} K_{v_j}}{L_{q_j}} (F_{\mathrm{hyb}_{r_j}} - F_{\mathrm{hyb}_j}) (\dot{x}_{r_j} -\dot{x}_j) \\
& \quad \quad \quad  + \frac{K_{1_j} \beta_{2_j} K_{i_j}}{2 L_{d_j}} \left[ (F_{\mathrm{hyb}_{r_j}} - F_{\mathrm{hyb}_j})^2 + (i_{d_{r_j}} -i_{d_j})^2 \right]\\
& \quad \quad \quad  - \frac{K_{2_j} K_{i_j}}{L_{d_j}} (i_{d_{r_j}} -i_{d_j})^2\\
&\quad \quad  = - K_{1_j} \left( \frac{\beta_{1_j} K_{f_j}}{L_{q_j}} - \frac{\beta_{2_j} K_{i_j}}{2 L_{d_j}} \right)(F_{\mathrm{hyb}_{r_j}} - F_{\mathrm{hyb}_j})^2 \\
& \quad \quad \quad  - \frac{K_{i_j}}{L_{d_j}}  \left( K_{2_j} - \frac{K_{1_j} \beta_{2_j}}{2} \right) (i_{d_{r_j}} -i_{d_j})^2 \\
& \quad \quad \quad - \frac{K_{1_j} \beta_{1_j} K_{v_j}}{L_{q_j}} (F_{\mathrm{hyb}_{r_j}} - F_{\mathrm{hyb}_j}) (\dot{x}_{r_j} -\dot{x}_j).
\end{aligned}
\label{eq:dv_EMLA2}
\end{equation}
Therefore, if the following conditions are satisfied:
\begin{equation}
\left\{
\begin{aligned}
\frac{\beta_{1_j} K_{f_j}}{L_{q_j}} &\ge \frac{\beta_{2_j} K_{i_j}}{2 L_{d_j}},\\
K_{2_j} &\ge \frac{K_{1_j} \beta_{2_j}}{2},\\
K_{1_j} &= \frac{L_{q_j}}{\beta_{1_j} K_{v_j}},
\end{aligned}
\label{eq:EMLA_stability_conditions}
\right.
\end{equation}
it can be concluded that:
\begin{equation}
\begin{aligned}
&\dot{\nu}_{act_j} \leq - K_{1_j} \left( \frac{\beta_{1_j} K_{f_j}}{L_{q_j}} - \frac{\beta_{2_j} K_{i_j}}{2 L_{d_j}} \right)(F_{\mathrm{hyb}_{r_j}} - F_{\mathrm{hyb}_j})^2 \\
& \quad \quad \quad - \frac{K_{i_j}}{L_{d_j}}  \left( K_{2_j} - \frac{K_{1_j} \beta_{2_j}}{2} \right) (i_{d_{r_j}} -i_{d_j})^2 \\
& \quad \quad \quad - (F_{\mathrm{hyb}_{r_j}} - F_{\mathrm{hyb}_j}) (\dot{x}_{r_j} -\dot{x}_j) .
\end{aligned}
\label{eq:dv_EMLA3}
\end{equation}

Eventually, by differentiating~\eqref{eq:nu_j} and substituting~\eqref{eq:d_nu_rb} and~\eqref{eq:dv_EMLA3} with the conditions~\eqref{eq:EMLA_stability_conditions}, one obtains:
\begin{equation}
\begin{split}
\dot{\nu}_1 &\leq -\sum_{ A \in \Upsilon_1} \left[ \left( { }^{\mathbf{A}} \boldsymbol{V_r} - { }^{\mathbf{A}} \boldsymbol{V} \right)^T\,\mathbf{K_A} \left( { }^{\mathbf{A}} \boldsymbol{V_r} - { }^{\mathbf{A}} \boldsymbol{V} \right) \right]\\
&\quad  + p_{T_{1}} - p_{B_{c_2}},\\
\dot{\nu}_j &\leq -\sum_{ A \in \Upsilon_j} \left[ \left( { }^{\mathbf{A}} \boldsymbol{V_r} - { }^{\mathbf{A}} \boldsymbol{V} \right)^T\,\mathbf{K_A} \left( { }^{\mathbf{A}} \boldsymbol{V_r} - { }^{\mathbf{A}} \boldsymbol{V} \right) \right]\\
&\quad  - K_{1_j} \left( \frac{\beta_{1_j} K_{f_j}}{L_{q_j}} - \frac{\beta_{2_j} K_{i_j}}{2 L_{d_j}} \right)(F_{\mathrm{hyb}_{r_j}} - F_{\mathrm{hyb}_j})^2\\
&\quad  - \frac{K_{i_j}}{L_{d_j}}  \left( K_{2_j} - \frac{K_{1_j} \beta_{2_j}}{2} \right) (i_{d_{r_j}} -i_{d_j})^2 \\
&\quad  + p_{B_{c_j}} - p_{B_{c_{j+1}}} \quad \quad \quad \quad \quad \quad \quad \quad  (j=2, 3),\\
\dot{\nu}_4 &\leq -\sum_{ A \in \Upsilon_4} \left[ \left( { }^{\mathbf{A}} \boldsymbol{V_r} - { }^{\mathbf{A}} \boldsymbol{V} \right)^T\,\mathbf{K_A} \left( { }^{\mathbf{A}} \boldsymbol{V_r} - { }^{\mathbf{A}} \boldsymbol{V} \right) \right]\\
&\quad  + p_{B_{c_4}} - p_{T_{4}},
\end{split}
\label{eq:virtaul_stability}
\end{equation}
demonstrating the virtual stability of each subsystem in the sense of Definition~\ref{Def_VS}.

\begin{thm}
    The entire system, decomposed into rigid body and actuator subsystems, is stable under the low-level control law (\ref{eq:voltages_Control}); high-level control laws~\eqref{Eq:Actuator_Tilt_Required_Dynamics}, \eqref{Eq:Actuator_Lift_Required_Dynamics}, and \eqref{Eq:Actuator_Base_Required_Dynamics2}; and adaptation law (\ref{L adapt}).
\end{thm}

\textit{Proof}: Consider the non-negative accompanying function for the overall system defined as:
\begin{equation}
\nu = \sum_{j=1}^{4} \nu_{j} = \sum_{j=1}^{4} \left( \nu_{rb_j} + \nu_{act_j} \right).
\label{total nu}
\end{equation}
Differentiating~\eqref{total nu} with respect to time and substituting from~\eqref{eq:virtaul_stability}, while incorporating the conditions in~\eqref{eq:EMLA_stability_conditions}, yields
\begin{equation}
\begin{split}
\dot{\nu} &\leq -\sum_{ A \in \Upsilon} \left[ \left( { }^{\mathbf{A}} \boldsymbol{V_r} - { }^{\mathbf{A}} \boldsymbol{V} \right)^T\,\mathbf{K_A} \left( { }^{\mathbf{A}} \boldsymbol{V_r} - { }^{\mathbf{A}} \boldsymbol{V} \right) \right]\\
&\quad -\sum_{j=2}^{3} \left[ K_{1_j} \left( \frac{\beta_{1_j} K_{f_j}}{L_{q_j}} - \frac{\beta_{2_j} K_{i_j}}{2 L_{d_j}} \right)(F_{\mathrm{hyb}_{r_j}} - F_{\mathrm{hyb}_j})^2 \right.\\
&\quad \quad \quad \left. + \frac{K_{i_j}}{L_{d_j}}  \left( K_{2_j} - \frac{K_{1_j} \beta_{2_j}}{2} \right) (i_{d_{r_j}} - i_{d_j})^2 \right],
\end{split}
\label{eq:virtaul_stability2}
\end{equation}
where $\Upsilon = \Upsilon_1 \cup \Upsilon_2 \cup \Upsilon_3 \cup \Upsilon_4$. This inequality establishes the stability of the overall system under the proposed controller. Furthermore, since the inputs to the neural networks are bounded, their outputs are also bounded. Therefore, all neural network outputs remain bounded, which completes the proof.

\section{Results and Discussion}
\label{sec:results_and_discussion}
\begin{figure}[ht]
\centering
\subfloat[]{\includegraphics[width=1.44in]{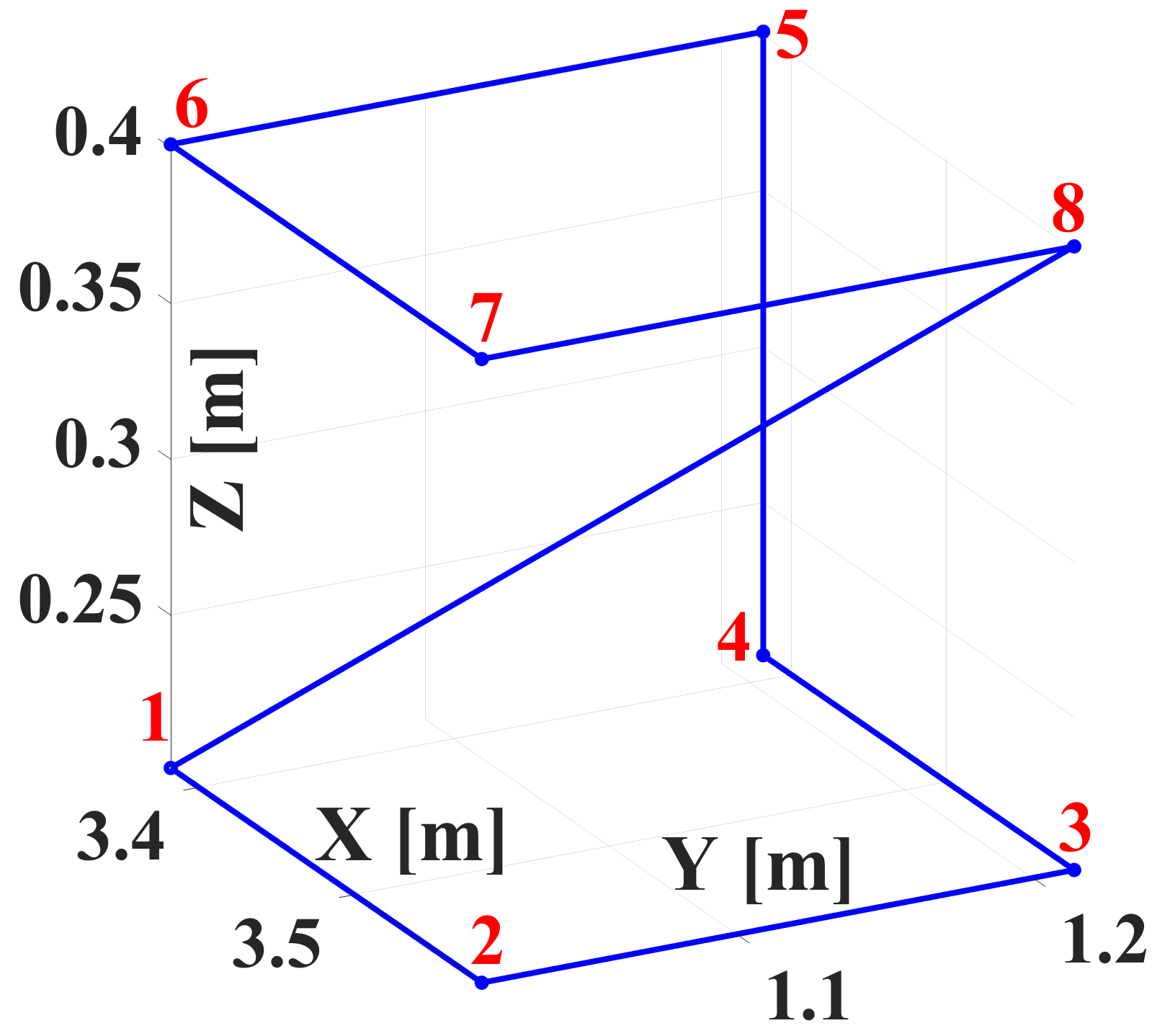}%
\label{Fig:SchematicISO}}
\subfloat[]{\includegraphics[width=1.08in]{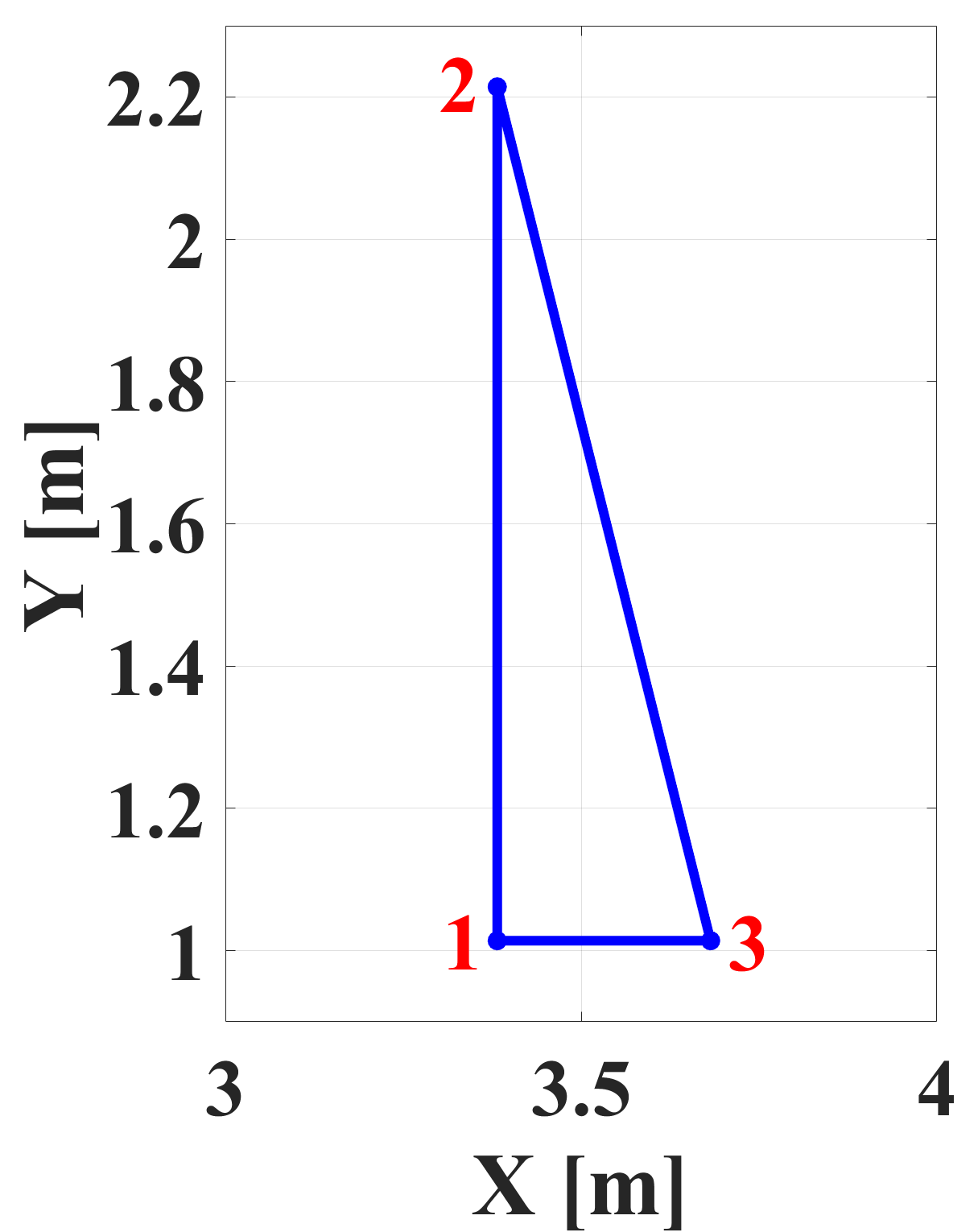}%
\label{Fig:SchematicTriangular}}
\caption{Schematic view of the trajectories. a) Cubic trajectory with $20$ cm length. b) Planar triangular trajectory.}
\label{Fig:SchematicTrajectories}
\end{figure}

To rigorously evaluate the proposed adaptive modular controller for the all-electric HDRM, its performance is tested under standardized motion scenarios. Two representative trajectories are considered: (1) a cubic trajectory, commonly used to benchmark industrial robot accuracy and repeatability, and (2) a planar triangular trajectory for assessing motion in constrained two-dimensional environments. Both are illustrated in Fig.~\ref{Fig:SchematicTrajectories}.

\subsection{Simulation Results}

In this section, the proposed adaptive modular controller is compared with two baselines: (1) a modular controller without adaptation, and (2) a PD controller without adaptation or model-based dynamic compensation, retaining only proportional-derivative feedback.

\begin{figure}[ht]
    \centering
    \includegraphics[width=0.4\textwidth]{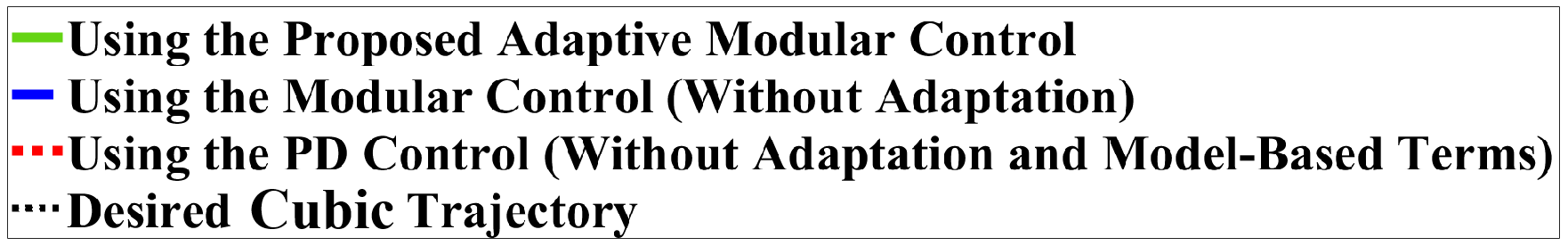}
    \vspace{1mm}
    
    \centering
    \begin{minipage}[b]{0.24\textwidth}
        \centering
        \includegraphics[width=\textwidth]{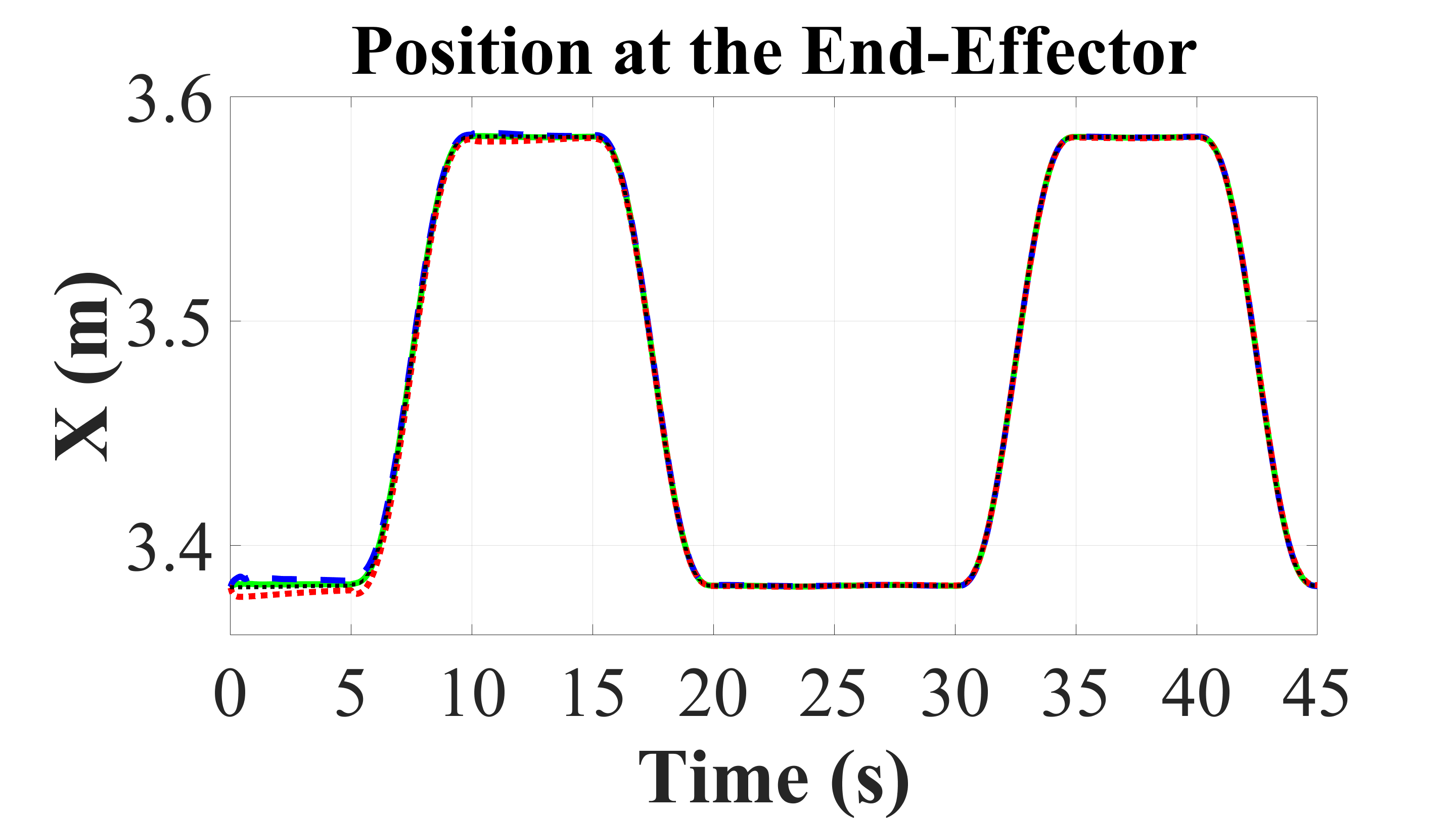}
    \end{minipage}
    \hfill
    \begin{minipage}[b]{0.24\textwidth}
        \centering
        \includegraphics[width=\textwidth]{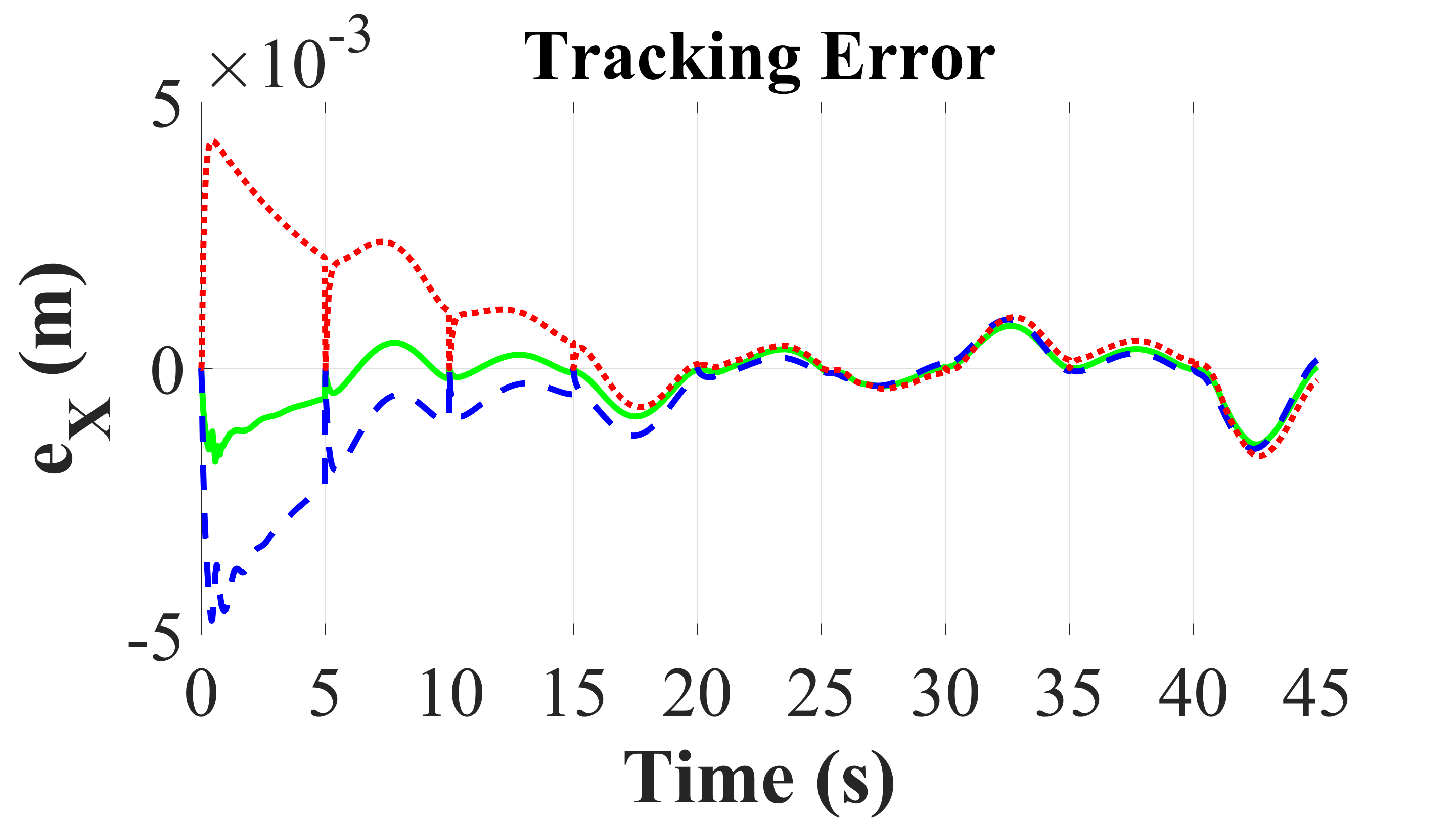}
    \end{minipage}
    \vspace{1mm}    
    
    \begin{minipage}[b]{0.24\textwidth}
        \centering
        \includegraphics[width=\textwidth]{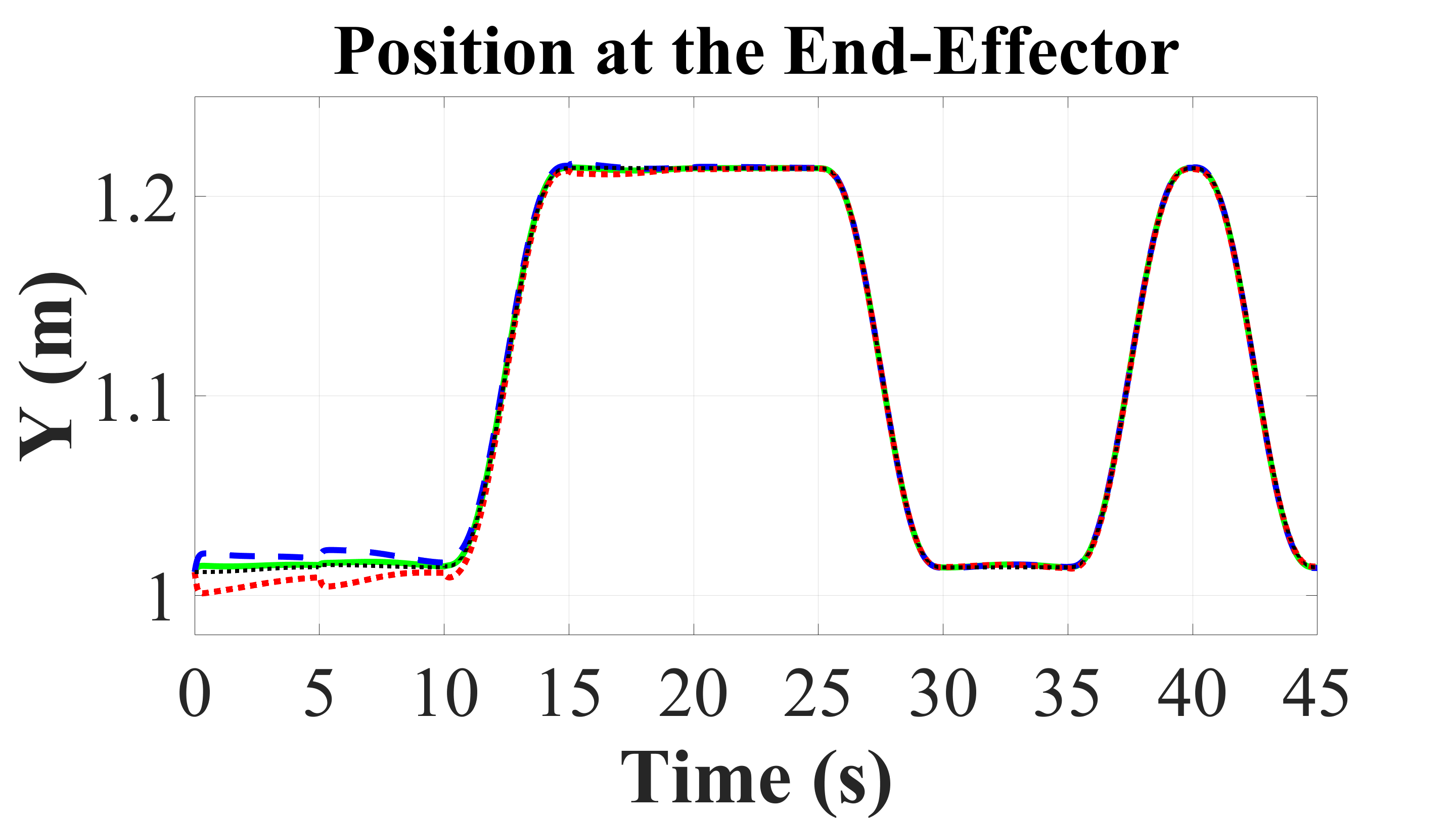}
    \end{minipage}
    \hfill
    \begin{minipage}[b]{0.24\textwidth}
        \centering
        \includegraphics[width=\textwidth]{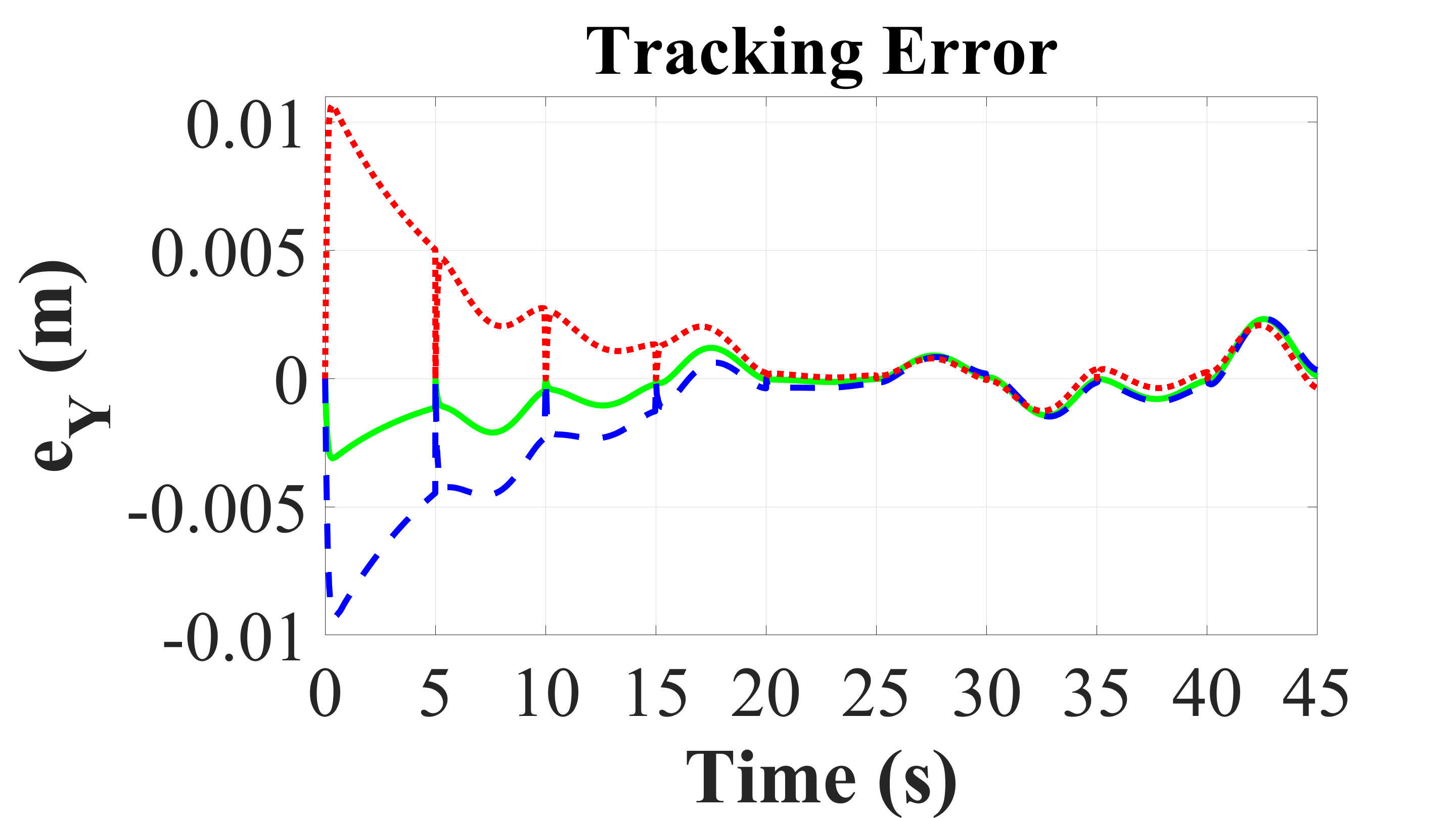}
    \end{minipage}
    \vspace{1mm}
    
    \begin{minipage}[b]{0.24\textwidth}
        \centering
        \includegraphics[width=\textwidth]{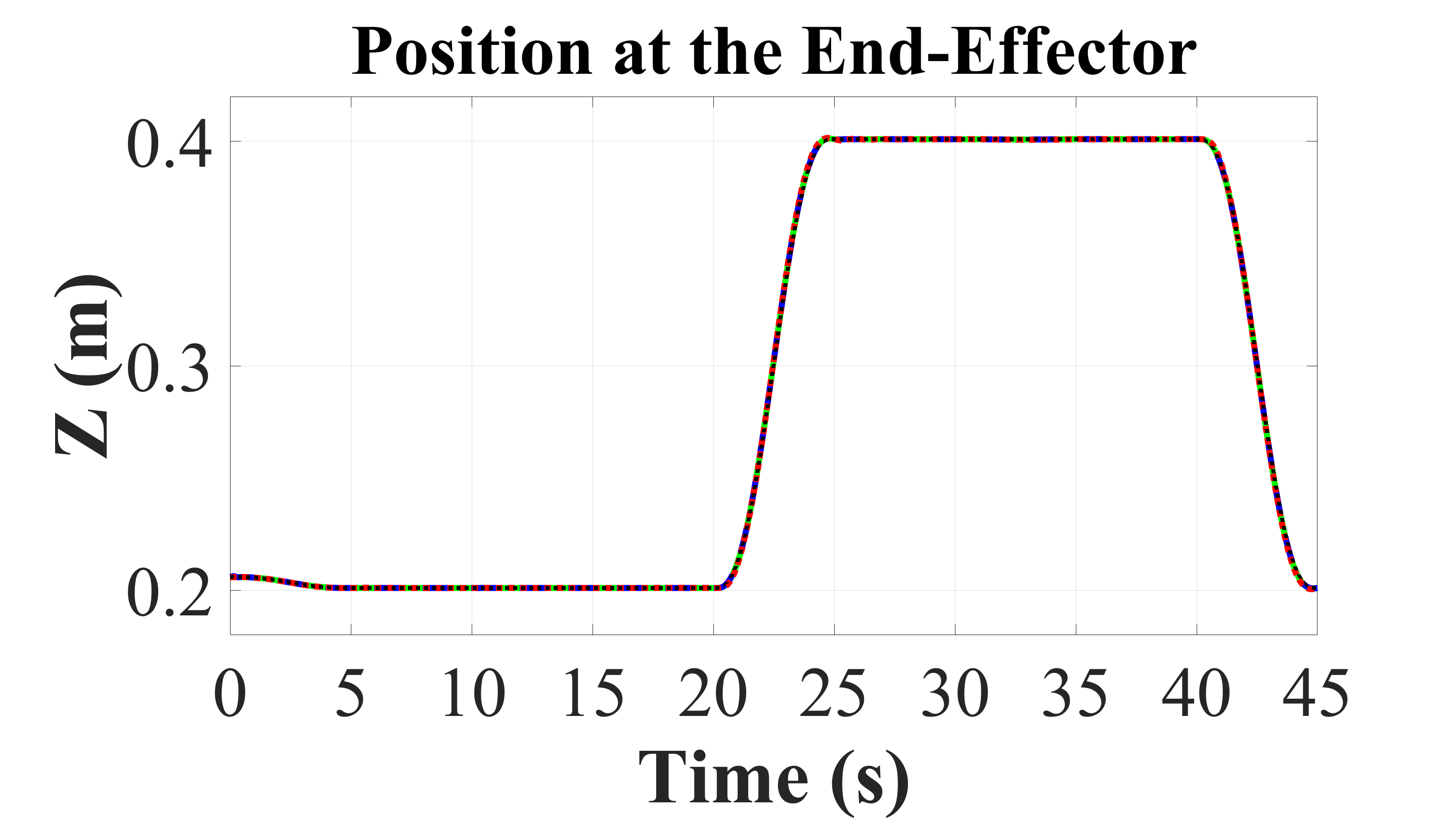}
    \end{minipage}
    \hfill
    \begin{minipage}[b]{0.24\textwidth}
        \centering
        \includegraphics[width=\textwidth]{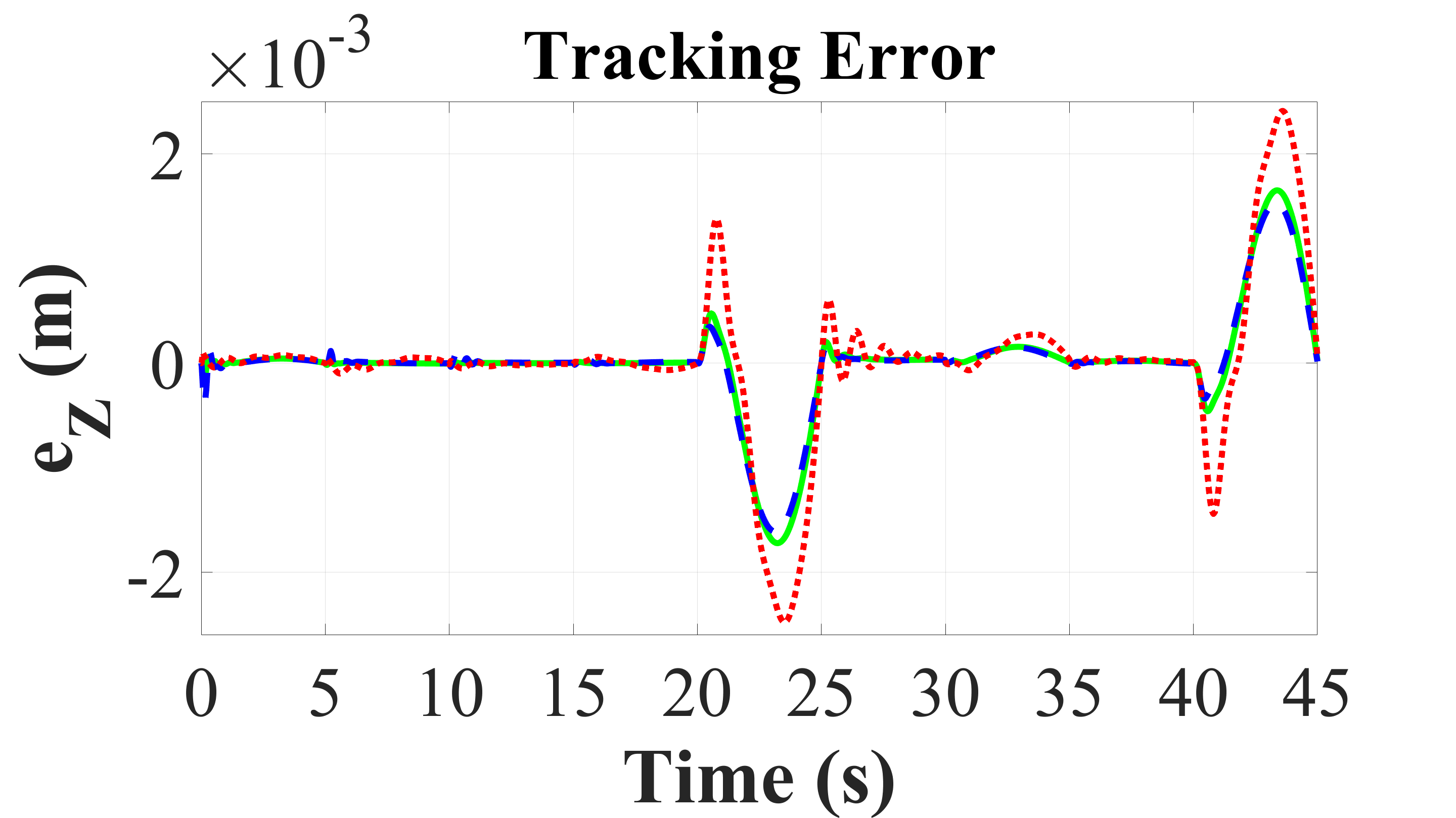}
    \end{minipage}

    \caption{Performance comparison of controllers using the cubic trajectory.}
    \label{Fig:SimulationISO}
\end{figure}

\begin{figure}[ht]
    \centering
    \includegraphics[width=0.4\textwidth]{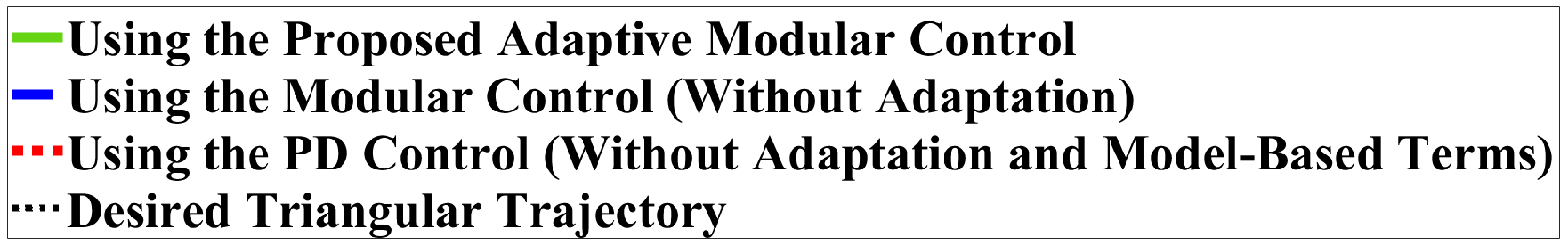}
    \vspace{1mm}
    
    \centering
    \begin{minipage}[b]{0.24\textwidth}
        \centering
        \includegraphics[width=\textwidth]{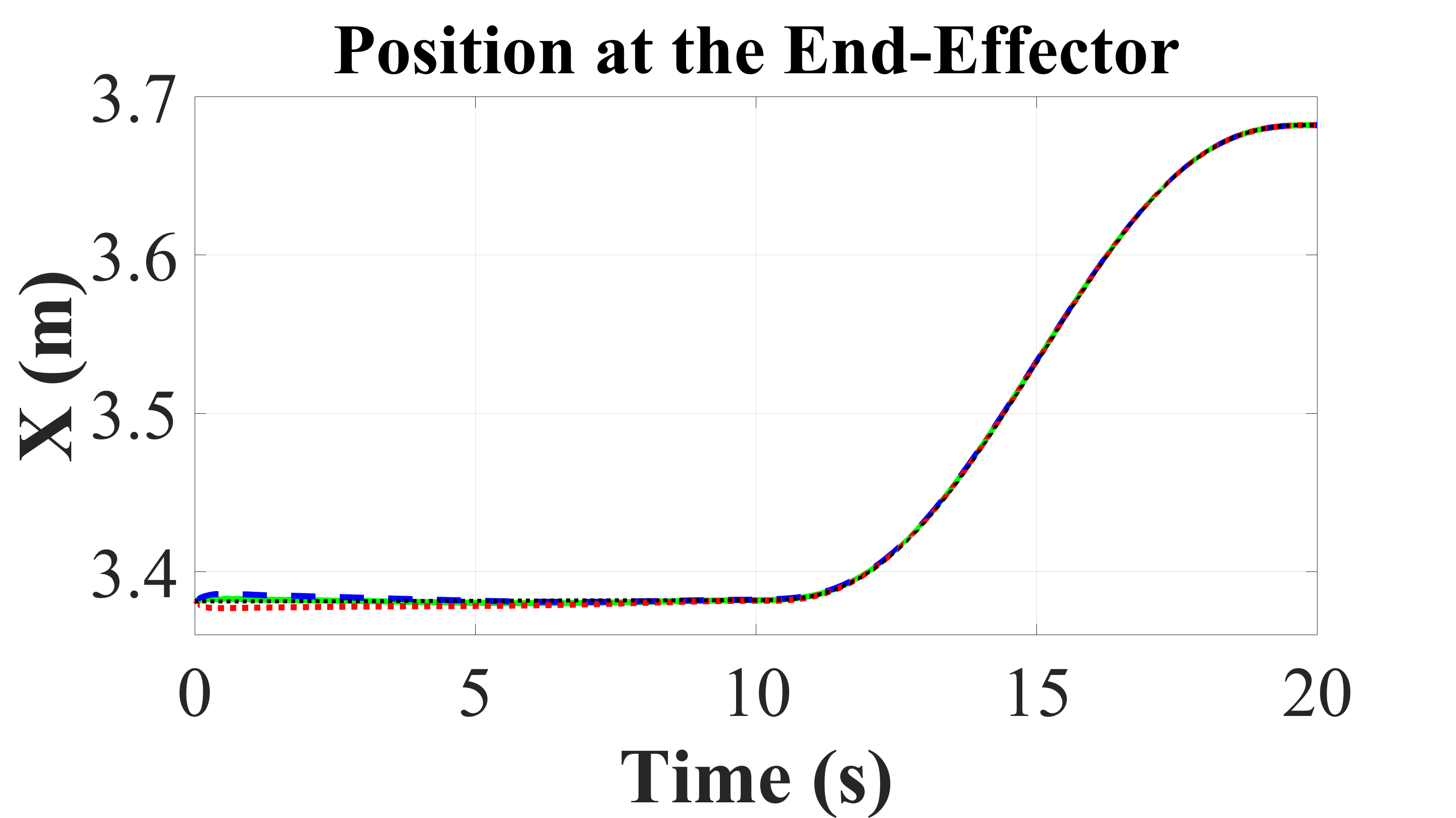}
    \end{minipage}
    \hfill
    \begin{minipage}[b]{0.24\textwidth}
        \centering
        \includegraphics[width=\textwidth]{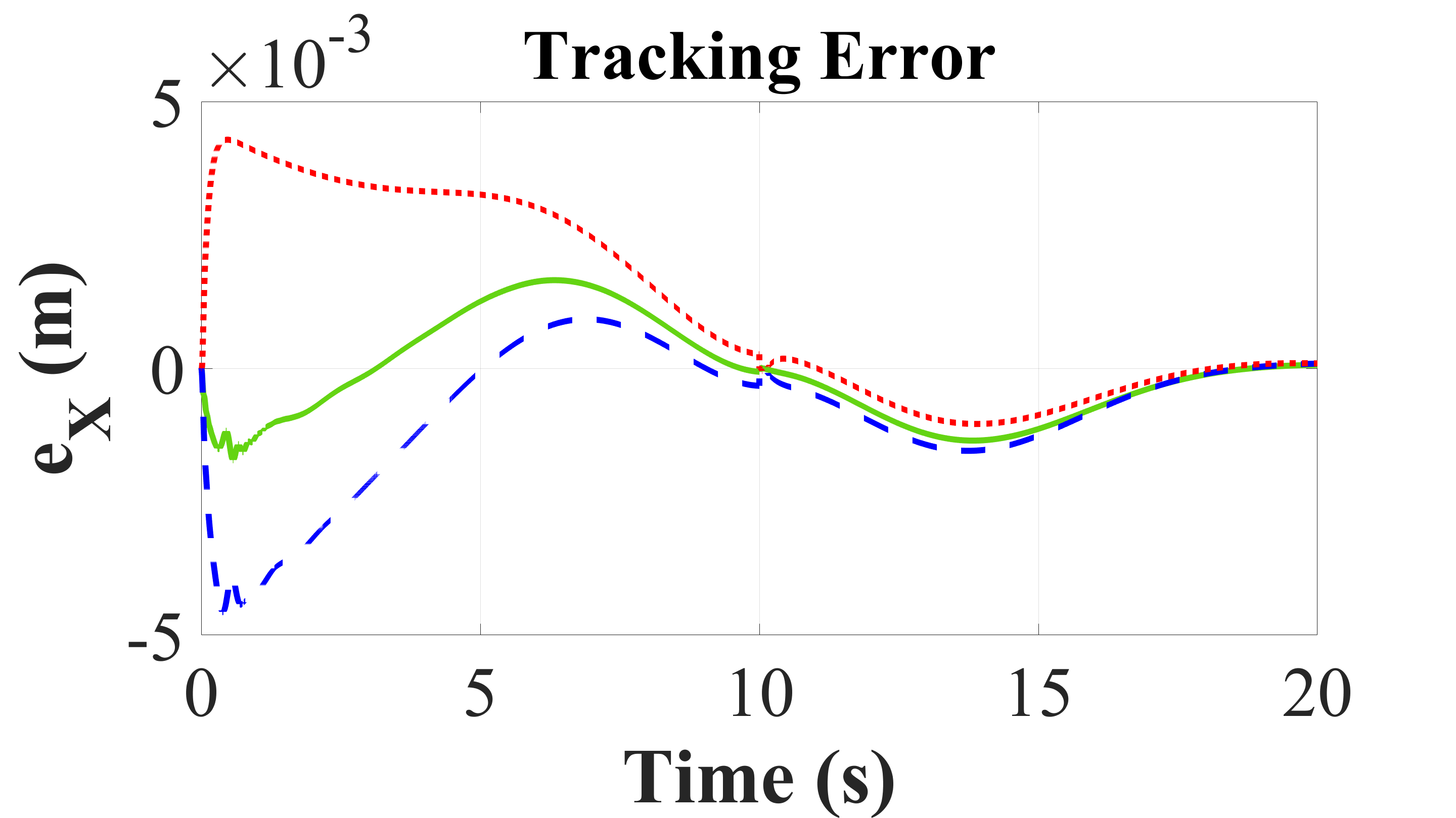}
    \end{minipage}
    \vspace{1mm}    
    
    \begin{minipage}[b]{0.24\textwidth}
        \centering
        \includegraphics[width=\textwidth]{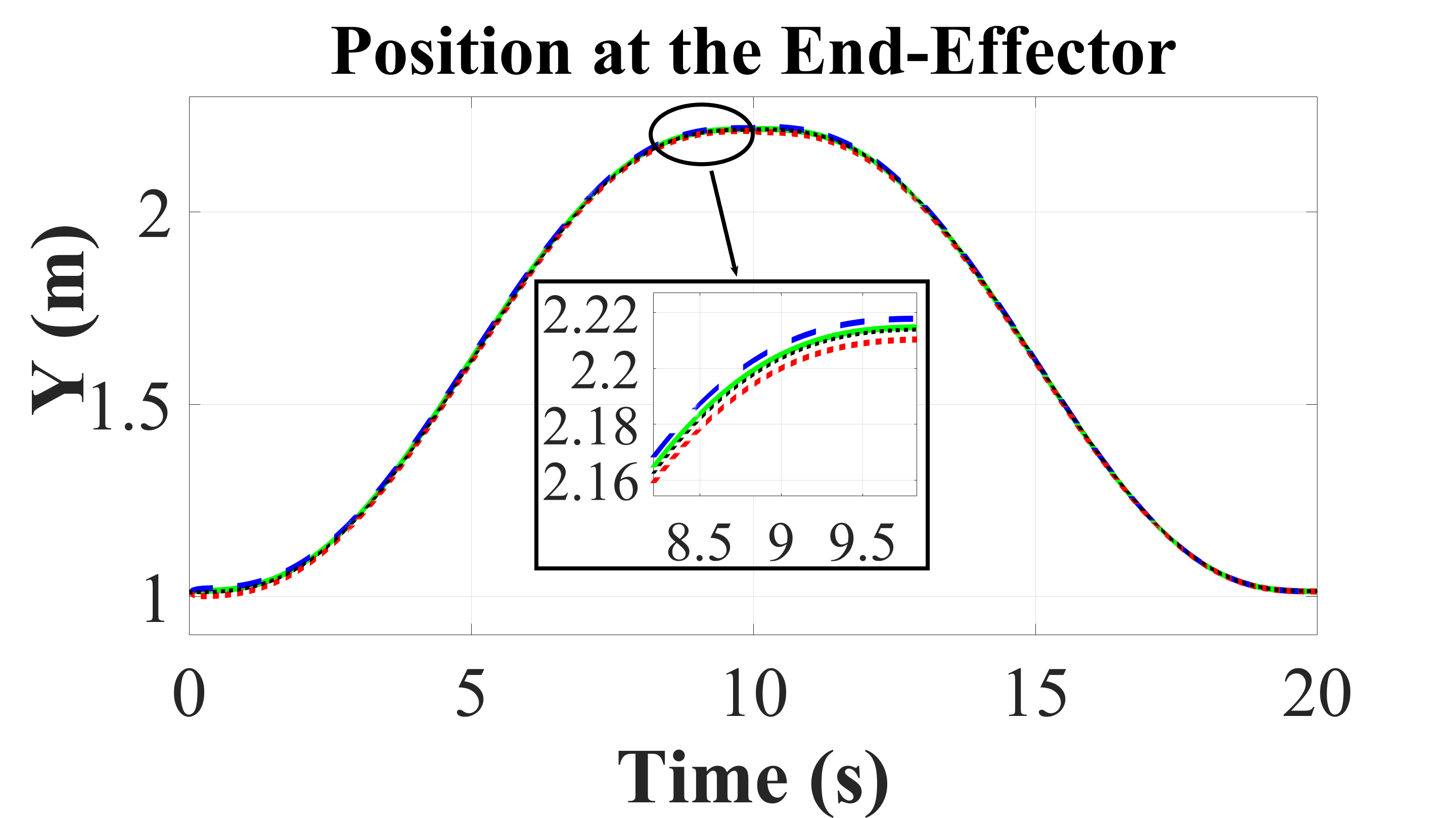}
    \end{minipage}
    \hfill
    \begin{minipage}[b]{0.24\textwidth}
        \centering
        \includegraphics[width=\textwidth]{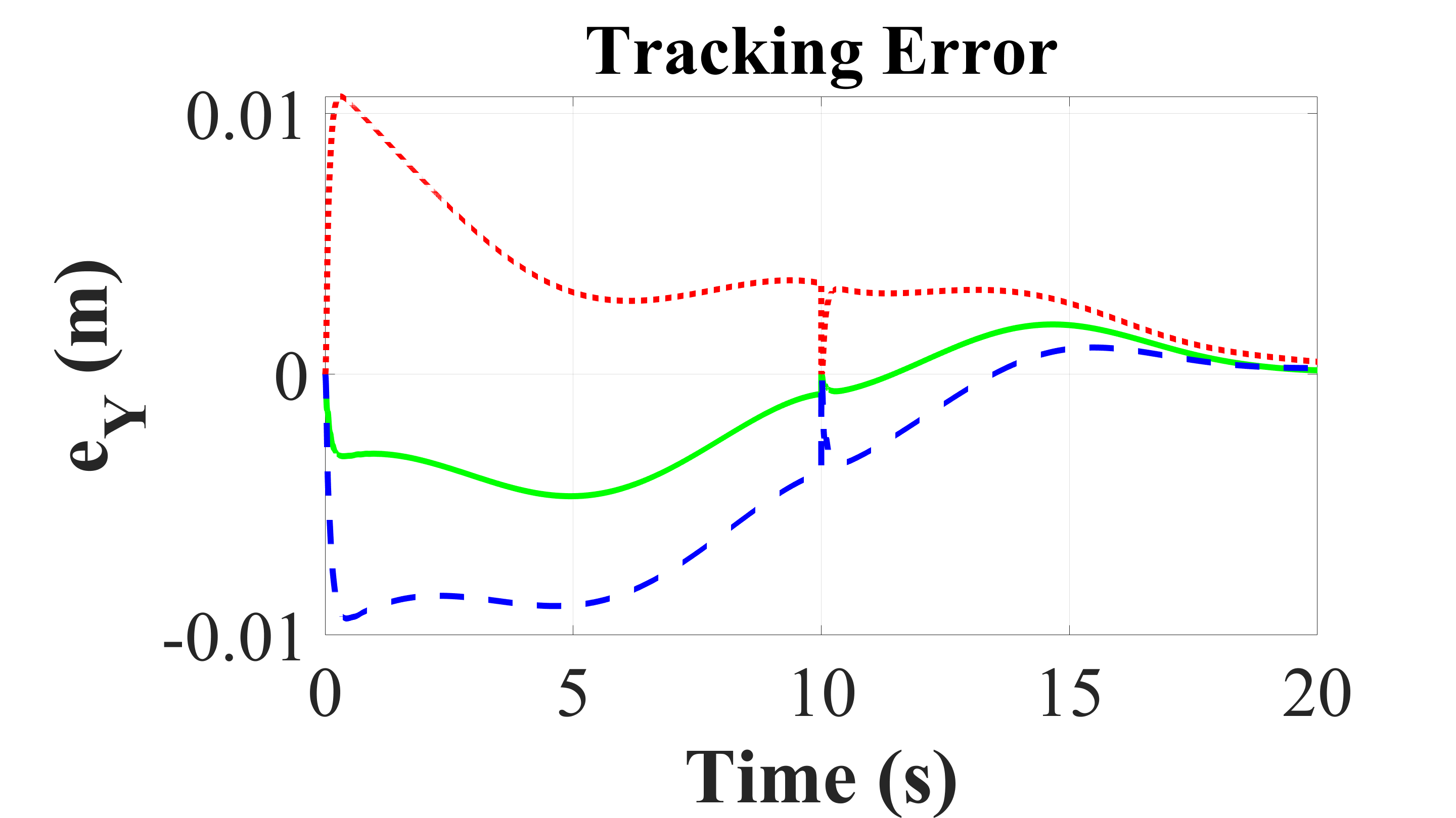}
    \end{minipage}

    \caption{Performance comparison of controllers using the planar triangular trajectory.}
    \label{Fig:SimulationArbitrary}
\end{figure}

Fig. \ref{Fig:SimulationISO} illustrates the Cartesian tracking performance in the end effector for the three controllers, adaptive modular, modular and PD, while performing the cubic trajectory shown in Figure \ref{Fig:SchematicISO}. This standardized 3D motion path provides a comprehensive assessment of each controller’s ability to handle complex spatial trajectories. It is observed that the adaptive modular controller achieves the lowest tracking errors, indicating better handling of uncertainties and improved accuracy.
The tracking results of the end-effector in Cartesian space for the three controllers during execution of the triangular trajectory (illustrated in Fig.~\ref{Fig:SchematicTriangular}) are presented in Fig.~\ref{Fig:SimulationArbitrary}.
It can be observed that the proposed adaptive modular controller demonstrates superior tracking performance compared to the other controllers when following the triangular trajectory. Specifically, it achieves significantly lower Cartesian tracking errors throughout the motion, indicating its effectiveness in compensating for dynamic uncertainties and improving end-effector tracking accuracy.

To assess the robustness of the proposed controller against parametric uncertainties, the root mean squared error (RMSE) is employed as a performance metric. The controller's tracking performance is evaluated under density variations of $+40\%$, $+20\%$, nominal, $-20\%$, and $-40\%$ for the rigid-body components while following the two previously described trajectories. The corresponding results are presented in Table~\ref{tab:rmse_uncertainties}.
\begin{table}[ht]
\caption{RMSE of the proposed controller under parametric uncertainties for two trajectories.}
\label{tab:rmse_uncertainties}
\centering
\begin{tabular}{|c|c|c|}
\hline
\textbf{Uncertainty} & \textbf{Cubic Trajectory} & \textbf{Triangular Trajectory} \\
\hline
+40 \% & 0.0024 m & 0.0042 m \\
+20 \% & 0.0019 m & 0.0035 m \\
0 \% (Nominal) & 0.0015 m & 0.0029 m \\
-20 \% & 0.0029 m & 0.0050 m \\
-40 \% & 0.0036 m & 0.0060 m \\
\hline
\end{tabular}
\end{table}

As observed in Table~\ref{tab:rmse_uncertainties}, the RMSE remains consistently low across all levels of uncertainty, with only a gradual increase as the deviation from nominal parameters grows. This behavior demonstrates the controller’s ability to maintain accurate trajectory tracking despite significant parametric variations.

\subsection{Experimental Results}

For an experimental validation of the performance of the proposed adaptive modular controller for the all-electric HDRM, a series of motion-tracking experiments was carried out using a dedicated EMLA-driven testbed. The primary objective was to assess the trajectory tracking fidelity and control robustness under realistic operating conditions.

\textcolor{black}{Although the proposed framework is formulated for a 6-DoF HDRM, the hardware validation is conducted on a 1-DoF EMLA testbed, since such a setup is necessary to isolate the actuator-level dynamics on which the method is built without interference from multi-body coupling and inter-joint effects. The scalability of the approach to the full HDRM is supported by the 6-DoF analytical model and the simulation results.}

The experimental platform consists of a three-phase, 380/480V, 8-pole Nidec PMSM rated at 11.6~kW. This motor is coupled to a reduction gearbox and a screw mechanism, forming a complete EMLA. The EMLA system is mechanically connected to another actuator that acts as an external load emulator, capable of generating dynamic and time-varying force profiles via an electrohydraulic human–machine interface (HMI). The load emulation closely replicates the operational disturbances typically encountered in the lift and tilt joints of an HDRM. Meanwhile, a Unidrive M700 servo controller governs the PMSM operation through a subsystem-based control interface \cite{bahari2025system}. The setup features real-time signal acquisition and actuation feedback, coordinated via an EtherCAT communication network, enabling synchronized voltage control, force feedback, and trajectory monitoring with high temporal resolution. The same benchmark trajectories used in the simulation section were employed for experimental validation: the cubic trajectory for assessing 3D tracking accuracy, and a planar triangular trajectory for evaluating controller performance in constrained planar motion.

\begin{figure}[ht]
\centering
\includegraphics[width=2.2in]{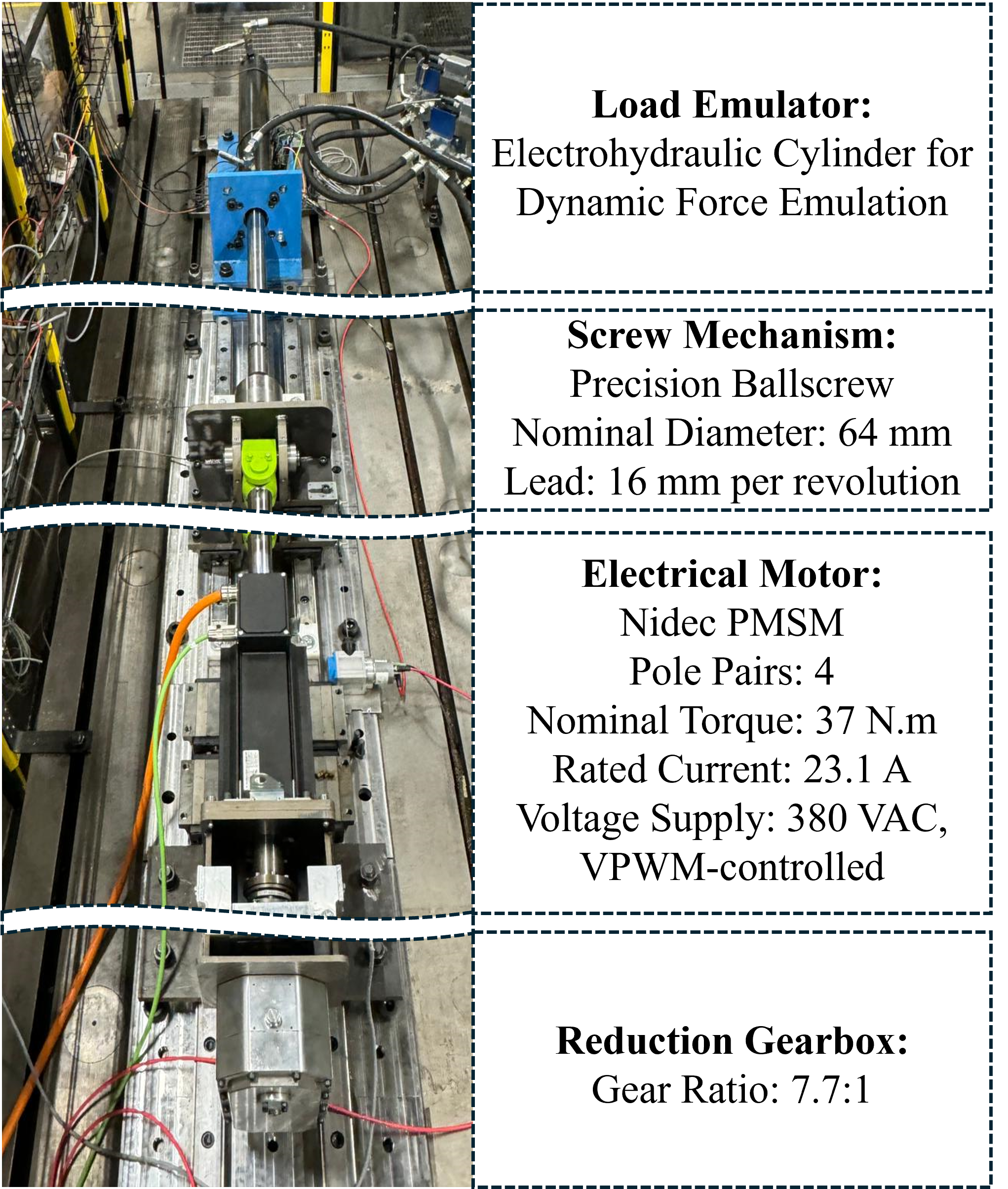}
\caption{Experimental setup used to validate the proposed control strategy. The testbed includes the Nidec PMSM, a reduction gearbox, and a high-precision screw forming the EMLA, as well as electrohydraulic cylinder as the load emulator.}
\label{fig:Experiment_testbed}
\end{figure}

\begin{figure}[ht]
    \centering
    \includegraphics[width=0.85\linewidth]{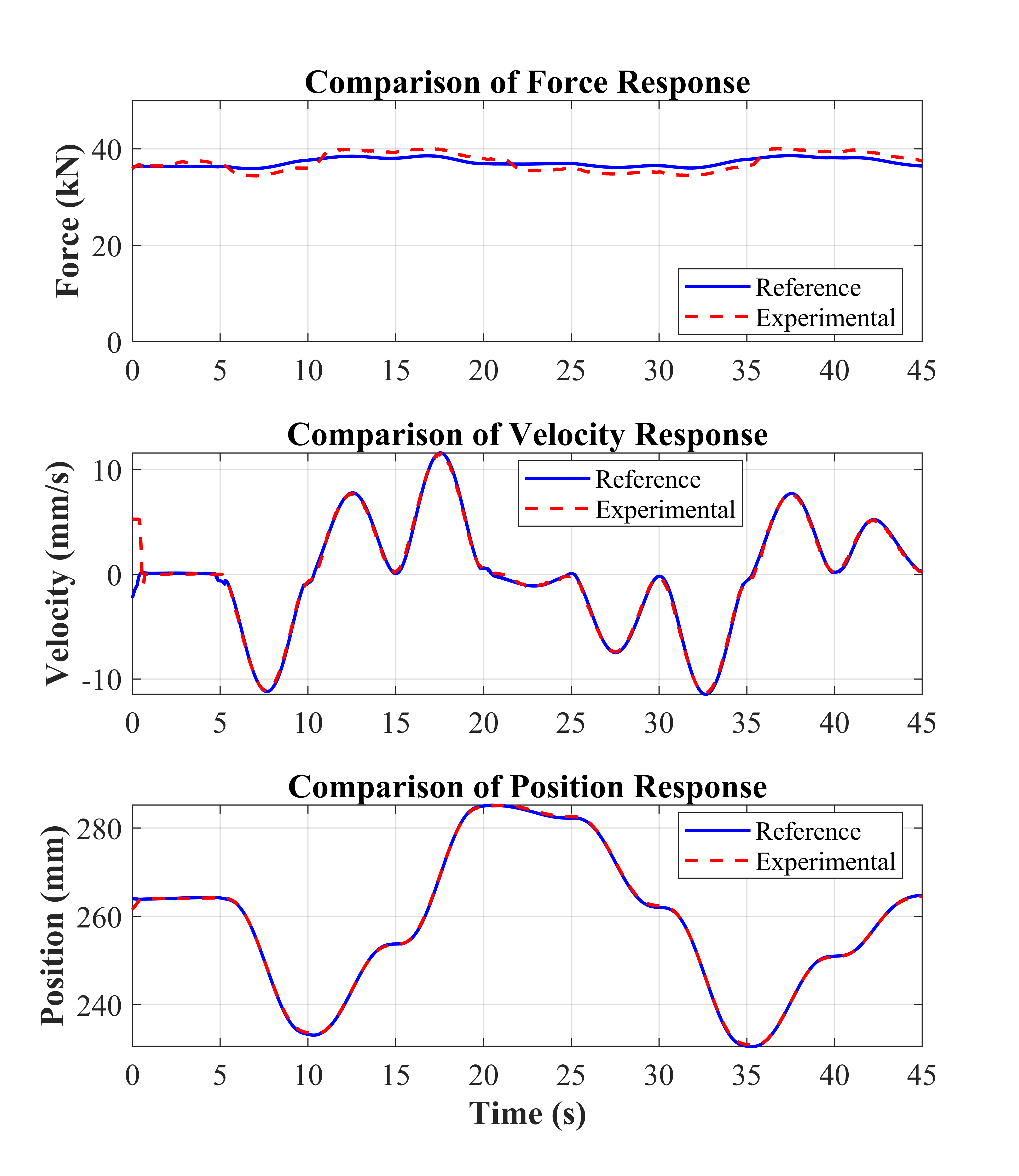}
    \caption{Experimental tracking results for the lift actuator under the cubic trajectory, illustrating tracking accuracy under three-dimensional motion.}
    \label{Fig:LiftISO}
\end{figure}

\begin{figure}[ht]
    \centering
    \includegraphics[width=0.85\linewidth]{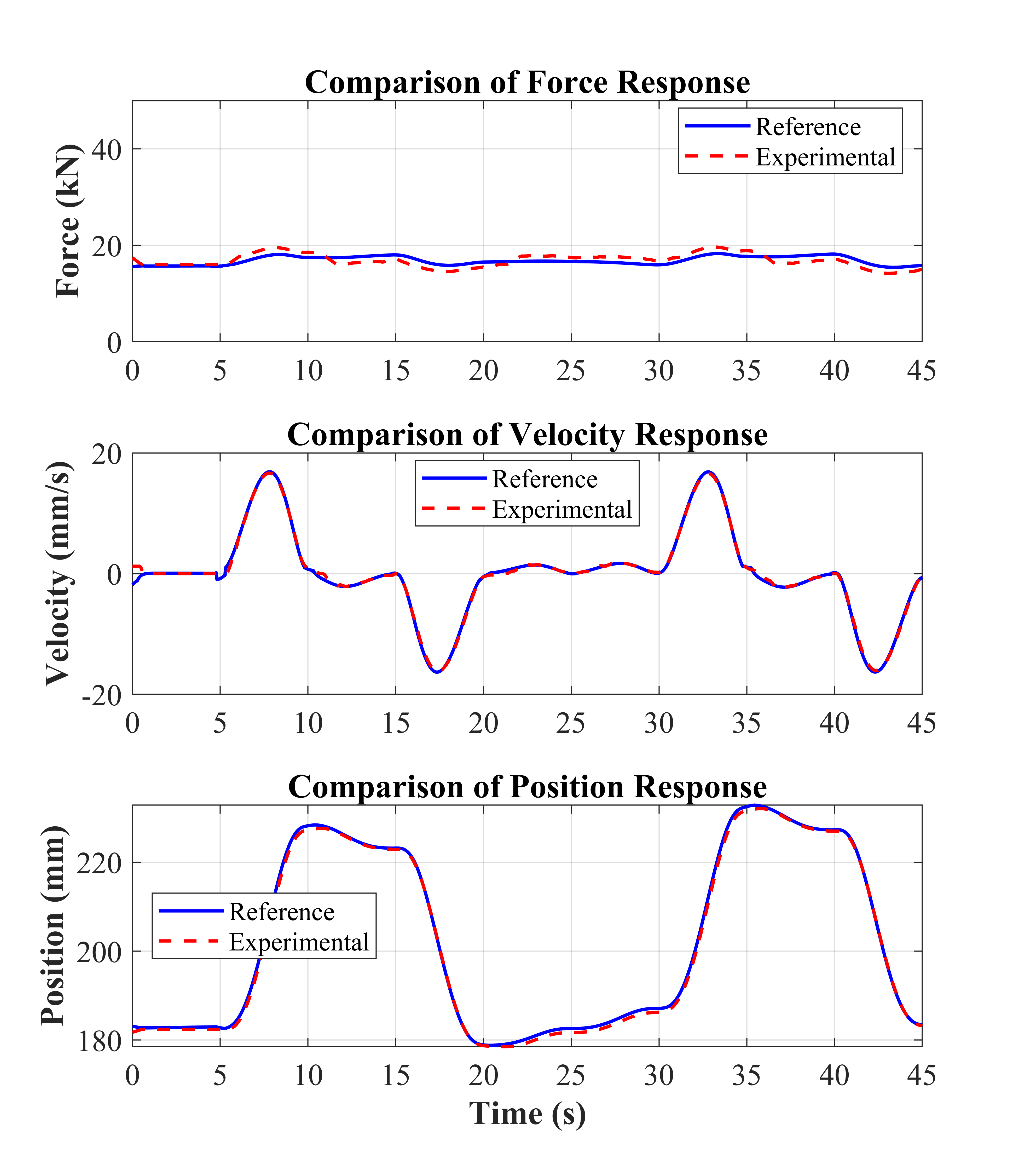}
    \caption{Experimental tracking results for the tilt actuator under the cubic trajectory, illustrating tracking accuracy under 3D motion.}
    \label{Fig:TiltISO}
\end{figure}

Fig.~\ref{fig:Experiment_testbed} shows the experimental testbed used for evaluating the lift and tilt actuators under these benchmark trajectories, while the results for the lift actuator tracking the cubic trajectory are illustrated in Fig.~\ref{Fig:LiftISO}, followed by the corresponding results for the tilt actuator in Fig.~\ref{Fig:TiltISO}. These figures present a comparison between the desired reference trajectories and the experimentally recorded force, velocity, and position responses.

\begin{figure}[ht]
    \centering
    \includegraphics[width=0.85\linewidth]{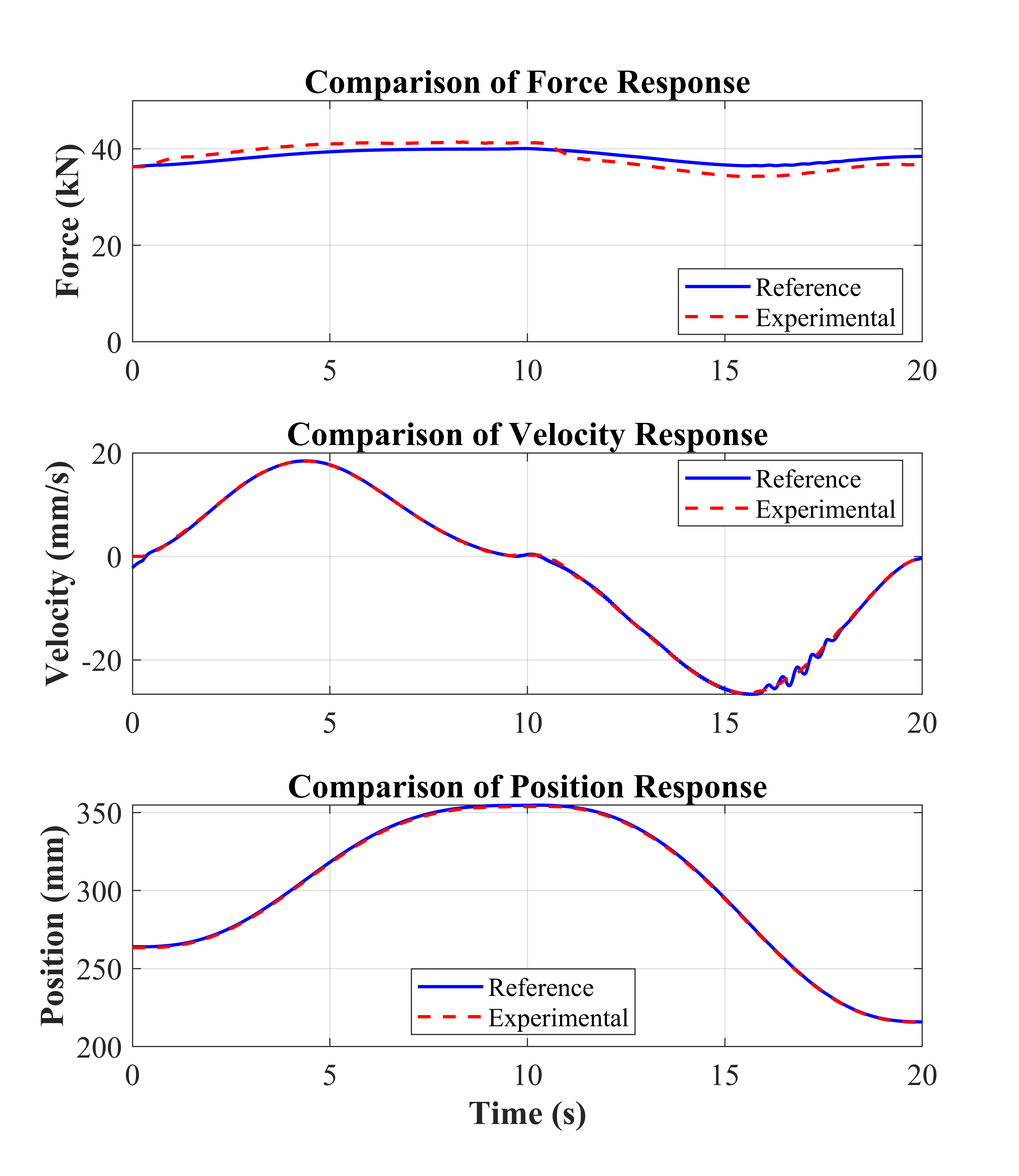}
    \caption{Experimental tracking results for the lift actuator under the planar triangular trajectory, illustrating performance under two-dimensional constrained motion.}
    \label{Fig:LiftTri}
\end{figure}

\begin{figure}[ht]
    \centering
    \includegraphics[width=0.85\linewidth]{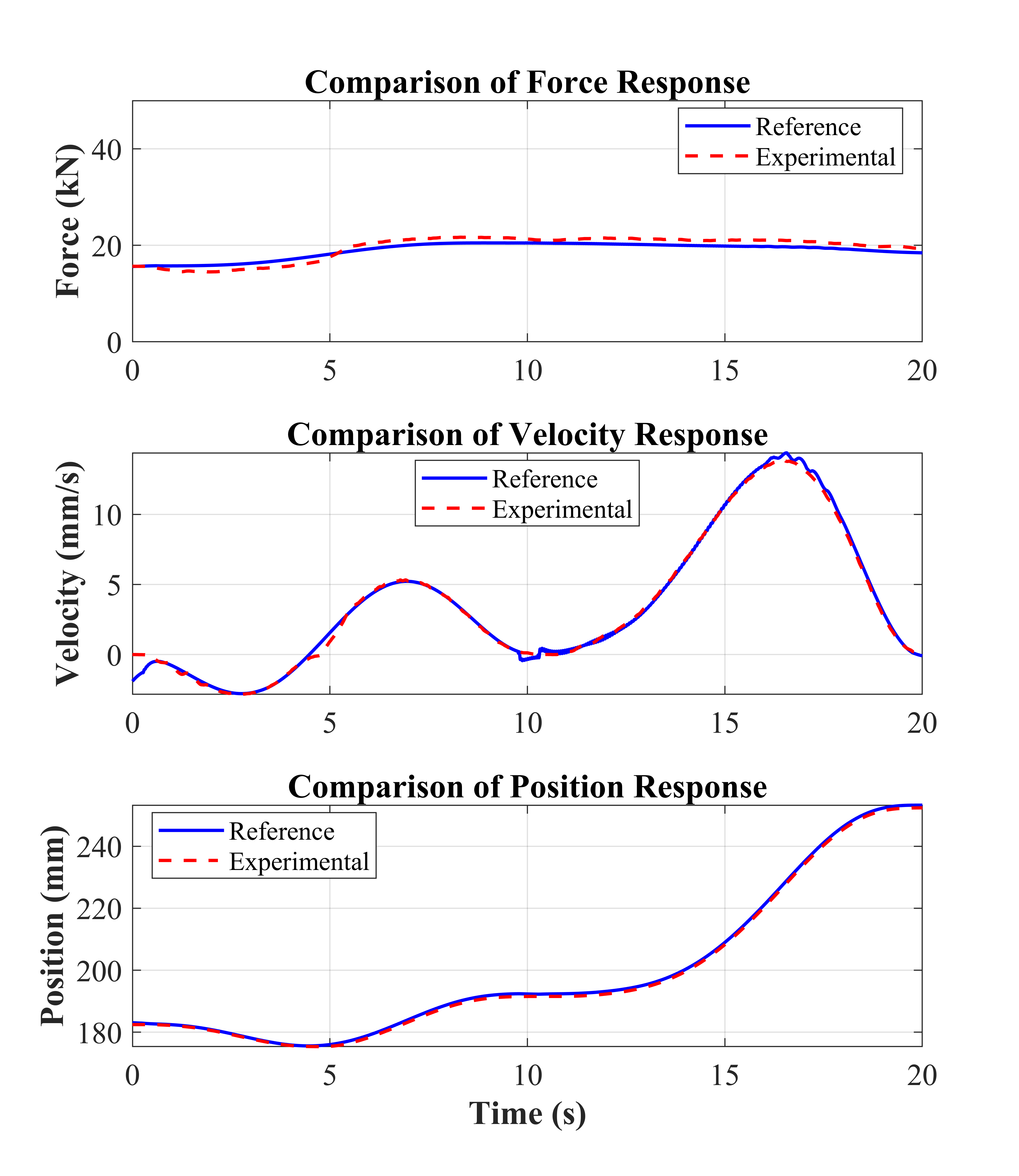}
    \caption{Experimental tracking results for the tilt actuator with planar triangular trajectory, illustrating performance under 2D constrained motion.}
    \label{Fig:TiltTri}
\end{figure}

To evaluate the performance further under 2D constraints, the planar triangular trajectory was applied to both joints. The resulting performance of the lift and tilt actuators under this trajectory is shown in Fig.~\ref{Fig:LiftTri} and Fig.~\ref{Fig:TiltTri}, respectively. In all cases, the proposed modular adaptive controller demonstrates accurate trajectory tracking and effective disturbance rejection, even in the presence of realistic dynamic loads.

\section{Conclusion}
\label{sec:conclusion}

This paper has presented a unified system-level modeling and control methodology for a 6-DoF HDRM, actuated exclusively by EMLAs. A hybrid actuator model was constructed by involving a physics-based description of the electromechanical dynamics and data-driven surrogate trained on a dedicated 1-DoF testbed. This surrogate-enhanced model was embedded in an extended VDC framework augmented by a natural adaptation law to guarantee robustness against parametric uncertainties.  The derived analytical manipulator model underpins a hierarchical control scheme that maps high‐level force and velocity objectives to low‐level voltage and current commands.

The complete architecture was evaluated in a multi-domain simulation with both the cubic benchmark and a custom planar triangular trajectory. In the simulation, the proposed adaptive modular controller achieved sub-centimeter Cartesian tracking accuracy, while experimental validation of the same 1-DoF platform under realistic load emulation confirmed a simulated performance and showcased precise motion tracking under dynamic force variations. A Lyapunov‐based stability proof verified that the combined rigid‐body VDC subsystems and low‐level EMLA controllers yield an asymptotically stable closed‐loop system.

These results illustrate that a hybrid modeling and VDC architecture can deliver high-precision, robust control of a fully electric HDRM while remaining modular and real-time implementable, positioning a fully electric HDRM as a viable alternative to traditional hydraulic systems in MWM applications. Future work will extend hardware-in-the-loop validation to the full 6-DoF manipulator, enable online adaptation during operation, and integrate holistic energy and thermal management to support sustained field deployments.

\bibliography{References}

\bibliographystyle{IEEEtran}

\vfill

\end{document}